%% file: main.tex
\documentclass[sn-mathphys]{sn-jnl}

\jyear{2025}%

\usepackage{amsfonts}
\usepackage{amsmath}
\usepackage{booktabs}
\usepackage{multirow}
\usepackage{amssymb}
\usepackage{bbding}
\usepackage{pifont}
\usepackage{wasysym}
\usepackage{color}
\usepackage{siunitx}
\usepackage{subfigure}
\usepackage{graphicx}
\usepackage{diagbox}
\usepackage{textcomp}
\usepackage{siunitx}
\usepackage{caption}
\usepackage{subcaption}
\usepackage{mdframed}  
\usepackage{xcolor}
\usepackage{listings}
\usepackage{longtable}
\usepackage{ragged2e}

\usepackage{amsthm}

\newenvironment{example} {\par\medskip\begin{mdframed}[backgroundcolor=white, linecolor=black, linewidth=1pt]}  
  {\end{mdframed}}

\theoremstyle{thmstyletwo}%

\theoremstyle{thmstylethree}%

\newcommand{\mol}{$\langle$\texttt{mol}$\rangle$}
\newcommand{\emol}{$\langle$\texttt{/mol}$\rangle$}
\newcommand{\pro}{$\langle$\texttt{protein}$\rangle$}
\newcommand{\epro}{$\langle$\texttt{/protein}$\rangle$}
\newcommand{\ant}{$\langle$\texttt{antibody}$\rangle$}
\newcommand{\eant}{$\langle$\texttt{/antibody}$\rangle$}
\newcommand{\rna}{$\langle$\texttt{rna}$\rangle$}

\newcommand{\dna}{$\langle$\texttt{dna}$\rangle$}
\newcommand{\erna}{$\langle$\texttt{/rna}$\rangle$}
\newcommand{\edna}{$\langle$\texttt{/dna}$\rangle$}

\newcommand{\material}{$\langle$\texttt{material}$\rangle$}
\newcommand{\ematerial}{$\langle$\texttt{/material}$\rangle$}

\newcommand{\product}{$\langle$\texttt{product}$\rangle$}
\newcommand{\eproduct}{$\langle$\texttt{/product}$\rangle$}

\newcommand{\reactant}{$\langle$\texttt{reactant}$\rangle$}
\newcommand{\ereactant}{$\langle$\texttt{/reactant}$\rangle$}

\newcommand{\ourM}{NatureLM}

\begin{document}

\title[\ourM{}]{
Nature Language Model: Deciphering the Language of Nature for Scientific Discovery}

\author{
{\centering
{
\ourM{} team\footnote{A full list of authors is available in the Author List section on Page \pageref{sec:authorlist}.} \\ Microsoft Research AI for Science\\
\url{https://NatureLM.github.io/}}
}}

\abstract{
Foundation models have revolutionized natural language processing and artificial intelligence, significantly enhancing how machines comprehend and generate human languages. Inspired by the success of these foundation models, researchers have developed foundation models for individual scientific domains, including small molecules, materials, proteins, DNA, RNA and even cells. However, these models are typically trained in isolation, lacking the ability to integrate across different scientific domains. Recognizing that entities within these domains can all be represented as sequences, which together form the ``language of nature'', we introduce Nature Language Model (\ourM{}), a sequence-based science foundation model designed for scientific discovery. Pre-trained with data  from multiple scientific domains, \ourM{} offers a unified, versatile model that enables various applications including: (i) generating and optimizing small molecules, proteins, RNA, and materials using text instructions; (ii) cross-domain generation/design, such as protein-to-molecule and protein-to-RNA generation; and (iii) top performance across different domains, matching or surpassing state-of-the-art specialist models.

\ourM{} offers a promising generalist approach for various scientific tasks, including drug discovery (hit generation/optimization, ADMET optimization, synthesis), novel material design, and the development of therapeutic proteins or nucleotides. We have developed \ourM{} models in different sizes (1 billion, 8 billion, and 46.7 billion parameters) and observed a clear improvement in performance as the model size increases.
}


\maketitle

\clearpage
\begin{figure*}[!htpb]
\centering
\includegraphics[width=\linewidth]{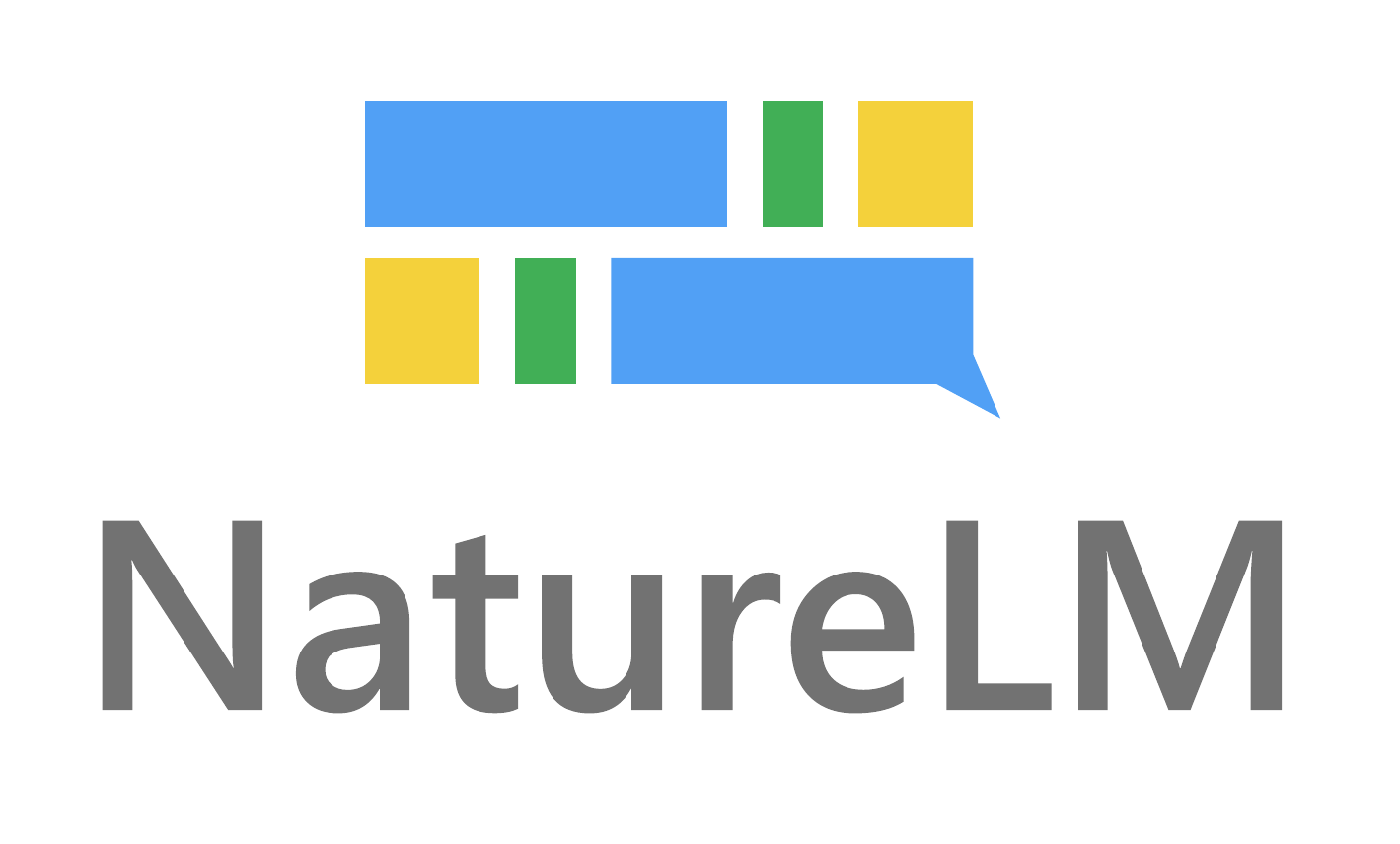}
\end{figure*}

Github Repository: \texttt{\url{https://github.com/microsoft/SFM/}}

Hugging Face Repositories: 



\texttt{\url{https://huggingface.co/microsoft/NatureLM-8x7B}}

\texttt{\url{https://huggingface.co/microsoft/NatureLM-8x7B-Inst}}

\clearpage
\tableofcontents  
\clearpage

\input{chapters/introduction.tex}
\clearpage
\input{chapters/method.tex}
\clearpage
\input{chapters/cmpd}
\clearpage
\input{chapters/protein}

\clearpage
\input{chapters/material}
\clearpage
\input{chapters/Nucleotide}
\clearpage
\input{chapters/prediction}
\clearpage
\input{chapters/performance_improvement}
\clearpage
\input{chapters/text}
\clearpage
\input{chapters/discussion}

\clearpage
\bibliography{sn-bibliography}

\clearpage
\appendix
\setcounter{figure}{0}  
\renewcommand{\thefigure}{S\arabic{figure}}
\setcounter{table}{0}  
\renewcommand{\thetable}{S\arabic{table}} 
\input{chapters/SI}

\end{document}

%% file: chapters/introduction.tex
\section{Introduction}
\label{sec:introduction}
Foundation models, including the GPT \cite{brown2020languagemodelsfewshotlearners,gpt4technicalreport,openai2024gpt4ocard}, Gemini \cite{geminiteam2024geminifamilyhighlycapable,geminiteam2024gemini15unlockingmultimodal}, Phi \cite{abdin2024phi3technicalreporthighly,phi4report}, Llama \cite{dubey2024llama3herdmodels}, Mistral \cite{jiang2023mistral7b,jiang2024mixtralexperts}, DeepSeek \cite{deepseekv3,deepseekai2025r1}, and Qwen \cite{yang2024qwen2technicalreport,qwen2025qwen25technicalreport}, represent a transformative advancement in artificial intelligence. 
These models, trained on massive web-scale datasets, are designed to serve as general-purpose tools, capable of handling a wide range of tasks with a single architecture. The most notable capabilities of foundation models include their abilities to perform tasks without fine-tuning, a phenomenon known as zero-shot learning, and their few-shot learning abilities which allow them to adapt to new tasks by drawing inferences from just a few examples. 

Despite their success in general-purpose tasks, early investigations \cite{ai4science2023impactlargelanguagemodels} highlight significant room for improvement in scientific tasks involving small molecules, proteins, DNA, RNA, or materials. In particular, foundation models struggle with precise quantitative predictions (e.g., ligand-protein binding affinity, protein-protein interactions, DNA properties) \cite{ai4science2023impactlargelanguagemodels}, as well as the rational design of small molecule compounds, proteins, or materials. Ensuring the scientific accuracy of outputs from these models remains a grand challenge.

Recently, there has been a concerted effort to develop large-scale foundation models specifically tailored for scientific tasks. These approaches can be broadly divided into four categories:
\begin{enumerate}
\item Domain-specific foundation models. These models, such as ProGen \cite{ProGen} and ESM3 \cite{esm3} for proteins, DNABERT \cite{zhou2024dnabert2} and Evo \cite{evo2024science} for DNA sequences, scGPT \cite{Cui2024scgpt} for single-cell data, and chemical language models \cite{liu2023molxpt,segler2018generating} for small molecules, are trained specifically on token sequence representations for individual scientific domains.
\item Fine-tuned general-purpose models. This approach adapts well-trained large language models for specific scientific domains, as demonstrated by Tx-LLM \cite{chaves2024tx} for small molecules and ProLLAMA \cite{lv2024prollamaproteinlanguagemodel} for proteins.
\item Scientific data enhanced large language models (LLMs). This approach, exemplified by works such as  \cite{BioGPT2022Luo,liu2023molxpt,galactica2022}, trains LLMs from scratch mainly with text data and a small portion of scientific data.
\item Integration of specific scientific modules. In this approach, external modules, such as pre-trained molecular or protein encoders, are integrated into general-purpose models (e.g., Llama) via lightweight adapters \cite{drugchat,proteinchat}.
\end{enumerate}

While these approaches have made considerable progress, they do face notable limitations. Domain-specific models (approach \#1) are restricted to their respective fields, limiting their ability to capture interdisciplinary insights for cross-domain applications. Fine-tuning general-purpose models (approach \#2) and scientific data enhanced LLMs (approach \#3) show promise but are often constrained by small-scale scientific datasets, e.g., around 90\% text data and only 10\% scientific data in \cite{galactica2022}, which hinders the models' capacity to capture the complexity of scientific tasks. The integration of external modules (approach \#4) faces challenges in aligning inputs effectively with large language models, and most implementations opt for limited fine-tuning with small datasets, leaving the core models largely unchanged.

These limitations emphasize the necessity for a science foundation model, to fulfill the sophisticated demands of scientific research. A model of this kind must not only be highly proficient in producing precise scientific predictions, but also adept at designing and optimizing scientific entities conditioned on context information. A good science foundation model ought to have the capacity to handle a diverse range of inputs. These inputs can span from literature text, to scientific sequence data such as protein or DNA sequences, and further to structural data like 3D protein/DNA structures and their dynamic behaviors. In the present study, our focus is on sequence-based data for representing biological, chemical, material systems, and textual human language (e.g., English):
\begin{itemize}
\item DNA, RNA, and proteins, which are often referred to as the ``language of nature", are intrinsically represented by sequences. Additionally, many other scientific entities like small molecules and materials can be effectively represented as sequences through well-established domain-specific techniques \cite{Weininger1988}.
\item Sequence data is highly compatible with the current mainstream large language models (LLMs). Through the continuous pre-training of LLMs, we are able to utilize the scientific knowledge embedded in these general-purpose LLMs to tackle complex scientific challenges.  
\item Sequential data provides remarkable flexibility when combined with autoregressive paradigms \cite{bond2021deep,yenduri2024gpt}%
. These paradigms, which are extensively employed in generative models, are capable of effectively modeling the highly complex distributions of any scientific object that can be presented in the form of a sequence.  
\end{itemize}

\begin{figure}[!htbp]
    \centering
    \includegraphics[width=\linewidth]{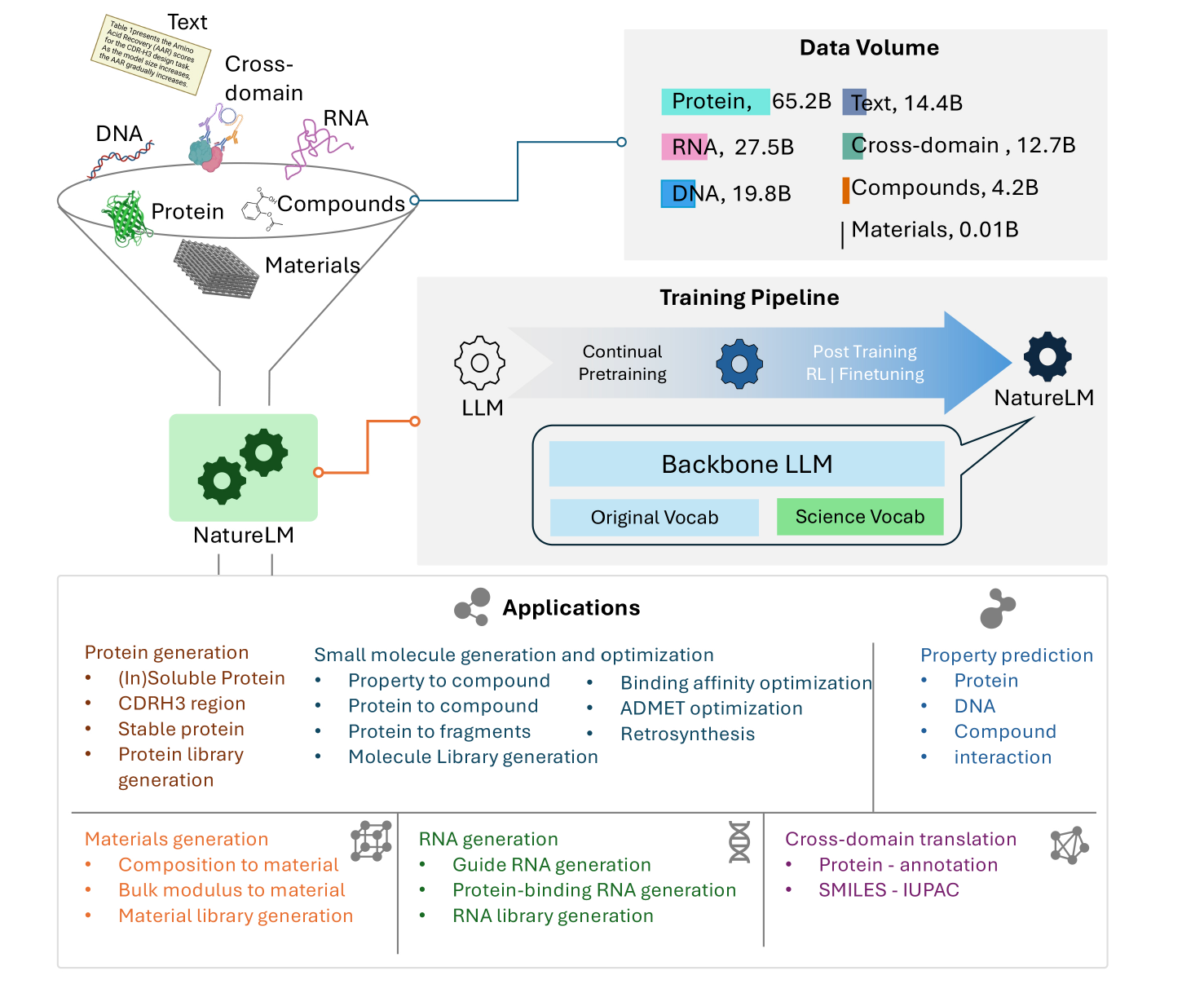}
    \caption{\ourM{} is a GPT-style generative model trained on a diverse range of data, including small molecule compounds, proteins, DNA, RNA, materials, and both general and scientific texts, amounting to a total of 143 billion tokens. It is built on existing large language models by integrating new vocabularies for scientific entities and jointly pre-training all components. After the pre-training, the model undergoes additional instruction tuning using millions of curated instructions from scientific fields. Options for reinforcement learning and dedicated fine-tuning are also available to boost performance on specific tasks. Users can engage with \ourM{} through natural language inputs. The model excels in various domains, achieving top results in tasks such as retrosynthesis (Section \ref{sec:retro}), SMILES-to-IUPAC translation (Section \ref{sec:smiles_iupac}), protein generation (Section \ref{sec_prot_generation}) and material property prediction (Section \ref{sec:dedicate_tune_matbench}), often matching or exceeding the capabilities of state-of-the-art specialized models.}
    \label{fig:naturelm_overview}
\end{figure}

\ourM{} is designed to handle the complexity of small molecules, proteins, DNA, RNA, materials, and their associated textual information. An overview of \ourM{} is in Fig. \ref{fig:naturelm_overview}. \ourM{} follows the Transformer decoder architecture and is trained on a corpus of 143 billion tokens collected from various scientific domains (Fig. \ref{fig:pretrain_data_pie}). Our experiments demonstrate that \ourM{} significantly outperforms general-purpose foundation models for scientific tasks. Specifically, \ourM{} excels in tasks such as:
\begin{enumerate}
    \item Following textual instructions to generate and optimize scientific molecular entities.
\item Performing cross-domain generation tasks, such as designing small molecules or RNA binders for proteins as well as designing guide RNA sequences given a DNA template for CRISPR systems.
\item  Achieving top performance on generation and translation tasks, such retrosynthesis (Section \ref{sec:retro}), SMILES-to-IUPAC translation (Section \ref{sec:smiles_iupac}), protein generation (Section \ref{sec_prot_generation}), matching or surpassing state-of-the-art specialist models. 
\end{enumerate}

\begin{figure}[!htb]
\centering
\includegraphics[trim= 1cm 2cm 3cm 4cm, clip, width=\linewidth]{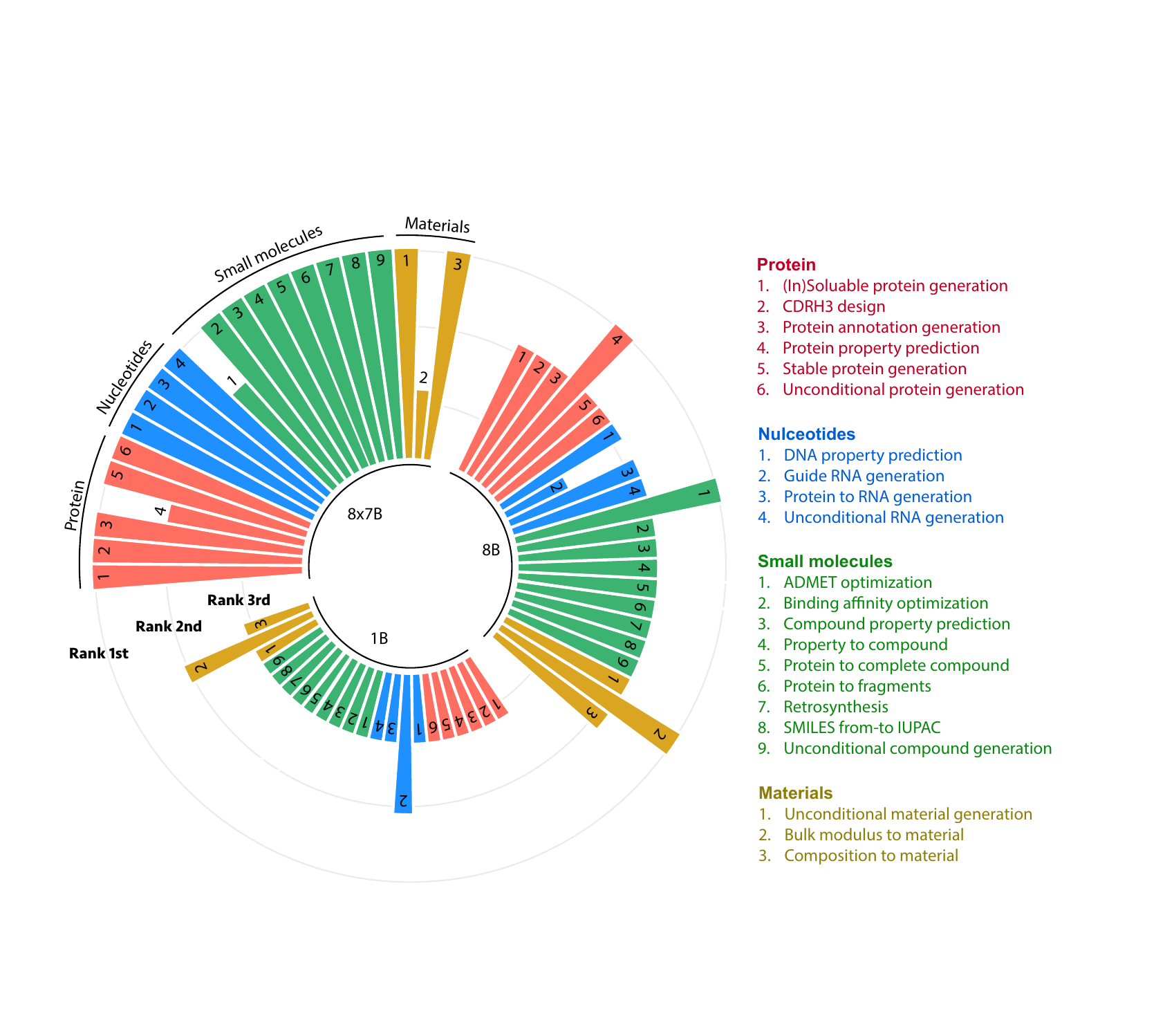}
\caption{The scaling effect in \ourM{} is obvious. The chart depicts the overall ranking of models with varying sizes, where a better rank is represented by the ``outsider'' bar. The 8x7B model achieves top performance in 19 tasks, while the 8B model excels in 3 tasks. 18 categories exhibited performance improvements with increasing model size (i.e., 8x7B demonstrated the best performance, followed by 8B, and then 1B), highlighting the potential of large foundation models for scientific applications. }
\label{fig:overallrank}
\end{figure}

To investigate the scalability of \ourM{} with respect to model size, we trained three versions of \ourM{} with varying parameter configurations. As illustrated in Fig. \ref{fig:overallrank}, among the 22 categories of tasks evaluated, 18 categories exhibited clear improvements with increasing model size (i.e., 8x7B demonstrated the best performance\footnote{The 8x7B model is a Mixture-of-Experts (MoE) model \cite{jiang2024mixtralexperts}, composed of eight expert models, each with 7 billion parameters. A portion of these expert models is shared across all models, resulting in a total parameter count of 46.7 billion.}, followed by 8B, and then 1B), underscoring the potential of large foundation models for scientific discovery. Additionally, we demonstrate the efficacy of \emph{reinforcement learning} in enhancing the post-training performance of \ourM{} for molecular property optimization and \emph{dedicated finetuning} for retrosynthesis.

In summary, \ourM{} represents a significant step towards building a generalist model across multiple scientific domains. By harnessing the capabilities of text-based instructions, \ourM{} serves as a powerful tool for scientific discovery, enabling cross-domain generation and optimization in areas such as drug discovery, materials science, and the development of therapeutic proteins and nucleotides. 
Ideally, a foundation model should support a broad range of tasks while demonstrating strong zero-shot and few-shot capabilities.
\ourM{} shows great promise, but its language capabilities and few-shot learning skills still lag behind leading large language models. We will address these limitations in future iterations, positioning \ourM{} as an essential component in the continued evolution of science foundation models.

%% file: chapters/method.tex
\section{Method}

\subsection{Pre-training data}\label{sec:pretraining}
\begin{figure}[!htpb]
\centering
\includegraphics[trim=3cm 0cm 6cm 0cm, clip, width=0.8\linewidth]{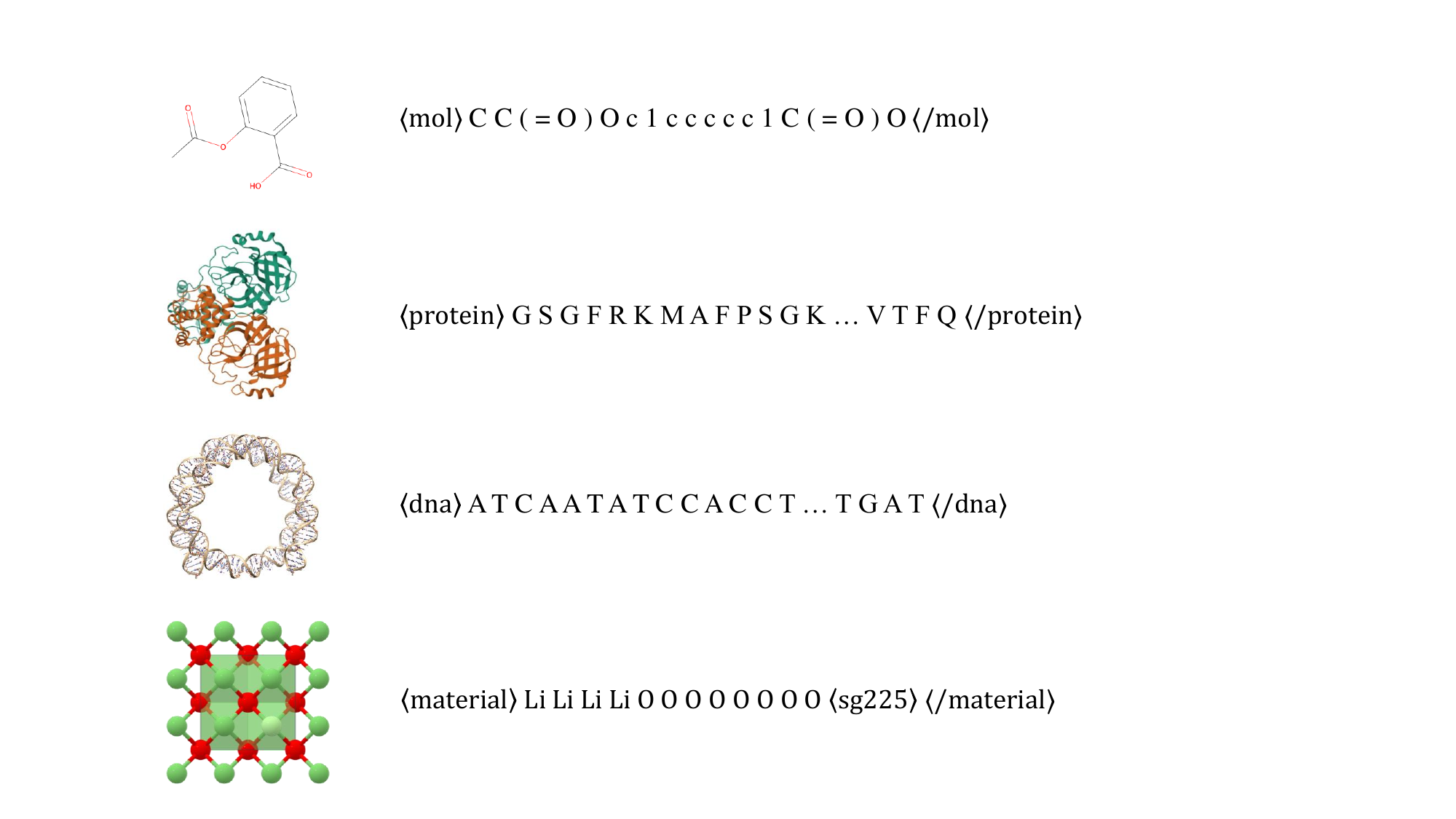}
\caption{Example data from each domain. The small molecule is Aspirin (PubChem CID: 2244) and visualized by RDKit \cite{greg_landrum_2025_14779836}. The protein snapshot is from the PDB bank with ID 7CAM \cite{Wang2020-vw}. The DNA structure is split into chain I and chain J from PDB 1KX5 \cite{Davey2002im} and visualized by UCSF Chimera \cite{chimera2004software}. The material snapshot is from the material project with ID mp-1960 \cite{osti_1194803}.}
\label{fig:example_pretrain_data}
\end{figure}

The pre-training data includes text, small molecules, proteins, materials, DNA, and RNA, all in the format of sequences:
\begin{enumerate}
\item Small molecules are converted into Simplified Molecular Input Line Entry System (SMILES) notations, obtained by applying depth-first search algorithm on molecular graphs to yield a linear representation of the chemical structure \cite{Weininger1988}. The SMILES are tokenized by the commonly used regular expression for molecules\footnote{\url{https://github.com/microsoft/DVMP/blob/main/molecule/tokenize_re.py\#L11}}.
\item Proteins, DNA and RNA are depicted using FASTA format, which sequentially lists the amino acids or nucleotides. The sequences are tokenized into individual units, with proteins broken down into their constituent amino acids and DNA/RNA into their respective nucleotides.
\item For crystal material data, both the chemical composition and the associated space group number\footnote{\url{https://en.wikipedia.org/wiki/List_of_space_groups}} are flattened into a sequence. For example, consider the material from the material project with ID mp-1960, as shown in Fig. \ref{fig:example_pretrain_data}. This material has 12 atoms in its cell, consisting of 4 Li and 8 O atoms. We flatten this information as depicted in the figure. The space group is Fm3m, which corresponds to the International Space Group Number 225, and we represent it with $\langle$sg$\cdots$$\rangle$. 
\end{enumerate}

An example of the data is in Fig. \ref{fig:example_pretrain_data}. The vocabulary sizes of small molecules, proteins, material, DNA and RNA are 1401, 26, 396, 16 and 16 respectively. To differentiate scientific entities from regular text, each scientific sequence is enclosed by a pair of special tokens: \mol{}$\cdots$\emol{} for small molecules, \pro{}$\cdots$\epro{} for proteins, \material{}$\cdots$\ematerial{} for materials, \dna{}$\cdots$\edna{} for DNA and \rna{}$\cdots$\erna{} for RNA. Specifically, we use \product{}$\cdots$\eproduct{} and \reactant{}$\cdots$\ereactant{} to represent products and reactants for small molecules in chemical reactions. We use \ant{}$\cdots$\eant{} to denote antibodies. For example, benzene is represented by \mol{}c1ccccc1\emol{}. More examples can be found within the following sections.

\begin{figure}[!htpb]
\centering
\includegraphics[width=0.7\linewidth]{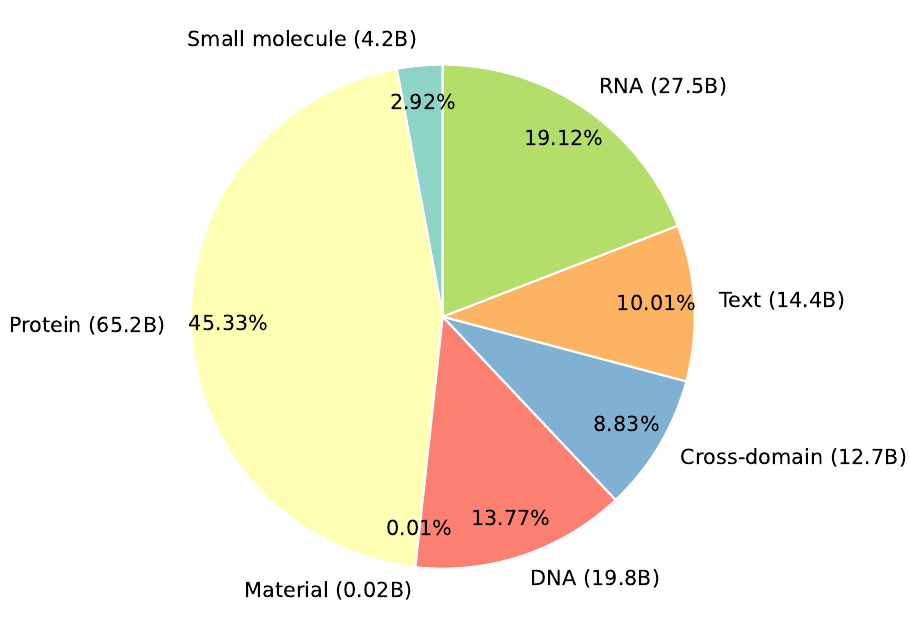}
\caption{Distribution of the pre-training data, measured by the number of tokens of each category.}
\label{fig:pretrain_data_pie}
\end{figure}

The pre-training data contains single-domain sequences and cross-domain sequences. A single-domain sequence comes from one domain, such as pure text sequences, SMILES sequences for small molecules, and FASTA sequences for proteins. A cross-domain sequence includes data from two different domains, building connections across domains. The distribution of our pre-training data is visualized in Fig. \ref{fig:pretrain_data_pie} and more details are left in Table \ref{tab:statistics_pretrain_data}. 

Our cross-domain data is organized into three categories.
\begin{enumerate}
    \item Interleaved Sequences: Inspired by \cite{liu2023molxpt}, we process scientific literature by initially employing a named entity recognition tool, BERN2 \cite{bern2}, to identify the mentions of small molecules and proteins within the corpus. These entities are then converted into their corresponding SMILES and FASTA sequences.
    Consequently, the small molecules and proteins are wrapped by text, creating an interleaved data structure that bridges the gap between textual information and scientific data. We also develop a quality filter to remove low-quality sentences.
    This formulation is also similar to the one that has been used in multi-modal LLMs where image tokens are wrapped inside text~\cite{Zhu2023MultimodalCA,Zhan2024AnyGPTUM,team2024chameleon}. We provide an example of interleaved sequences.
    \begin{example}
        A prospective, randomized clinical trial was performed to study the efficacy of povidone iodine \mol{}C=CN1CCCC1=O.II\emol{} ( Betadine \mol{}C=CN1CCCC1=O.II\emol{}) suppositories for the treatment of bacterial vaginosis (BV) in comparison to capsules containing lactobacilli (Dderlein Med).
    \end{example}
    \item Parallel Text and Scientific Entities: Leveraging databases such as PubChem\footnote{\url{https://pubchem.ncbi.nlm.nih.gov/}}, UniProt\footnote{\url{https://www.uniprot.org/}}, and NCBI\footnote{\url{https://www.ncbi.nlm.nih.gov/}}, we extract descriptive information about specific proteins and small molecules. Additionally, from the Materials Project website\footnote{\url{https://next-gen.materialsproject.org/}}, material-related data such as bandgap, energy above hull, and other properties are gathered and translated into textual descriptions. This process results in parallel datasets that align scientific facts with their textual counterparts, enhancing the richness of the information. 
    \item Linking DNA with Proteins Through the Central Dogma: For DNA sequences, we identify segments that can be transcribed and translated into proteins, following the central dogma of molecular biology. These identified DNA segments are then replaced with the equivalent protein sequences, establishing a direct connection between the genetic blueprint and its functional protein products. This method not only reflects the biological process but also creates a dataset that encapsulates the relationship between nucleotide and amino acid sequences. We retrieved data from the RefSeq database\footnote{\url{https://www.ncbi.nlm.nih.gov/refseq/}} and extracted protein sequences, including their isoforms, from annotated genes, along with their corresponding flanking DNA sequences.
\end{enumerate}

\begin{table}[!htbp]
\centering
\begin{tabular}{lcccc}
\toprule
& Samples & Tokens & Samples & Tokens \\
& (by million) & (by billion) & (\%) & (\%) \\ 
\midrule
Interleaved Sequence & 4.3 & 4.0 & 10.2 & 31.3 \\
Text-SMILES & 33.0 & 3.0 & 78.8 & 24.0 \\
Text-protein & 1.9 & 1.4 & 4.6 & 10.8\\
Text-material & 1.7 & 0.2 & 4.0 & 1.6 \\
DNA-protein & 1.0 & 4.1 & 2.4 & 32.3 \\
\midrule 
Total& 41.9 & 12.7 & 100 & 100 \\
\bottomrule
\end{tabular}
\caption{Statistics of cross-domain data.}
\label{tab:statistics_multimodal_data}
\end{table}

The statistics of cross-domain data is in Table \ref{tab:statistics_multimodal_data}. Both interleaved sequences and text-science parallel data are types of cross-domain data that aim to facilitate cross-domain connections. For interleaved sequences, the sources are literature, which covers a broader range of general topics and wider domains. In contrast, parallel data sources are existing databases that focus on specific properties. Although the topics covered by parallel data are not as diverse as those in interleaved sequences, the amount of data available for each given property is larger. These distinctions highlight the complementary nature of the two types of cross-domain data.

\subsection{Post-training data}\label{sec:supervised_ft_data}
\begin{figure}[!htbp]
    \centering
    \includegraphics[width=0.7\linewidth]{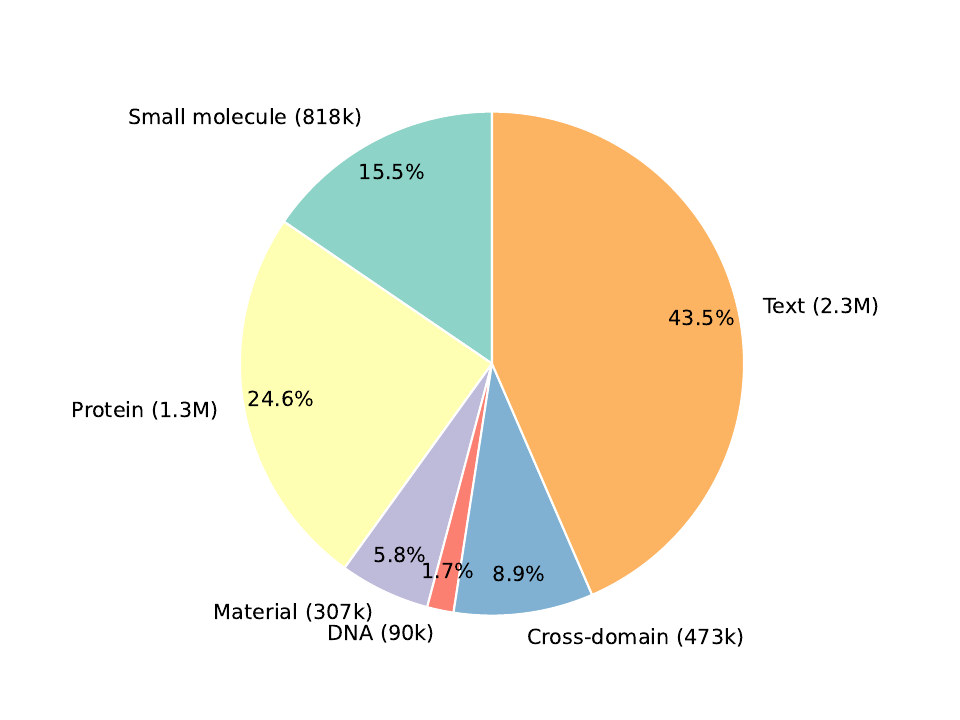}
    \caption{Statistics of post-training data, measured by the number of sequences.}
    \label{fig:sft_data}
\end{figure}

We curated a dataset for post-training with about 5.1 million instruction-response pairs encompassing six domains, small molecules, proteins, materials, DNA, RNA and general text (Figure \ref{fig:sft_data}). The dataset includes over 60 categories of tasks. For each task category, multiple prompts were manually crafted to form diverse instruction-response pairs, covering essential scientific tasks such as molecular optimization, antibody design, and guide RNA design. We provide two examples below:
\begin{example} 
$\rhd$ {\bf Example 1:}

\noindent\textbf{Instruction:}\\ 
\texttt{Create a guiding RNA to interact with the DNA sequence }\\
\noindent\dna{}CCCAGAGC$\cdots$GGGCCTGTC\edna{}.\\   \noindent\textbf{Response}:\rna{}AGGGGACAAACCTTCATCCA\erna{}

\noindent$\rhd$ {\bf Example 2}
\noindent\textbf{Instruction: }\\\texttt{What can be produced when these reactants combine?} $\langle$\texttt{reactants}$\rangle$CNC.C1(=O)CCCC1Cc1c[nH]c2ccc(C\#N)cc12$\langle$\texttt{/reactants}$\rangle$\\   
\textbf{Response: }\\$\langle$\texttt{product}$\rangle$CN(C)C1CCCC1Cc1c[nH]c2ccc(C\#N)cc12$\langle$\texttt{/product}$\rangle$
    \end{example}  

The text data were sourced from open-source instruction tuning datasets like OIG \footnote{\url{https://huggingface.co/datasets/laion/OIG}}, aiming to ensure that the model not only excels in scientific tasks but also maintains general language capabilities.

\subsection{Model architecture}
\ourM{} models are built upon well-trained large language models (LLMs) with some additional parameters for newly introduce scientific tokens. We used Llama 3 8B \cite{dubey2024llama3herdmodels} and Mixtral 8x7B \cite{jiang2024mixtralexperts} to initialize the main part of \ourM{} and continued pre-training using the science data described in Section \ref{sec:pretraining}. Additionally, we trained a model with 1B parameters, which replicates the structural design of Llama 3 but with a reduced number of layers and smaller hidden dimensions. The pre-training of \ourM{} 1B begins with a random selection of 300 billion pure text tokens from the SlimPajama dataset \cite{cerebras2023slimpajama}, followed by the science data we collected in Section \ref{sec:pretraining}. This approach ensures a consistent training methodology across all three models. The details of the model architecture are provided in Table \ref{tab:model_arch}.

\begin{table}[!htbp]
\centering
\begin{tabular}{cccccccc}
\toprule
Model Parameters & 1B & 8B & 8x7B \\
\midrule
Hidden Dimensions & 2048 & 4096 & 4096 \\
FFN Dimensions & 5504 & 14336 & 14336 \\
Attention Heads & 32 & 32 & 32 \\
KV Heads & 8 & 8 & 32 \\
Number of Layers & 16 & 32 & 32 \\
Vocabulary Size & 130,239 & 130,239 & 38,078 \\
\bottomrule
\end{tabular}
\caption{Model parameters of different sizes of \ourM{}.}
\label{tab:model_arch}
\end{table}

\subsection{Continued pre-training}
To address the intricate comprehension required for scientific tasks, \ourM{} introduces specific tokens for scientific entities. Consequently, we augment the vocabulary of the chosen LLMs. The embedding weights for these newly introduced tokens are randomly initialized. Directly tuning from pre-training usually causes instability and potentially compromises the language capabilities of the original LLMs. This is primarily due to the introduction of new tokens and the mismatch between the well-trained text tokens and randomly initialized scientific tokens. 

To circumvent this issue, we have devised a two-stage pre-training procedure:

Stage 1: Training is exclusively concentrated on the newly introduced tokens. During this phase, the parameters of the existing model are frozen. This allows the new tokens to adapt to the model gradually, mitigating the risk of instability.

Stage 2: Once the new tokens are adequately trained, we proceed to the second phase where the entire network, including both new and existing parameters, is trained. This joint optimization process ensures that the new tokens are seamlessly integrated with the existing ones, enhancing the model's overall performance.

This two-stage training approach not only fosters a thorough understanding of the scientific domain but also preserves the integrity and robustness of the underlying language model by preventing potential instabilities. The detailed training recipe is summarized in Table \ref{tab:training_recipe}.

\begin{figure}[!htbp]
    \centering
    \includegraphics[width=0.75\linewidth]{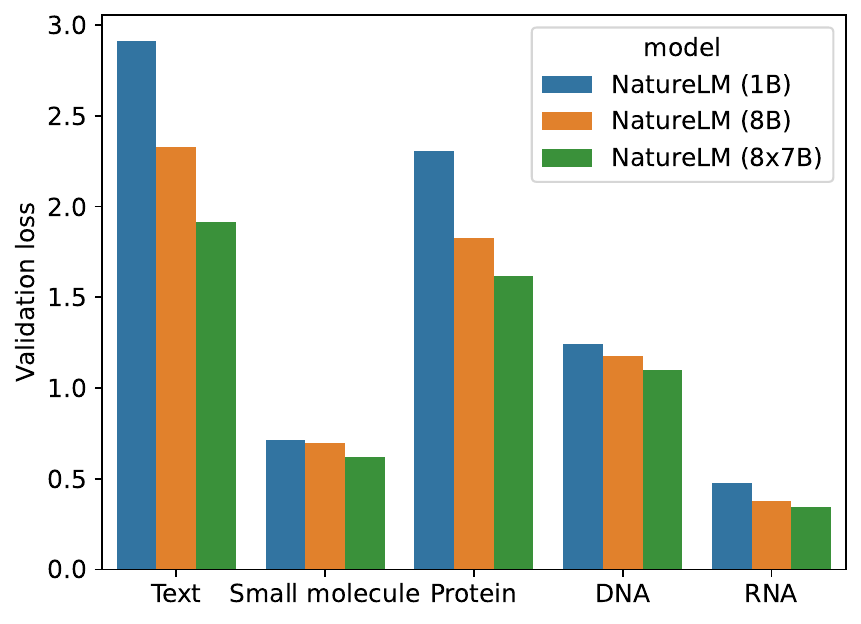}
    \caption{Validation loss for the 1B, 8B, and 8x7B \ourM{} models. Larger models result in smaller validation losses across all domains. \ourM{} (8B) is short for ``Llama 3 8B \ourM{}'' throughout this paper.}
    \label{fig:dev_loss}
\end{figure}

The validation loss for the three versions of the models is illustrated in Fig. \ref{fig:dev_loss}. All validation losses decrease as the model size increases. This indicates that larger models are better at capturing the underlying patterns or rules in the data, which is expected due to their increased capacity. The most significant decreases are observed in the text and protein data, suggesting that these datasets benefit more from enlarged models.

\subsection{Post-training}\label{sec:post_train}
In the post-training phase, we mainly employ supervised fine-tuning (SFT) using the instruction-response pair data outlined in Section~\ref{sec:supervised_ft_data}. These pairs are structured into sequences utilizing the template  ``\texttt{Instruction: \{instruction\}\textbackslash{}n\textbackslash{}n\textbackslash{}nResponse: \{response\}}'' where ``\{\texttt{instruction}\}'' and ``\{\texttt{response}\}'' serve as placeholders. During the model optimization, the training loss is computed solely on the response part of the sequence. Unlike in the pre-training phase, each sequence contains a single instruction-response pair rather than multiple pairs packed into one sequence. Empirical evidence suggests that this approach aids in stabilizing the post-training process. The 1B and 8B models are trained for 20k steps, while the 8x7B model is trained for 7.8k steps (due to resource constraint). We also explore using RLHF after supervised finetuning and results are discussed in Section \ref{sec:RL}.

\subsection{Inference acceleration}
As \ourM{} will be tested on many downstream tasks, we need to accelerate inference speed to reduce computational cost. We adopted the following approaches: (1) PagedAttention \cite{pagedattention}, which optimizes LLM serving by partitioning the key-value (KV) cache into fixed-size, non-contiguous blocks, reducing memory fragmentation and enabling efficient memory sharing; and (2) Selective Batching \cite{orca}, which batches compatible operations while handling attention separately, allowing for flexible and efficient processing of requests with varying input lengths. We employed the vLLM framework \cite{vllm2024} to serve \ourM{} models, leveraging its implementations of both PagedAttention and Selective Batching. These optimizations were applied to the 1B, 8B, and 8×7B models. Consequently, the inference speed for the \ourM{} 8×7B model reached approximately 525 tokens per second with Brain Float 16 (BF16) precision on two NVIDIA A100 GPUs.

%% file: chapters/cmpd.tex
\section{Small molecule tasks}\label{sec:smallmol}
We assess the capabilities of \ourM{} in terms of small molecule generation from the following perspectives: 
\begin{enumerate}
    \item The unconditional generation ability (Section \ref{sec:smallmol_uncondition});
    \item The basic properties (such as QED, TSPA, etc.) to small molecule generation (Section \ref{sec:basic_property_to_mol});
    \item The translation between small molecule SMILES and IUPAC (Section \ref{sec:smiles_iupac}); 
    \item Utilize \ourM{} to aid the drug discovery pipeline, which encompasses the generation and optimization of hit compounds (Section \ref{sec:tamgen}), optimization of binding affinity (Section \ref{sec:binding_affinity}), ADMET optimization (Section \ref{sec:admet}), and the synthesis routes of the compounds (Section \ref{sec:retro}).
\end{enumerate}

\subsection{Unconditional molecular generation}\label{sec:smallmol_uncondition}
We input the special token \mol{} to \ourM{} and let the model generate SMILES. The generation process stops upon encountering the special token \emol{}. We assess the validity of the generated SMILES by checking if they can be converted into molecules using RDKit. Additionally, we evaluate the uniqueness of the valid SMILES by calculating the ratio of unique valid SMILES to the total valid SMILES.

The evaluation results are presented in Table \ref{tab:unconditional_mol_eval}. The results demonstrate a clear trend: as the model size increases, the performance in terms of validity improves. \ourM{} exhibits a consistent increase in uniqueness as the model's capacity grows.  We also establish comparisons between \ourM{} and three generalist models: Llama 3 (8B), Mixtral (8x7B), and GPT-4. Our \ourM{} significantly outperforms the others in terms of uniqueness. As for validity, the results show that GPT-4 demonstrates a remarkable ability to generalize chemically valid SMILES.

\begin{table}[!htbp]
\centering
\begin{tabular}{lccccc}
\toprule 
& Validity (\%) & Uniqueness (\%) \\
\midrule
Llama 3 (8B)    & 77.9 & 35.1  \\ 
Mixtral (8x7B) & 72.6 & 35.1  \\
GPT-4          & \textbf{99.6} & 54.6  \\
\midrule
\ourM{} (1B)       & 94.9 & 91.1 \\
\ourM{} (8B)       & 96.8 & 96.6 \\
\ourM{} (8x7B)     & 98.8 & \textbf{98.8} \\
\bottomrule
\end{tabular}
\caption{Unconditional evaluation of small molecules generation. RDKit is used to convert the generated SMILES strings into molecular structures and check validity. The uniqueness ratio is calculated among the valid molecules.}
\label{tab:unconditional_mol_eval}
\end{table}

\subsection{Property-to-molecule generation}
\label{sec:basic_property_to_mol}
The task is to generate molecules with specified properties, which is a critical aspect of molecular design. An example is shown as follows:
\begin{example}
 \textbf{Instruction: }\\\texttt{Generate a molecule with four hydrogen bond donors.}\\
\textbf{Response: }\\\mol{}C(C[C@@H](C(=O)O)N)CN=C(N)N\emol{}
\end{example}  

\begin{figure}[!htbp]
    \centering
    \includegraphics[width=\linewidth]{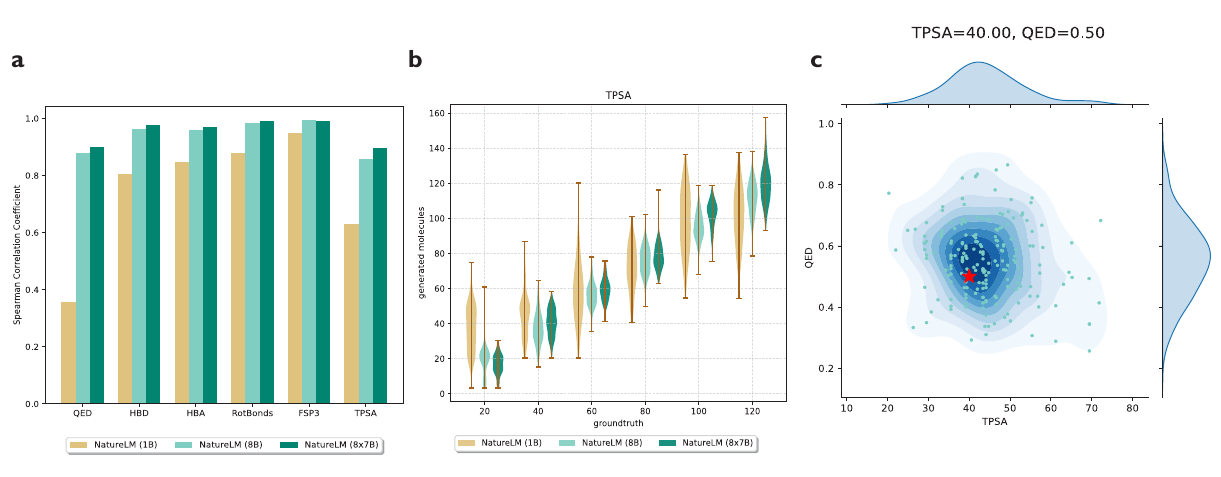}
    \caption{Evaluation of property-to-molecule generation. (a) Bar plot of the Spearman correlation coefficients between the input property values and generated molecules' property values. (b) Violin plot showing the input TPSA values and generated molecules' TPSA values. More properties are shown in Fig. \ref{fig:basic_to_cmpd_violinplot}. (c) The joint distribtion of the generated molecules' TPSA and QED given the input ``TPSA=40, QED=0.5'' (see Fig. \ref{fig:qed_fsp3_joint_optim} for more cases).}
    \label{fig:property2mol}
\end{figure}

We conduct evaluations of \ourM{} on six distinct properties: Quantitative Estimate of Drug-likeness (QED), hydrogen bond acceptors (HBA), hydrogen bond donors (HBD), fraction of sp3 hybridized carbons (FSP3), rotatable bonds (RotBonds), and topological polar surface area (TPSA). All these properties can be calculated using RDKit. For each property, we select multiple values as inputs to the model (see Table \ref{tab:property_values}). We generate 100 molecules for each input and evaluate them with metrics including the Spearman correlation (Fig. \ref{fig:property2mol}a) and the correct ratio (Fig. \ref{fig:prop2mol_correct}). 
Our findings reveal that on certain property, such as TPSA, the model demonstrates a Spearman correlation greater than 0.8, illustrating the consistency between the generated molecules and the input specifications (Fig. \ref{fig:property2mol}b).

Additionally, our model can handle the combination of multiple properties. For example, when given the command ``\texttt{Generate a compound with QED 0.5 and TPSA 40}'', the model generates compounds that meet both specified criteria. The results are shown in Fig. \ref{fig:property2mol}c. The majority of the generated compounds have QED and TPSA values centered around our desired properties (i.e., 0.5 and 40), demonstrating the versatility and effectiveness of \ourM{} in multi-property molecular generation.

\subsection{Translation between SMILES and IUPAC}\label{sec:smiles_iupac}
We evaluate NatureLM on the translation between SMILES and IUPAC on NC-I2S and NC-S2I \cite{yu2024llasmol}, the bidirectional IUPAC-SMILES translation dataset comprising 2993 pairs of SMILES and their corresponding IUPAC names (Table \ref{tab:text_to_cmpd_eval}). We ensure that there is no test set leakage in this setting. On both text-to-SMILES and SMILES-to-text translation tasks, NatureLM (8x7B) outperforms all competing language models in terms of accuracy, demonstrating our model's strong capability for text-molecule correspondence. NatureLM significantly outperforms GPT-4 and Claude 3 Opus \cite{claude3}, strong generalist large language models (LLMs), highlighting the necessity of training on scientific data. Compared with another LLM trained on text and SMILES corpus LlaSMol$_{\rm Mistral}$ \cite{yu2024llasmol}, NatureLM also obtains significantly better performance. Moreover, NatureLM (8x7B) performs comparably with STOUT \cite{rajan2024stout},  the widely-used model trained specially for IUPAC-SMILES translation task, demonstrating NatureLM's potential as a scientific generalist in specific domains. The performance increases from NatureLM (1B) to NatureLM (8x7B), exhibiting the scaling benefits of larger models. A case study is presented in Fig. \ref{fig:case_study_iupac_to_smiles}, comparing \ourM{} with general large language models and highlighting the advantages of \ourM{} in scientific tasks.

\begin{table}[!htbp]
\centering
\begin{tabular}{lcc}
\toprule
& IUPAC-to-SMILES & SMILES-to-IUPAC\\
\midrule
STOUT &\textbf{0.735}&0.565\\ 
GPT-4&0.033&0\\ 
Claude 3 Opus & 0.177 & 0\\
LlaSMol$_{\rm Mistral}$ & 0.701 & 0.290\\
\ourM{} (1B) & 0.476 & 0.284\\
\ourM{} (8B) & 0.679 & 0.517\\
\ourM{} (8x7B) & 0.704 & \textbf{0.607}\\
\bottomrule
\end{tabular}
\caption{IUPAC-SMILES translation results. Models are evaluated by top-5 accuracy.}
\label{tab:text_to_cmpd_eval}
\end{table}

\subsection{Target-aware hit compound generation and optimization}\label{sec:tamgen}
The task is to generate small molecule compounds given the target protein sequence. The combination of \ourM{} and structure-based compound design, such as TamGen \cite{TamGen} and TargetDiff \cite{targetdiff}, will be explored in the future. We test \ourM{} within two distinct scenarios: 

(1) Generate compounds from the target protein sequences. This process is crucial for the hit identification stage of drug discovery, with the goal of discovering  chemical entities that exhibit specific interactions with the target protein. 

(2) Generate molecular fragments based on the target protein sequences and partial molecular structures as inputs. This method is instrumental during the lead optimization phase, where we scrutinize and refine the molecular architecture to improve efficacy and precision.

The examples are shown below:

\begin{example}  

\noindent$\rhd${\em Scenario 1}: Complete molecule generation

\noindent\textbf{Instruction: }\\Produce a compound guided by the target \\
\noindent\pro{}LALSLTADQMVSALL...SYDLLLEMLDAH\epro{} \\
\noindent\textbf{Response:}\mol{}CC1=C(c2cccc(O)c2)C(c2ccc(I)cc2)Oc2ccc(O)cc21\emol{}


\noindent$\rhd${\em Scenario 2}: Fragment generation

\noindent\textbf{Instruction: }\\Design a compound with reference to the target \\
\pro{}DTKEQRILR$\cdots$EKAIYQGP\epro{} and the fragment  $\langle$\texttt{fragA}$\rangle$O=c1[nH]cnc2c(O)cc([*:1])c([*:2])c12$\langle$\texttt{/fragA}$\rangle$\\
        \textbf{Response: }\\$\langle$\texttt{fragB}$\rangle$Fc1ccc([*:1])cc1.Fc1ccc([*:2])cc1$\langle$\texttt{/fragB}$\rangle$
\end{example}  

\begin{figure}[!htbp]
\centering
\includegraphics[trim=1cm 0 1cm 0, clip, width=\linewidth]{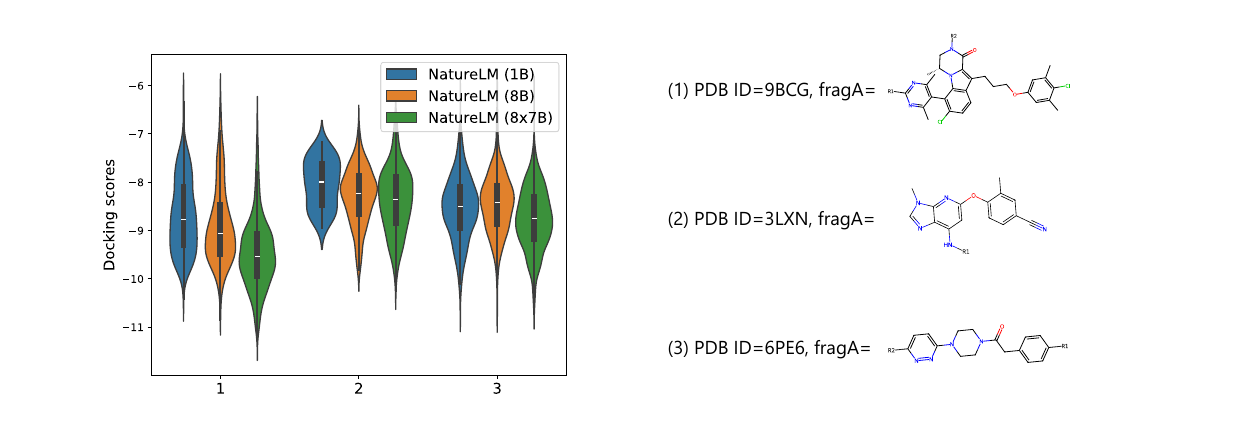}
\caption{Docking scores for molecules in target-to-fragment generation.  This violin plot presents the docking scores of molecules involved in target-to-fragment generation. We selected three recent papers that focus on fragment optimization: \cite{Tarr2024pdb9BCG}, \cite{Mammoliti2024pdb3xln} and \cite{Tangallapally2024pdb6PE6}, which utilize PDB IDs 9BCG, 3LXN, and 6PE6, respectively. The input fragment is visualized alongside its corresponding PDB ID for clarity.}
\label{fig:target_to_frag}
\end{figure}

Here, ``[*:digit]'' refers to the connection point of the molecular fragment, like the R1 and R2 in Fig. \ref{fig:target_to_frag}.

In the first scenario, we compare \ourM{} with a sequence generation method, TamGen \cite{TamGen}, and two other approaches that design compounds in 3D space based on the input target structure: a diffusion-based method, TargetDiff \cite{targetdiff}, and an autoregressive generation method in 3D space, Pocket2Mol \cite{pocket2mol}. We follow the evaluation procedure outlined in the TamGen paper \cite{TamGen}, which includes calculating the docking score using AutoDock Vina \cite{autodock_vina_1,autodock_vina_2}, as well as assessing the QED, synthetic accessibility scores (SAS), diversity of the generated compounds, the percentage of compounds with logP in the range [0,5], and the percentage of compounds satisfying the rule-of-five. The results are presented in Table \ref{tab:tamgen_results}. We can see that in terms of docking score, QED and synthesis ability, \ourM{} surpasses previous baselines, highlighting its effectiveness. For the other metrics, \ourM{} achieves comparable performance to the baselines.


\begin{table}[!htbp]
\centering
\begin{tabular}{lccccccc}
\toprule
& Vina ($\downarrow$) & QED & SAS & Diversity & LogP$\in[0,5]$ & Ro5  \\
\midrule
Pocket2Mol    & -4.90 & 0.52 & \textbf{0.84} & \textbf{0.87} & 0.76 & \textbf{1}     \\
TargetDiff    & -6.08 & 0.55 & 0.67 & 0.83 & 0.74 & 0.98  \\
TamGen        & -6.66 & 0.56 & 0.76 & 0.75 & 0.84 & 0.99  \\
\ourM{} (1B)  & -6.80 & \textbf{0.64} & 0.82 & 0.77 & \textbf{0.85} & 0.99  \\
\ourM{} (8B)  & -6.92 & 0.62 & 0.81 & 0.73 & 0.84 & 0.99  \\
\ourM{} (8x7B)& \textbf{-6.95} & 0.62 & 0.82 & 0.75 & 0.84 & 0.99 \\
\bottomrule
\end{tabular}
\caption{Statistics of target to complete compound sequence generation.}
\label{tab:tamgen_results}
\end{table}

Additionally, we utilize \ourM{} for fragment generation. We selected three papers published after May 2024 \cite{Tangallapally2024pdb6PE6,Tarr2024pdb9BCG,Mammoliti2024pdb3xln}, where part of their task is to solve the issue of compound optimization. From these papers, we choose the target proteins and the fragments to retain, allowing \ourM{} to generate the remaining fragments.
The results are illustrated in Fig. \ref{fig:target_to_frag}. In this instance, it is evident that larger \ourM{} models  yield superior docking scores in general.

\subsection{Text-guided binding affinity optimization}\label{sec:binding_affinity}
To further improve the binding affinity between a target and a molecule, we propose a text-guided binding affinity optimization task. Given a target name and a molecule with a known binding affinity for that target, we aim to generate molecules with higher binding affinity, which is a crucial component for lead optimization. An example is shown below:
\begin{example}
    \noindent\textbf{Instruction: }\\\texttt{Improve the binding affinity on Uridine-cytidine kinase 2 of }\mol{}Cc1ccc(-c2nc3c(c(SCC(=O)Nc4ccccc4)n2)Cc2cccc(C)c2O3)cc1\emol{}\\
    \textbf{Response: }\\\mol{}Cc1ccc(-c2nc3c(c(SCC(=O)Nc4cccc(C(=O)O)c4)n2)Cc2cccc(C)
    c2O3)cc1\emol{}
\end{example} 
Here, the target information is provided in text format, which complements the FASTA representation used in Section \ref{sec:tamgen}. We will combine text and FASTA  in the future.

We test \ourM{} on 12 targets that are not present in the post-training data and use a hybrid retrieval and docking approach for evaluation. Specifically, for the generated molecules, if we can retrieve their binding affinity values from the ChEMBL database, we compare these values with the original molecule's binding affinity. Otherwise, we resort to docking scores as the comparison metric. For the 12 selected targets, their Spearman correlation between the docking score and the actual binding affinity for known molecules exceeds 0.5, indicating the reliability of using docking for assessment (Table \ref{tab:targets}).

\begin{figure}
    \centering
    \includegraphics[width=\linewidth]{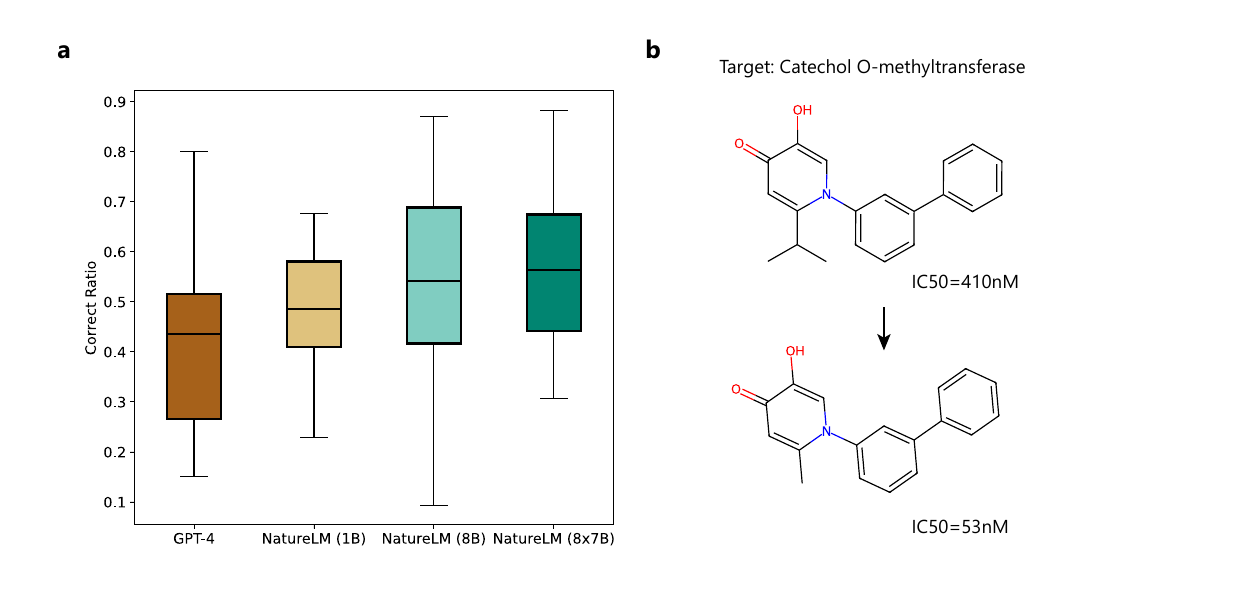}
    \caption{Evaluation of text-guided binding affinity optimization. (a) Box plot of the correct ratios of GPT-4, NatureLM (1B), NatureLM (8B) and NatureLM (8x7B) on 12 targets. (b) A case on the binding affinity optimization for Catechol O-methyltransferase. With small modifications, the binding affinity improves from 410nM to 53nM.}
    \label{fig:affnity}
\end{figure}

We observe that \ourM{} can successfully improve the molecule's binding affinity by making modifications on its chemical components, in a manner similar to what a chemist would typically do (Fig. \ref{fig:affnity}b). Compared with GPT-4, \ourM{} can generate  molecules (Fig. \ref{fig:affnity}a) with higher binding affinity, making it a better tool for molecule optimization than general purpose LLM. Another observation is that more than 90\% molecules generated by \ourM{} do not have known binding affinity score in ChEMBL database. For 8 out of the 12 targets, over 50\% of the generated novel molecules successfully improved the docking scores (Fig. \ref{fig:binding_docking}),  demonstrating the model's potential in exploring chemical spaces and discovering novel drugs. We observe that \ourM{} (8x7B) and \ourM{} (8B) outperform \ourM{} (1B) as they generate more correct molecules for the majority of targets (Fig. \ref{fig:binding_correct}).

\subsection{Text-guided metabolism \& distributional property optimization}\label{sec:admet}

We next assess the molecular optimization capabilities of \ourM{} regarding metabolism and its potential to cross the blood-brain barrier (BBB), a crucial factor in drug distribution. For metabolism, the goal is to reduce the inhibition of five key metabolic enzymes: CYP1A2, CYP2C19, CYP2C9, CYP2D6 and CYP3A4. An example is shown below:

\begin{example} 
{{
\small
\textbf{Instruction: }\\\texttt{Transform} \mol{}COCCn1c(=O)c(-c2ccccc2)nc2cnc(N3CCOCC3)nc21\emol{} \texttt{to yield lower CYP1A2 inhibition and present a better metabolic safety record.}\\
    \textbf{Response: }\\\mol{}COc1ccc(-c2nc3cnc(N4CCOCC4)nc3n(Cc3cccc(OC)c3)c2=O)cc1\emol{}
}}
\end{example}  

In terms of BBB permeability (BBBP), we evaluate the enhancement BBB permeability. An example is provided below:

\begin{example}
{{
\small
\noindent\textbf{Instruction:}\\ \texttt{Adjust the molecule} \mol{}CC[C@H](NC(=O)c1c(OCCCC(=O)O)c

\noindent(-c2ccccc2)nc2ccccc12)c1ccccc1\emol{} \texttt{to facilitate its passage through the blood-brain barrier.}
\newline
\noindent\textbf{Response:} \\\mol{}CC[C@H](NC(=O)c1c(O)c(-c2ccccc2)nc2ccccc12)c1ccccc1\emol{}
}}
\end{example}

For each test sample, we used random search to generate four cases. To determine whether NatureLM effectively refined the input molecule, we trained six groups of deep learning models for this evaluation. For assessing BBBP, we utilized the state-of-the-art model, BioT5 \cite{PeiQizhi2023BioT5}, to determine whether a compound is capable of crossing the BBB. For metabolism optimization, we used ChemProp \cite{yang2019analyzing} to train classifiers to test if a molecule has the ability to inhibit enzymes from the cytochrome P450 (CYP) superfamily. We evaluated the percentage of molecules that were successfully optimized according to the specified criteria (see Section \ref{app:more_eval_method} for details). 

Table \ref{tab:cyp_optimization} displays the outcomes of BBBP and metabolism optimization. The success rates for optimizing BBBP with the 1B, 8B, and 8x7B versions of NatureLM are 0.482, 0.549, and 0.552, respectively. Larger models show better performance, suggesting potential for enhancement opportunities in the future. For metabolism optimization, generally, the 8B model outperforms the others in terms of success rate, followed by the 8x7B model and lastly the 1B model. The 1B and 8B models share the same architecture (dense models, large vocabulary size), whereas the 8x7B model has a distinct one (mixture-of-expert model, relative small vocabulary size). In this particular task, the progression from the 1B model to the 8B model is consistent. However, a detailed analysis contrasting the 8x7B model is to be conducted in subsequent studies. Additionally, we jointly optimized metabolism and a basic property. The findings indicate that larger models generally yield better results (see Table \ref{tab:joint_basic_cyp}).

\begin{table}[!htpb]
\centering
\resizebox{\columnwidth}{!}{
\begin{tabular}{lccccccc}
\toprule
 & BBBP & CYP1A2 & CYP2C19 & CYP2C9 & CYP2D6 & CYP3A4 & CYP Average \\
\midrule
1B  & 0.482 & 0.805 & 0.815 & 0.770 & 0.750 & 0.831 & 0.794\\
8B & 0.549 & \textbf{0.882} & 0.813 & \textbf{0.882} & \textbf{0.833} & \textbf{0.913} & \textbf{0.865}\\
8x7B & \textbf{0.552} &  0.837 & \textbf{0.834} & 0.838 & 0.812 & 0.853 & 0.835\\
\bottomrule
\end{tabular}
}
\caption{Optimization results of BBBP metabolism and CYP enzymes. Measured by success rate. }
\label{tab:cyp_optimization}
\end{table}

\subsection{Retrosynthesis prediction}\label{sec:retro}
Retrosynthesis aims to identify synthesis routes for target molecules using commercially available compounds  as starting points, a critical task in the discovery and manufacture of functional small molecules~\cite{corey1969computer, segler2018planning, maziarz2024chimera}.
The applicability of ML-based retrosynthesis tools largely depends on the accuracy of single-step retrosynthesis prediction.
We evaluate the capability of \ourM{} for single-step retrosynthesis prediction on USPTO-50K ~\cite{schneider2016uspto50k}.
\ourM{} is prompted with the task description and the chemical SMILES of the product molecule, and is expected to generate potential reactants. 

We followed the common practice for splitting the USPTO-50K dataset~\cite{dai2019retrosynthesis, maziarz2024re}, and evaluated the performance using the 5007 reactions included in the test set.
We ensured that there is no test set leakage in this setting.
As outlined in Table~\ref{tab:retro-uspto50k},
all sizes of \ourM{} models surpass other methods in terms of top-$k$ accuracy, demonstrating our
model’s accurate predictive ability for retrosynthesis prediction.
\ourM{} significantly outperforms GPT-4, suggesting that training on scientific data is crucial for models to excel in scientific tasks. 
Furthermore, \ourM{} outperforms the state-of-the-art domain-specific models such as LocalRetro~\cite{chen2021localretro} and R-SMILES~\cite{Zhong2022rsmiles}, showing NatureLM’s potential as a scientific generalist in critical scientific tasks. We also note an increase in performance from \ourM{} (1B) to \ourM{} (8x7B), demonstrating the scaling advantages of larger models.

\begin{example}
{{
\small
        \textbf{Instruction: }\\
        \texttt{Please suggest possible reactants for the given product}
        $\langle$\texttt{product}$\rangle$CC(=O)c1ccc2c(ccn2C(=O)OC(C)(C)C)c1$\langle$\texttt{/product}$\rangle$ \\
        \textbf{Response: }\\ 
        $\langle$\texttt{reactant}$\rangle$
        CC(=O)c1ccc2[nH]ccc2c1.CC(C)(C)OC(=O)OC(=O)OC(C)(C)C
        $\langle$\texttt{/reactant}$\rangle$
}}
\end{example}



\begin{table}[!h]
\centering
\begin{tabular}{lcc}
\toprule
& Top-1 accuracy & Top-3 accuracy \\
\midrule
GPT-4 & 22.4\% & N/A \\
LocalRetro~\cite{chen2021localretro} & 51.5\% & 76.5\% \\ 
R-SMILES~\cite{Zhong2022rsmiles} & 56.0\%  & 79.1\% \\
EditRetro~\cite{han2024editretro} & 60.8\% & 80.6\% \\ 
\midrule
\ourM{} (1B) & 68.6\% & 86.8\%\\
\ourM{} (8B) & 70.2\% & 85.9\% \\
\ourM{} (8x7B) & 71.9\% & 87.4\% \\
\bottomrule
\end{tabular}
\caption{Retrosynthesis prediction results on USPTO-50K dataset.}
\label{tab:retro-uspto50k}
\end{table}

\begin{figure}[!htbp]
\centering
\includegraphics[width=0.8\linewidth]
{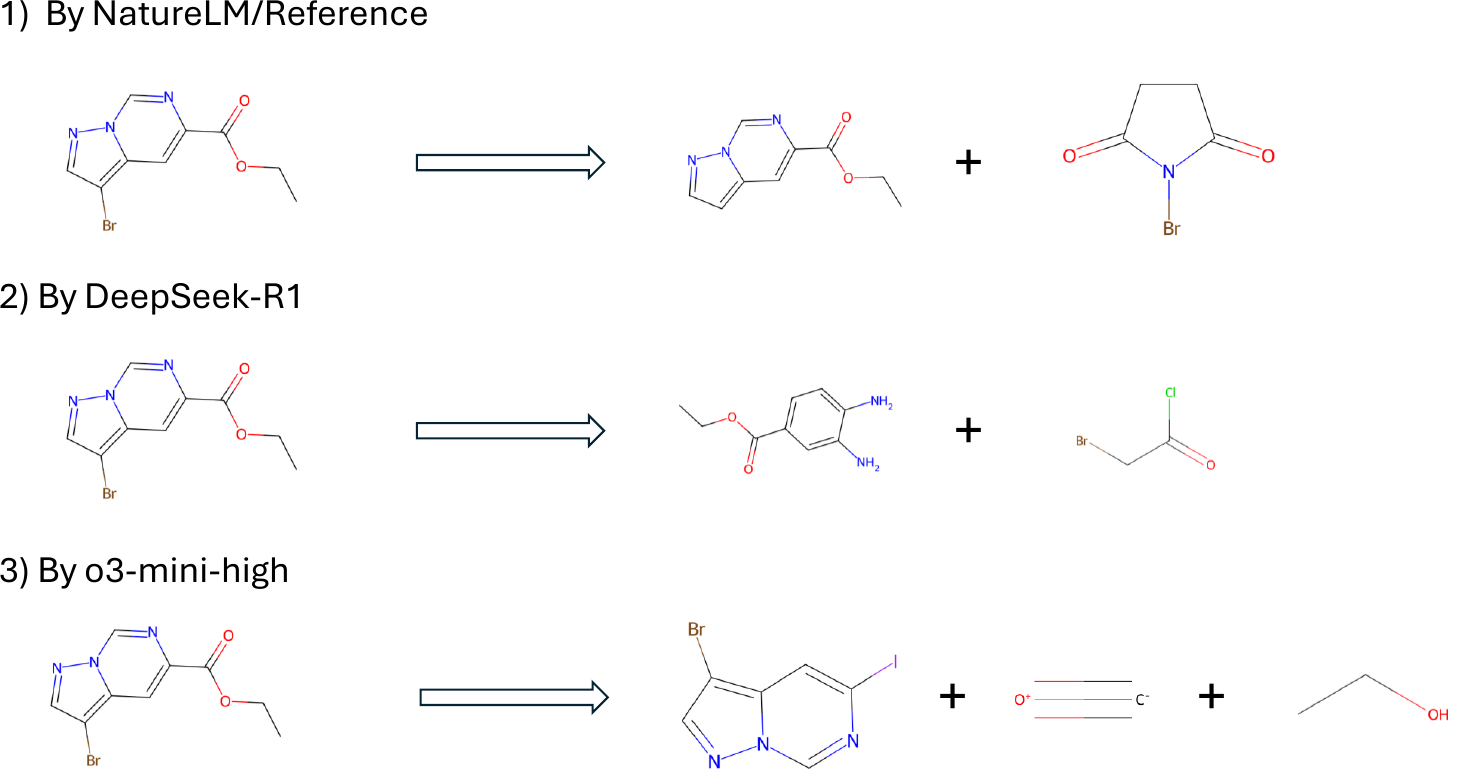}
\caption{Case study on retrosynthesis prediction. We evaluated the performance of \ourM{}, DeepSeek-R1, and o3-mini-high for retrosynthesis prediction using a reaction from U.S. Patent ID US12018024B2 (not included in our training set), granted to Novartis on June 25, 2024. \ourM{} successfully proposed the ground-truth reactants from the patent, while the outputs from the other two methods required further refinement to achieve the same level of accuracy.
}
\label{fig:case_study_reaction}
\end{figure}

We selected a newly discovered reaction from a U.S. patent with ID US12018024B2 (granted to Novartis on June 25, 2024) as the case study. As shown in Figure~\ref{fig:case_study_reaction}, the product is a brominated heterocyclic compound. To synthesize such molecules, organic chemists typically begin by constructing the ring system, then followed by a halogenation reaction on the ring. The halogenated site is subsequently used in further syntheses, often coupling with other ring systems. In this context, our NatureLM model accurately predicted one of the most common brominating agent NBS (\textit{N}-Bromosuccinimide) in this step, aligning perfectly with experimental results. DFT optimization calculations to reactant CCOC(=O)c1cc2ccnn2cn1 show that 3-position is the most nucleophilic. So bromination prefers at 3-position. This demonstrates NatureLM’s capability to effectively predict useful reactants in chemical reactions.

In contrast, DeepSeek-R1 \cite{deepseekai2025r1} model selected incorrect reactants. DeepSeek-R1 focused on constructing a heterocyclic ring system in this step. However, it incorrectly positioned the nitrogen and carbon atoms in the substrate, leading to the wrong outcome. Meanwhile, o3-mini-high model selected another correct route but not so convenient for the whole synthetic process. o3-mini-high might attempt to construct the acetyl side chain by transition-metal-catalyzed CO insertion reaction. Although radical mechanism can achieve this since iodo atom is more reactive than bromo atom, introducing the side chain at this stage may still lead to a side reaction at the bromo atom. In the actual synthesis, this acetyl side chain is generally constructed in previous steps. These cases illustrate that, so far, general LLMs cannot fully grasp the rationale behind chemical synthesis. Although they can generate some reasons and predictions, these do not align with practical strategies.

Another example is shown in Fig. \ref{fig:case_study_reaction2}, where \ourM{} accurately predicts the reactants, while the outputs from DeepSeek-R1 and o3-mini-high require additional refinement.

%% file: chapters/protein.tex
\section{Protein tasks}
\label{sec:protein}

Our model's capabilities with respect to proteins are assessed through several distinct types of tasks:

\begin{enumerate}
\item Unconditioned protein generation: The model generates protein sequences from scratch without any specific conditions or prompts.
\item Text-guided protein generation: This task involves guiding the model to generate protein sequences based on given natural language descriptions.
\item Antibody design: The model designs the Complementary-Determining Region H3 (CDR-H3) of antibodies to effectively bind to target antigens.
\item Protein characteristics description generation: This task is to generate clear, human-readable explanations of protein sequences, describing their properties and functions.
\item Case study: The model designs heme-binding protein design driven by text and SMILES.
\end{enumerate}

\subsection{Unconditioned generation}\label{sec_prot_generation}

The first capability of the model is generating protein sequences from scratch freely, prompted by the start token for proteins only, i.e., \text{$\langle$protein$\rangle$}. 
However, since there is no golden standard for evaluating proteins when no conditions are specified, it is difficult to assess the generation results. We focus on foldability, measured by pLDDT score \cite{Mariani2013-av}, as well as lengths and diversity of the sequences, for the valid sequences.

\begin{table}[!h]
\centering
\begin{tabular}{lccc}
\toprule
Model & Avg Length & Diversity & AVG pLDDT \\
\midrule
Mixtral 8x7b & 53 & 0.906 & 69.9 \\
GPT-4 & 46 & 0.816 & 65.1 \\
\midrule
\ourM{} (1B) & 288 & 0.985 & 69.8 \\
\ourM{} (8B) & 285 & 0.973 & 71.8 \\
\ourM{} (8x7B) & 318 & 0.989 & 75.9 \\
\bottomrule
\end{tabular}
\caption{Protein Sequence Generation Comparison. The average length of natural proteins (calculated from a subset of proteins randomly sampled from UR50) is about 311. The diversity was calculated by the number of clusters with 50\% sequence identity divided by the total generated sequence count. The pLDDT scores were calculated by OmegaFold \cite{omegafold} on the generated sequences with length less than 100 for a fair comparison. The length distribution is shown in Figure \ref{fig:protein:unconditioned_generation_sequence_length}. }
\label{tab:protein:unconditioned_generation}
\end{table}

As shown in Table \ref{tab:protein:unconditioned_generation}, \ourM{} consistently outperform Mixtral 8x7b and GPT-4 in terms of average sequence length, diversity, and average pLDDT score. The \ourM{} (8x7B) model achieves the best performance across all metrics, with an average length of 318, diversity of 0.989, and average pLDDT score of 75.9. ProLLAMA~\cite{lv2024prollamaproteinlanguagemodel}, a fine-tuned LLM for protein, generates proteins without explicitly defined constraints on length, achieving a pLDDT score of 66.5. In contrast, our approach, which does not impose length constraints, results in pLDDT scores of 69.8 and 78.1 for the 8B and 8x7B models, respectively, demonstrating our significant advancement in protein sequence generation.

\subsection{Text-guided protein generation}\label{sec:text_guided_protein_design}


For text-guided protein generation, we evaluated our models' ability to generate proteins with specific properties based on natural language prompts. In this study, we focused on two key properties: solubility and stability, leaving the exploration of additional properties for future work.
For stability, the models were tasked with generating protein sequences that exhibit stable properties. Regarding solubility, since both soluble and insoluble proteins exist in nature, we instructed \ourM{} to generate sequences of both types. 
Sample prompts are shown below, and a full list of prompts can be found in Figure~\ref{fig:protein:conditioned_prompts_full}.

\begin{example} 
$\rhd$ An example prompt for ``stable protein generation''\\
\texttt{I require a stable protein sequence, kindly generate one.}\\
$\rhd$ An example prompt for ``soluble protein generation''\\
\texttt{Generate a soluble protein sequence.}\\
$\rhd$ An example prompt for ``insoluble protein generation''\\
\texttt{Produce a protein sequence that is not soluble.}
\end{example} 

To evaluate the stability and solubility of a generated protein sequence, we utilized two specialist models fine-tuned from the protein foundation model, SFM-Protein~\cite{he2024sfm}, as oracle models. One model was used for stability classification, while the other was used for solubility classification. 
The oracle models provide probabilities that suggest the likelihood of the sequence possessing the desired property. To verify the efficiency of our model against random sampling, we have also chosen a subset of 1000 natural protein sequences from the UR50 dataset and assessed them using the same oracle models.


\begin{figure}[!htbp]
\centering
\subfigure[\ourM{} (1B)]{
\includegraphics[width=0.33\linewidth]{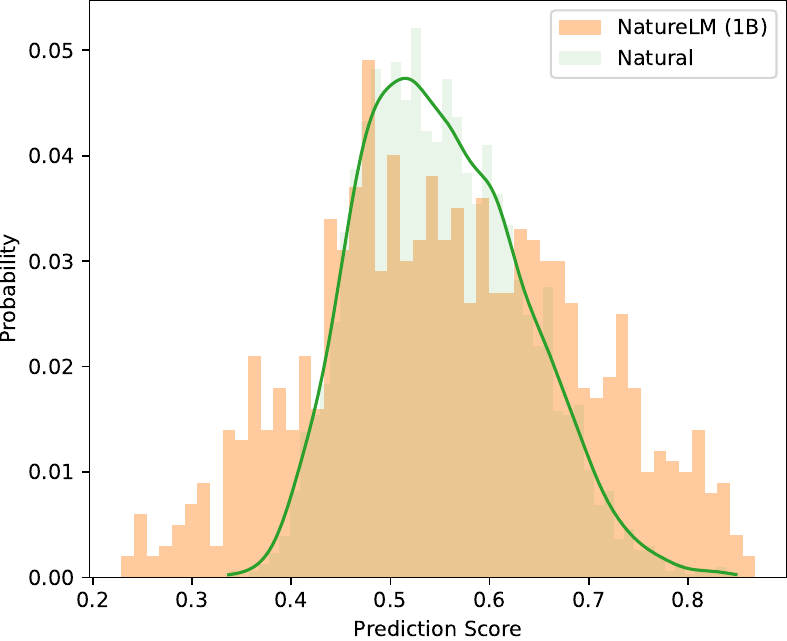}
}%
\subfigure[\ourM{} (8B)]{
\includegraphics[width=0.33\linewidth]{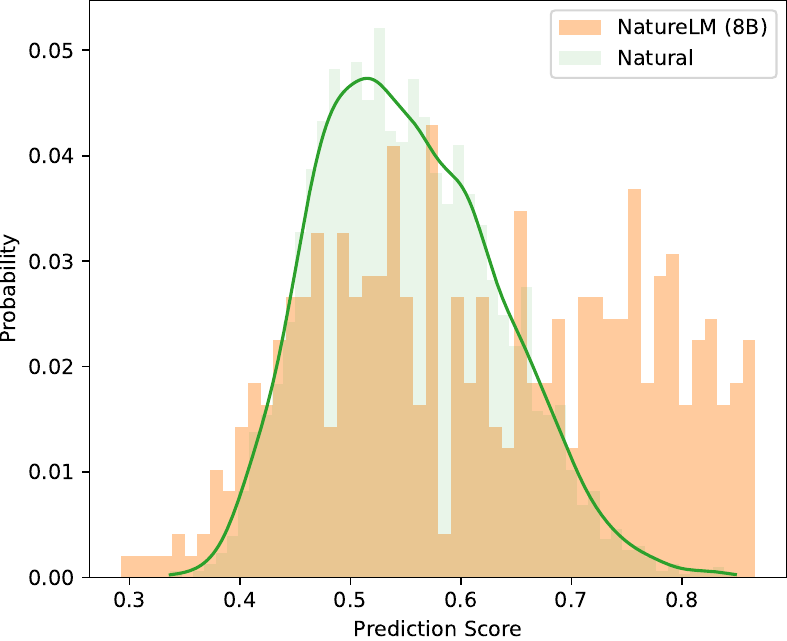}
}%
\subfigure[\ourM{} (8x7B)]{
\includegraphics[width=0.33\linewidth]{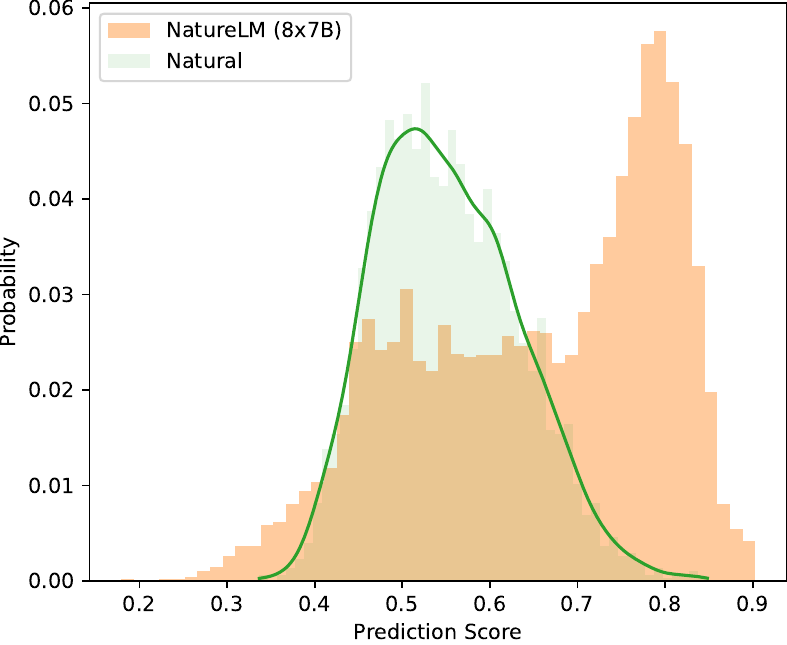}
}
\caption{Stability score distribution of the generated sequences.}
\label{fig:protein:conditioned_generation_stability}
\end{figure}

\begin{table}[!h]
\centering
\begin{tabular}{ccc}
\toprule
Source & AVG Prediction & Data Ratio (Score $>0.5$) \\
\midrule
Natural & 0.552 & 0.704 \\
\ourM{} (1B) & 0.559 & 0.644 \\
\ourM{} (8B) & 0.619 & 0.757 \\
\ourM{} (8x7B) & 0.655 & 0.812 \\
\bottomrule
\end{tabular}
\caption{Stability score ratio of the generated sequences.}
\label{tab:protein:conditioned_generation_stability}
\end{table}

\begin{figure}[!htbp]
\centering
\subfigure[\ourM{} (1B)]{
\includegraphics[width=0.33\linewidth]{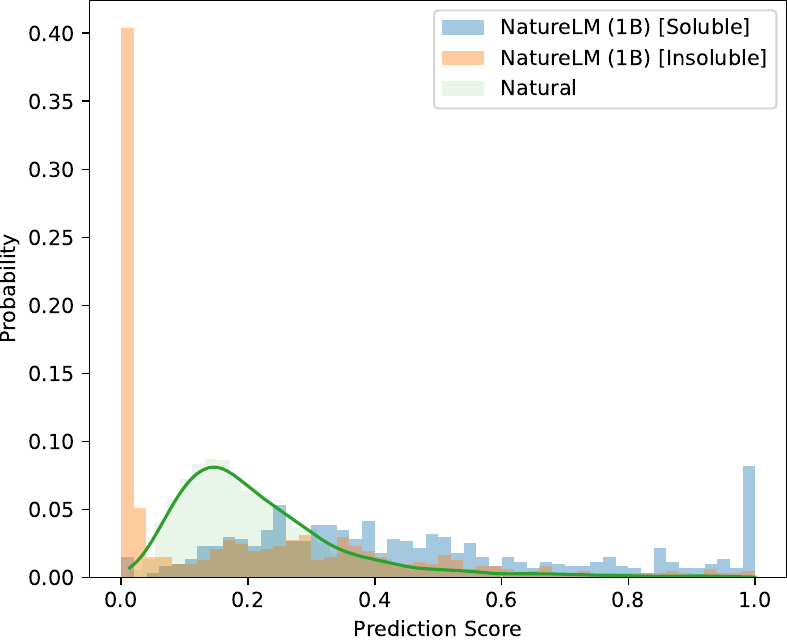}
}%
\subfigure[\ourM{} (8B)]{
\includegraphics[width=0.33\linewidth]{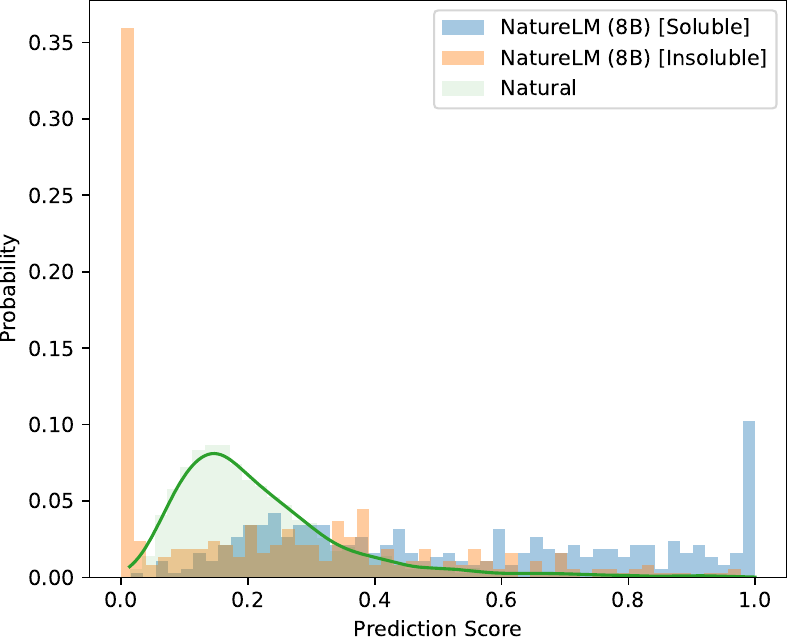}
}%
\subfigure[\ourM{} (8x7B)]{
\includegraphics[width=0.33\linewidth]{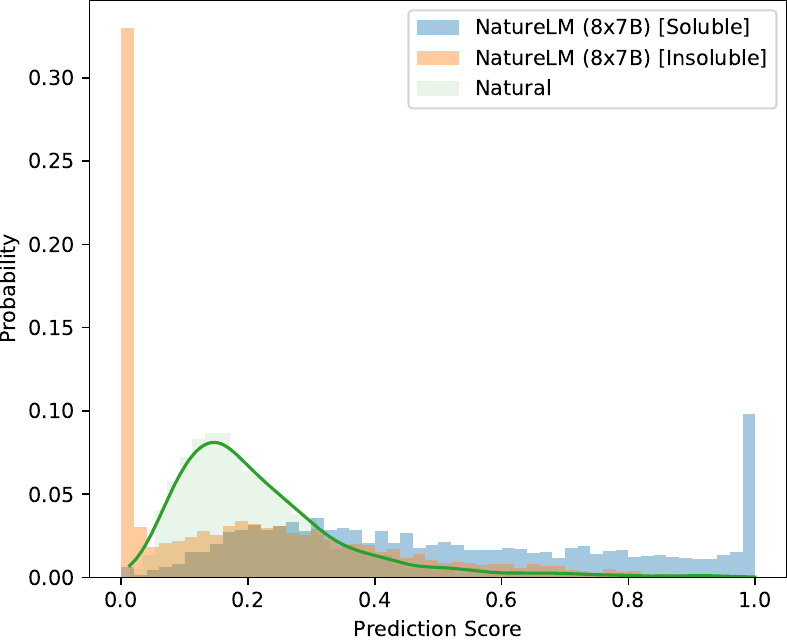}
}
\caption{Solubility score distribution of the generated sequences.}
\label{fig:protein:conditioned_generation_solubility}
\end{figure}


Figures \ref{fig:protein:conditioned_generation_stability} and \ref{fig:protein:conditioned_generation_solubility} show the distributions of stability and solubility scores for the generated sequences, respectively. The \ourM{} models demonstrate controlled distribution shift in generating proteins with desired properties compared to the natural sequences. 
In the task of generating more stable proteins, as shown in Figure~\ref{fig:protein:conditioned_generation_stability}, a clear trend emerges: as the model size increases, the proportion of sequences classified as stable grows, with a pronounced peak in the \ourM{} (8x7B) results. The quantified data, summarized in Table~\ref{tab:protein:conditioned_generation_stability}, further supports this observation. All three models produce proteins that are more stable than natural sequences based on average stability scores. Additionally, two of the models outperform natural proteins in terms of the number of sequences that exceed a stability threshold of 0.5.
For the solubility condition, Figure~\ref{fig:protein:conditioned_generation_solubility} reveals a similar trend. As the model size increases, the separation between the distributions of soluble and insoluble scores becomes more distinct, with less overlap. 

\subsection{Antigen-binding CDR-H3 design}

The task of antigen-binding CDR-H3 design focuses on constructing the Complementary-Determining Region H3 (CDR-H3) of an antibody to bind effectively to a target antigen. We employed the RAbD benchmark dataset~\cite{adolf2018rabd}, comprising 60 antibody-antigen complexes. The example is shown below:

\begin{mdframed}
\noindent\textbf{Instruction: }\\
\texttt{Using antigen} \pro{}TQVCTGTDMKLR$\cdots$GESSEDCQS\epro{} \texttt{and antibody frameworks} \ant{}IVLTQTPS$\cdots$LAVYYC\eant{} \texttt{and} \ant{}FGGGTRLEIEVQ\eant{}, \texttt{create the CDR3 regions.}\\
\textbf{Response: }\\ 
\ant{}QQYSNYPWT\eant{}
\end{mdframed}

The generation quality is evaluated by the Amino Acid Recovery (AAR) scores for the CDR-H3 design task. We use $r$ and $\hat{r}$ to represent the reference and generated sequences respectively, while $L(r)$ and $L(\hat{r})$ denote the number of amino acids in $r$ and $\hat{r}$. The $i$-th residue in the two sequences is denoted by $r_i$ and $\hat{r}_i$. The AAR is defined as follows:
\begin{equation}
{\rm AAR}(r,\hat{r}) = \frac{1}{L(r)}\sum_{i=1}^{L(r)}\mathbb{I}(r_i = \hat{r}_i). 
\end{equation}
In case $L(\hat{r})>L(r)$, only the first $L(r)$ elements are verified. If $L(\hat{r})<L(r)$, we assign $\mathbb{I}(r_i = \hat{r}_i)=0$ for $i>L(\hat{r})$.

\begin{table}[!h]
\centering
\begin{tabular}{lc}
\toprule
Method & AAR ($\uparrow$) \\
\midrule
GPT-4 & 0.312 \\
RefineGNN~\cite{jin2021refinegnn} & 0.298 \\
HSRN~\cite{jin2022hsrn} & 0.327 \\
MEAN~\cite{kong2022mean} & 0.368 \\
ABGNN~\cite{gao2023abgnn} & 0.396 \\
\midrule
Llama 3 (8B) & 0.275 \\
\ourM{} (1B) & 0.273 \\
\ourM{} (8B) & 0.368 \\
\ourM{} (8x7B) & 0.376 \\
\bottomrule
\end{tabular}
\caption{AAR of the CDR-H3 design. Please note that the \ourM{} models utilize sequence-only input for this task, whereas the baseline models may incorporate additional information, such as structural data.}
\label{tab:protein:antibody:cdr3GenGiven_antigen}
\end{table}

Table \ref{tab:protein:antibody:cdr3GenGiven_antigen} presents the Amino Acid Recovery (AAR) scores for the CDR-H3 design task. As the model size of \ourM{} increases, the AAR gradually increases. The \ourM{} (8x7B) model achieves competitive performance with an AAR of 0.376, outperforming several specialized GNN-based models, except ABGNN~\cite{gao2023abgnn} in our study. While SFM-protein, a BERT-like model trained on protein sequences, holds the top performance, our results demonstrate the potential of \ourM{} in CDR-H3 design, particularly as the model scales and undergoes further refinement.

\subsection{Protein characteristics description generation}\label{sec:protein_to_desc}
Despite the rapid discovery of natural protein sequences facilitated by advanced sequencing techniques, the functions of many of these proteins remain largely unknown. This knowledge gap restricts our ability to exploit these proteins for engineering and therapeutic purposes. In this study, we explored the annotation generation capabilities of the \ourM{} series.

\begin{figure}[!htbp]
\centering
\includegraphics[trim=3cm 2cm 7cm 1cm, clip, width=0.85\linewidth]{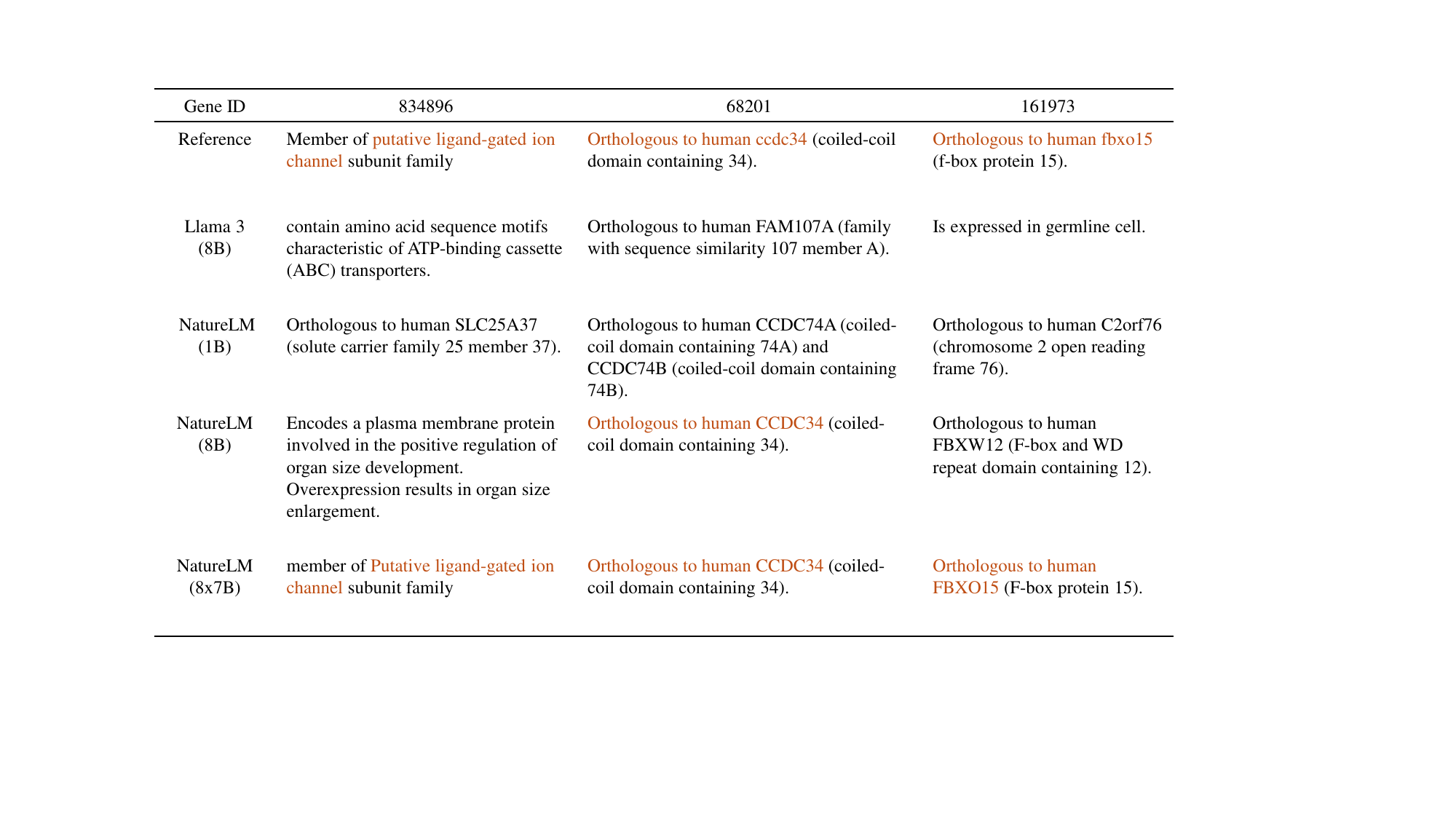}
\caption{Generated descriptions for different proteins.}
\label{fig:protein:protein_understanding}
\end{figure}

To achieve this, we compiled pairs of protein sequences and their human-readable annotations from various species, sourced from the NCBI database. We divided the dataset temporally.
Model performance was evaluated using Rouge-L scores. As shown in Table \ref{tab:protein:ncbi_description}, \ourM{} models consistently outperformed Llama 3 8B in Rouge-L scores, with performance differences widening as model size increased. Notably, the \ourM{} (8x7B) model achieved the highest score of 0.585. A detailed analysis presented in Figure \ref{fig:protein:protein_understanding} revealed that the \ourM{} (8x7B) model not only generates protein annotations with higher accuracy but also successfully identifies orthologues and functions of proteins, while \ourM{} (8B) is also able to generate reasonable results in many cases.

\begin{table}[!htbp]
\centering
\begin{tabular}{lc}
\toprule
Model Setting & Rouge-L\\
\midrule
GPT-4o & 0.091 \\
Fine-tuned Llama 3 (8B) & 0.324 \\
\ourM{} (1B) & 0.548 \\
\ourM{} (8B) & 0.572 \\
\ourM{} (8x7B) & 0.585 \\
\bottomrule
\end{tabular}
\caption{Performance of protein description generation, measured by Rouge-L. Llama 3 (8B) serves as a baseline, which is directly fine-tuned on the data collection described in Section \ref{sec:supervised_ft_data}. More details about this baseline in Section \ref{sec:ablation_study}.}
\label{tab:protein:ncbi_description}
\end{table}

In conclusion, \ourM{} demonstrates strong performance across a wide range of protein-related tasks, from unconditioned generation to specific design tasks like CDR-H3 design. The scalability of our approach is evident, with larger models consistently outperforming smaller versions and often achieving state-of-the-art results.

\subsection{Heme-binding protein design driven by text and SMILES}\label{sec:heme_case_study}
\ourM{} is capable of designing proteins to meet various requirements. In this case study, we demonstrate how to use \ourM{} to design a protein that binds to heme cofactor.

Heme is an essential molecule, playing critical roles in oxygen transport and various enzymatic processes. Designing a protein that can bind to heme can contribute to advancements in therapeutic development, biochemical research, and synthetic biology.

The protein design practice is outlined as the following 4 steps:

\noindent{\textbf{Step 1: protein generation}}. We prompt \ourM{} to generate proteins using two approaches:
\begin{enumerate}
\item Text-based protein design, where we provide a text instruction (here in English);
\item Small-molecule-based protein design, where we use the SMILES representation of heme as part of the input instruction.
\end{enumerate}

Examples of these prompts and their corresponding responses are shown below:

\begin{example}
{{
\small
\noindent$\rhd$ {Generate proteins based on text} (Section \ref{sec:text_guided_protein_design})

\noindent\textbf{Instruction:} \texttt{Design a heme-binding protein sequence.} 

\noindent\textbf{Response:} \pro{}MSAAEGAVVFSEEKEALVLK$\cdots$\epro{}

\noindent$\rhd$ { Generate proteins based on the SMILES of the heme}

\noindent\textbf{Instruction:} \texttt{Generate a protein that can bind to} \\
\mol{}[Fe+2].C=CC1=C(C)c2cc3[n-]c(cc4[n-]c(cc5nc(cc1n2)C(C)$\cdots$\emol{}

\noindent\textbf{Response:} \pro{}ETIGKRVFVHYCHGCHSQNALGI$\cdots$\epro{}
}}
\end{example}

\noindent{\textbf{Step 2: description generation}}. For each generated protein, we utilize the protein-to-text functionality in \ourM{} (Section \ref{sec:protein_to_desc}), to obtain a description of the properties and potential functions of the generated protein.

\noindent{\textbf{Step 3: screen proteins through keyword matching}}. In this step, we use GPT-4o to generate a keyword list, called \texttt{HemeList}, containing characteristics associated with heme-binding proteins. For every protein description generated in Step 2, we check whether it contains keywords from \texttt{HemeList}. If a description matches these criteria, the corresponding protein is added to a list called \texttt{HemeProtein}.

\noindent{\textbf{Step 4: structure prediction and validation}}. For each protein in \texttt{HemeProtein}, we use Protenix~\cite{Protenix2025} to predict the complex structure of the generated proteins bound to heme. The predicted structures are then inspected to ensure that the proteins can form the critical interaction with heme for stable binding.

\begin{figure}[!htpb]
    \centering
    \includegraphics[width=0.9\linewidth]{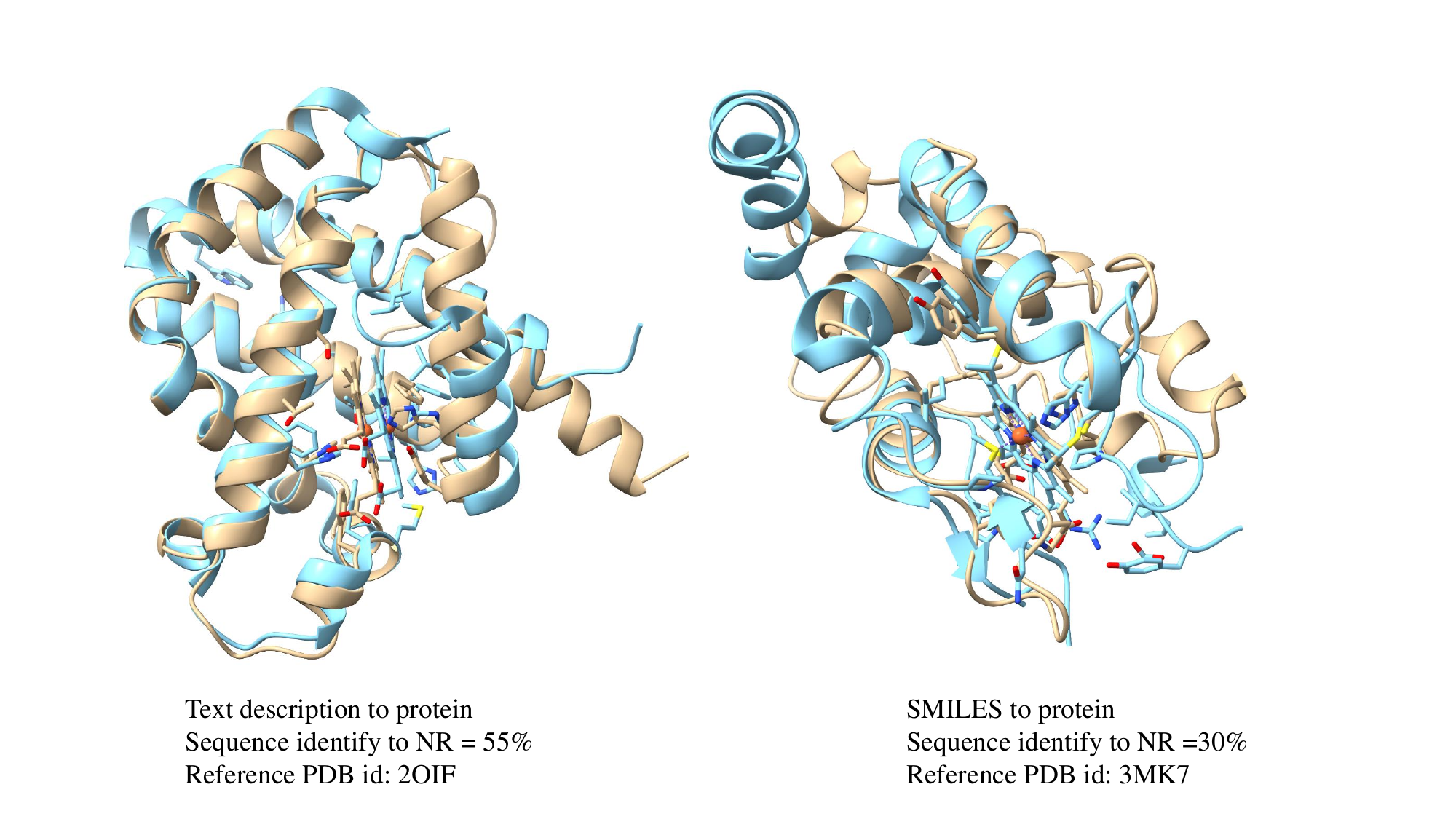}
    \caption{Two examples of proteins with plausibility of binding to heme. The yellow models represent the generated protein structures, while the blue models correspond to the reference structures retrieved using the built-in ``blast protein'' function in ChimeraX \cite{chimerax2023}. In each model, the heme binding region is highlighted by showing the nearby residues in stick representations. We use the protein-to-text functionality of \ourM{} to generate  functional annotations for these proteins, and the original outputs are provided here: (left protein) ``Heme-binding protein''; (right protein) ``Transfers electrons from cytochrome c551 to cytochrome oxidase; C-type cytochrome; Part of the cbb3-type cytochrome c oxidase complex.''}
    \label{fig:heme_bind_prot}
\end{figure}

In the text-based design case study (Fig.\ref{fig:heme_bind_prot} left), two histidine residues are positioned in close proximity to the iron located in the center of heme, enabling the formation of coordinated bonding interactions with the heme group. Similarly, in the SMILES-based design, the \ourM{} generates proteins with binding motifs similar to those generated in the text-based example (Fig. \ref{fig:SI:moreHemeCases}). However, as shown in Fig. \ref{fig:heme_bind_prot} (right), we show a representative case where a methionine and histidine residue are observed to interact closely with the iron ion (see Fig. \ref{fig:SI:prot_hem_hec} for more discussion on this case). These residues effectively coordinate the metal ion through their respective side chains, demonstrating alternative structural strategies for heme binding. Furthermore, the designed protein sequences differ significantly from those present in the database, indicating that \ourM{} can generate novel sequences with distinct structural properties. Collectively, these results demonstrate the effectiveness of \ourM{} in designing functional heme-binding proteins with diverse and novel structural features. We also compare the apo and holo structures of the generated proteins in Fig. \ref{fig:compare_apo_holo}, which shows that the key residues involved in heme binding, such as histidine and methionine, occupy similar positions in both structures.

%% file: chapters/material.tex
\section{Material tasks}\label{sec:material}
To evaluate the capabilities of \ourM{} for material generation, it is prompted to generate material's compositions in both unconditional and conditional way. For unconditional generation, the model is prompted with a special token indicating the start of material (i.e., \material{}) and is expected to generate the composition of the material (Section \ref{sec:uncondition_mat}). For conditional generation, the model is prompted to generate material formula and structure under specific human instructions, including: (1) Material generation for given composition (Section \ref{sec:comp_to_mat}); (2) Material generation for desired properties (Section \ref{sec:bulk_to_mat}). 

Crystal structures are fundamental to determining the physical, chemical, and mechanical properties of materials. Complementing the sequence outputs of \ourM{}, we finetune \ourM{} into a specialized model, \ourM{}-Mat3D, which predicts crystal structures from the generated chemical formulas (see Section \ref{sec:material_structure_predictor}). After generating material formulas and space groups with \ourM{}, we then utilize \ourM{}-Mat3D to convert them into crystal structures for further evaluation and practical application.


\subsection{Unconditional material generation}\label{sec:uncondition_mat}
The model is tasked with generating materials with arbitrary compositions. The input to \ourM{} is \material{}, and it produces material compositions with a specified space group. An example is provided below,
\begin{example}
\noindent\textbf{Instruction}: \material{} \\
\noindent\textbf{Response}: \material{} A A B B B $\langle$sg12$\rangle$\ematerial{}
\end{example}
where A, B refer to elements and $\langle$sg12$\rangle$ denotes the space group.

We evaluated the SMACT validity of the generated materials. 
Furthermore, we utilized \ourM{}-Mat3D to predict the crystal structures of a randomly selected subset of valid compositions. 
The energy above hull (abbreviated as ehull) of the predicted structures was then evaluated using MatterSim~\cite{yang2024mattersim}.
The distribution of ehull is shown in Fig. \ref{fig:mat_uncon}. We also assessed the ratio of stable materials, defining a generated material as stable if its ehull$<0.1$eV/atom. The results are presented in Table \ref{tab:uncon_mat}. It is evident that as the model size increases, the SMACT validity and stability of the generated materials improve.
\begin{table}[!htbp]
    \centering
    \begin{tabular}{lcc}
        \toprule
        Model & SMACT (\%) & Stability (\%) \\
        \midrule
        \ourM{} (1B) & 49.20 & 10.12\\ 
        \ourM{} (8B) & 63.42 & 12.47\\ 
        \ourM{} (8x7B) & 66.07 & 17.86\\
        \bottomrule
    \end{tabular}
    \caption{The SMACT validity and stability (with ehull$<0.1$eV/atom) for unconditional material generation. The distribution of ehull for the generated materials is illustrated in Fig. \ref{fig:mat_uncon}.}
    \label{tab:uncon_mat}
\end{table}

\subsection{Material generation for given composition}\label{sec:comp_to_mat}
The model is tasked with generating materials containing specific elements:
\begin{example}
\noindent\textbf{Instruction}: \texttt{Build a material that has Li, Ti, Mn, Fe, O}\\
\noindent\textbf{Response}: \material{} Li Li Li Li Ti Ti Ti Mn Mn Fe Fe Fe O O O O O O O O O O O O O O O O 
 $\langle$sg8$ \rangle$\ematerial{}
\end{example}
We evaluated the SMACT validity, stability, novelty, and precision of the generated materials. The novelty is measured as the ratio of unique generated materials that are not present in our instruction tuning data. The composition precision is calculated as 
\begin{equation}
{\rm composition\ precision} = \frac{1}{N} \sum_{i=1}^N \frac{\lvert E_{pi}\cap E_{gi}\rvert}{\lvert E_{pi}\rvert},
\label{eqn:mat_compt_to_mat_precision}
\end{equation}
where $E_{pi}$ and $E_{gi}$ stand for the sets of elements in the $i$-th prompt and corresponding generated material respectively.

\begin{figure}
    \centering
    \subfigure[\ourM{} (1B)]{
    \includegraphics[width=0.45\linewidth]{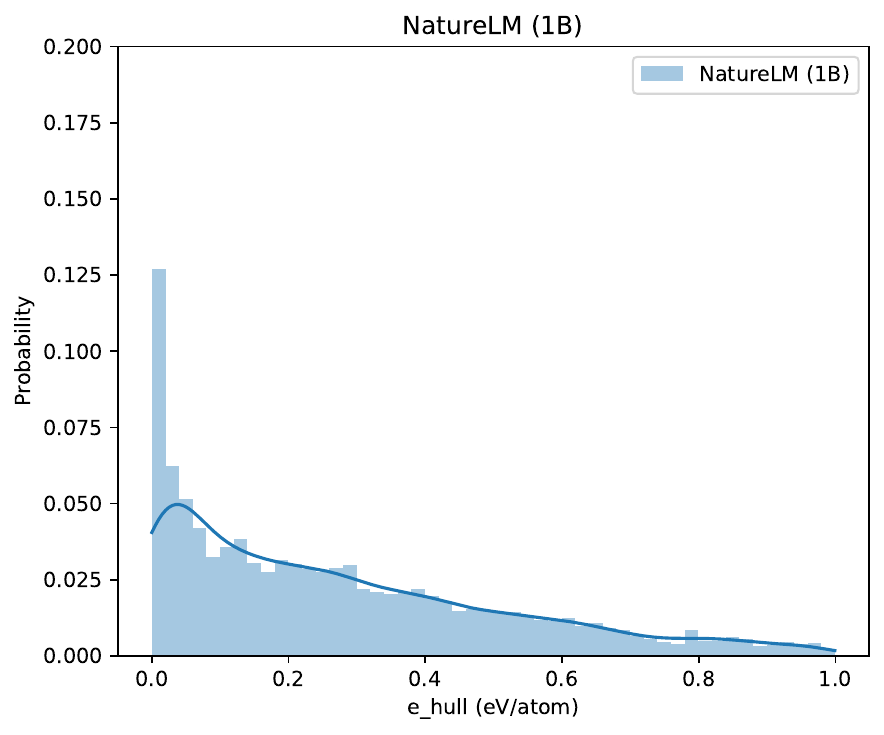}
    }%
    \subfigure[\ourM{} (8B)]{
    \includegraphics[width=0.45\linewidth]{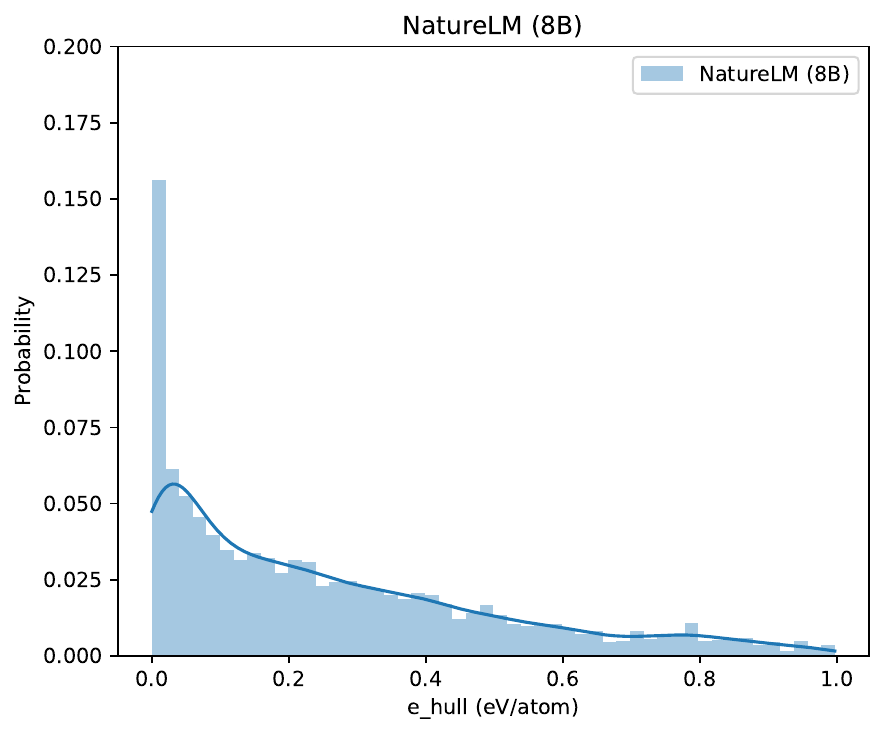}
    }%
    \vskip\baselineskip
    \subfigure[\ourM{} (8x7B)]{
    \includegraphics[width=0.45\linewidth]{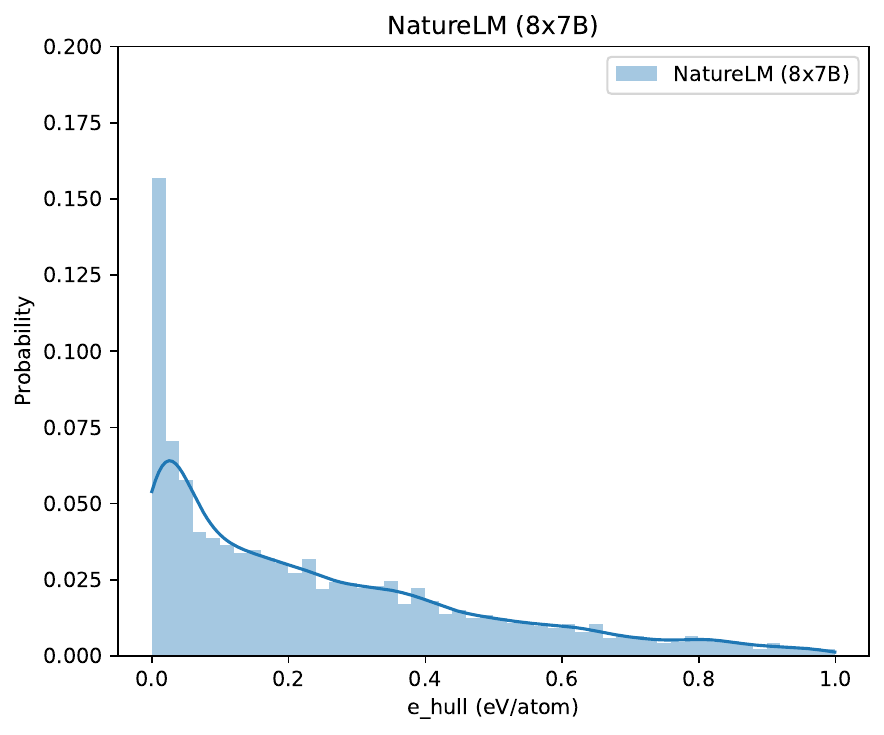}
    }
    \subfigure[Accumulated distribution]{
    \includegraphics[width=0.45\linewidth]{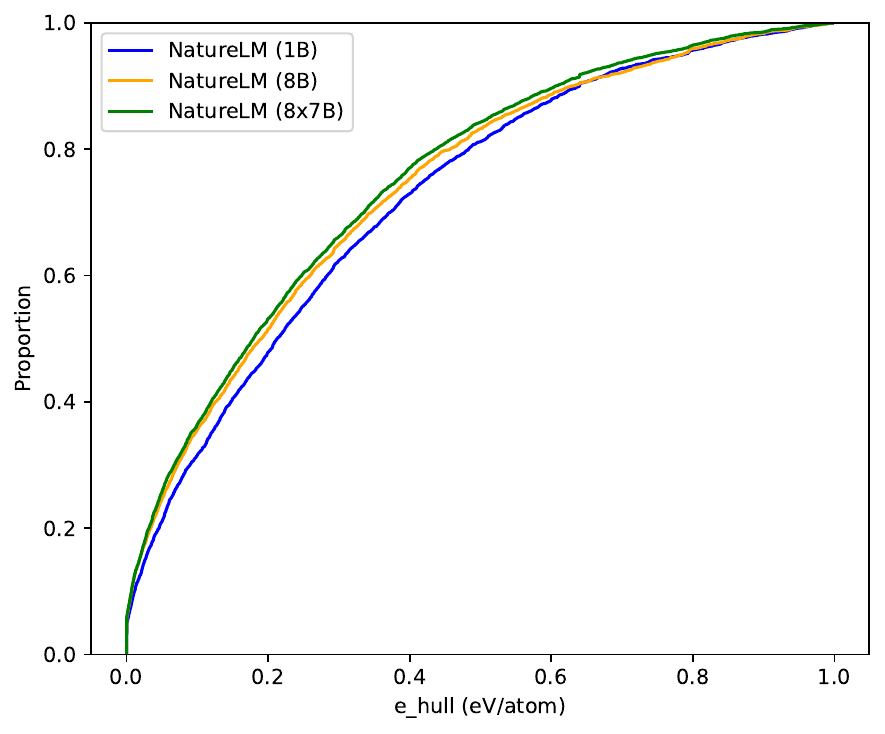}
    }
    \caption{Energy above hull (ehull) distribution for composition to material generation.}
    \label{fig:mat_comp_to_mat_ehull}
\end{figure}

The results are demonstrated in Table \ref{tab:mat_comp_to_mat}, and the distribution of ehull is depicted in Figure \ref{fig:mat_comp_to_mat_ehull}. Table \ref{tab:mat_comp_to_mat} shows a significant improvement in SMACT validity scores due to instruction tuning compared to unconditional generation. The precision for all three models is close to 100\%, indicating their strong capability to follow language instructions for generating material formulas with expected elements. Additionally, the high novelty demonstrates the models' generative abilities. Furthermore, stability improves with model size, highlighting their scalability. Figure \ref{fig:mat_comp_to_mat_ehull} illustrates this more clearly: as model size increases, the ehull distribution shifts closer to zero, indicating that more materials have lower energy and are in a more stable state. 
\begin{table}[!htbp]
\centering
\begin{tabular}{lcccc}
\toprule
Model & SMACT (\%) & Stability (\%) & Precision (\%) & Novelty (\%)\\
\midrule
\ourM{} (1B) & 79.38 & 31.56 & 97.95 & 97.13 \\
\ourM{} (8B) & 83.36 & 35.56 & 98.44 & 95.51 \\
\ourM{} (8x7B) & 81.56 & 36.46 & 97.68 & 94.83 \\
\bottomrule
\end{tabular}
\caption{The SMACT validity, stability, precision, and novelty for composition to material generation.}
\label{tab:mat_comp_to_mat}
\end{table}

\subsection{Material generation for desired properties}\label{sec:bulk_to_mat}
\ourM{} can generate materials with desired properties, and here we demonstrate this capability by generating materials that have specific bulk modulus.
The bulk modulus of a substance is a measure of the resistance of a substance to bulk compression. As a proof-of-concept, the \ourM{} is prompted to generate materials with the following instructions:
\begin{example}
\noindent\textbf{Instruction}: \texttt{Construct the composition for a material with a specified bulk modulus of 86.39 GPa.}\\
\noindent\textbf{Response}: \material{} Se Se Pd Sc 
 $\langle$sg164$ \rangle$\ematerial{}
\end{example}
We evaluated the SMACT validity, stability, novelty, and precision of the generated materials. Success rate is defined as the ratio of generated materials whose bulk modulus is within 10\% of the instructed value, compared to other generated materials. 

The results in Table \ref{tab:bulk_to_mat} indicate improved SMACT validity and stability as the model scales. Figure \ref{fig:mat_bulk_to_mat_ehull} depicts the distribution of ehull for the generated materials, showing a shift closer to zero with increasing model size.

\begin{figure}[!htbp]
    \centering
    \includegraphics[width=0.8\linewidth]{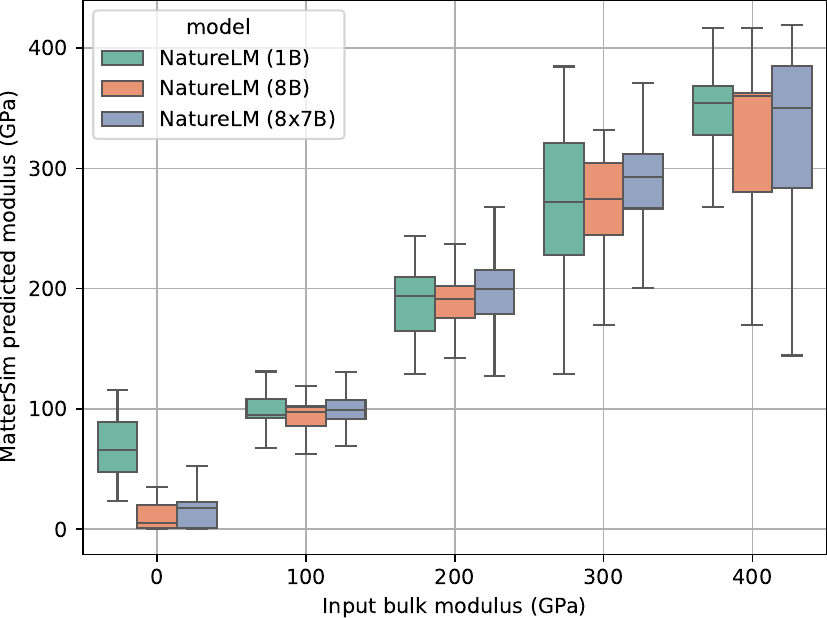}
    \caption{Distribution of predicted bulk modulus values for generated materials. The $x$-axis represents the input bulk modulus values from the instructions, while the $y$-axis shows the predicted values for the generated molecules calculated by MatterSim. }
    \label{fig:mat_bulk_to_mat}
\end{figure}

Further, to demonstrate how \ourM{} follows the instruction to generate materials with expected bulk modulus, we depict the distribution of the bulk modulus of generated materials under the instructions in Figure \ref{fig:mat_bulk_to_mat} where the $x$-axis denotes the bulk modulus in the instruction prompt and the $y$-axis denotes the predicted bulk modulus of the generated materials calculated by MatterSim. We can see that, as the model scales, the distribution aligns more closely with the ideal linear diagonal. 

To assess how many novel materials \ourM{} can generate, we prompted the model with a single instruction and allowed it to produce up to 1,000,000 material formulas. We then plotted the count of novel material formulas against the total number generated. Novel materials are defined as those passing the SMACT validity check, not present in the instruction tuning data, and not previously generated. Figure \ref{fig:mat_novelty} shows that the number of novel materials increases with the total generated. Even at 1 million generated materials, novel ones continue to appear, highlighting the model's strong generative capability.

\begin{table}[!htbp]
\centering
\begin{tabular}{lccccc}
\toprule
Model & SMACT (\%) & Stability (\%) & Precision (\%) & Novelty (\%) \\
\midrule
\ourM{} (1B) & 86.76 & 39.34 & 40.00 & 52.38 \\
\ourM{} (8B) & 87.21 & 52.81 & 44.06 & 36.31 \\
\ourM{} (8x7B) & 94.75 & 53.60 & 44.62 & 32.42 \\
\bottomrule
\end{tabular}
\caption{The SMACT validity, stability, precision, and novelty of generated materials conditioned on bulk modulus.}
\label{tab:bulk_to_mat}
\end{table}

\begin{figure}[!htbp]
    \centering
    \includegraphics[trim=5cm 3cm 5cm 1cm, clip, width=\linewidth]{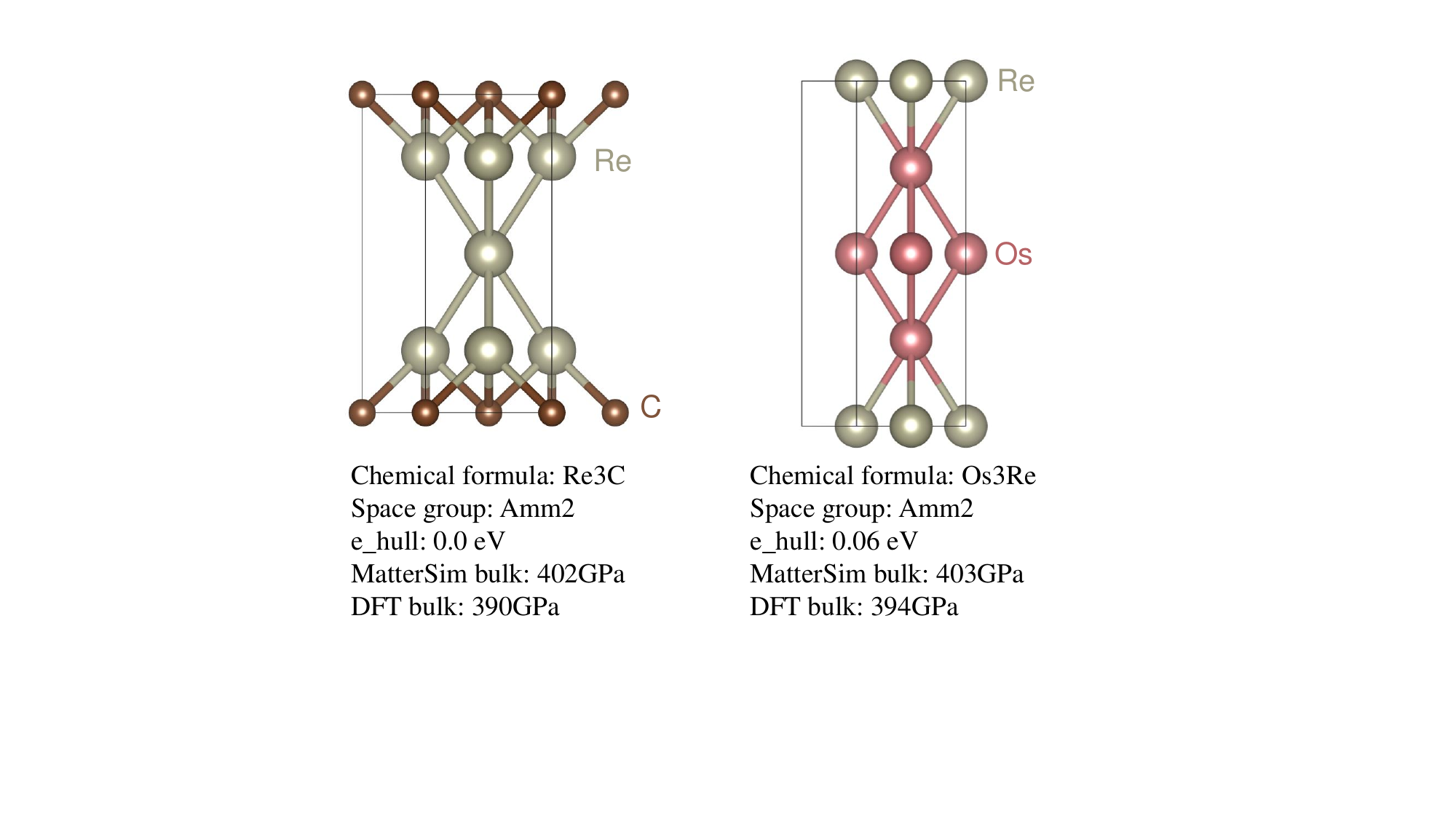}
    \caption{Two cases with bulk modulus values near 400 GPa (evaluated via DFT) were identified. The chemical formulas, space groups, energy above the hull (e\_hull), and bulk modulus values obtained from MatterSim and DFT calculations are provided.}
    \label{fig:bulk_caseStudy}
\end{figure}

Materials with an ultra-high bulk modulus are highly sought after due to their exceptional stiffness and incompressibility, making them indispensable for applications in extreme environments, such as aerospace, industrial tooling, and advanced engineering. To evaluate the potential of \ourM{} in generating materials with high bulk modulus, we conducted a detailed analysis of the generated compositions targeted at a bulk modulus of 400 GPa. From the generated outputs, we manually identified cases where MatterSim~\cite{yang2024mattersim} predicted bulk modulus values within a 5 GPa range of the target. Two such cases were selected for further validation using density functional theory (DFT) calculations (see Fig. \ref{fig:bulk_caseStudy}). The DFT results revealed bulk modulus values of 390 GPa and 394 GPa, which closely align with the target value of 400 GPa. 

Beyond achieving the bulk modulus target, the two generated structures were confirmed to be novel compared to those available in the Materials Project database. This novelty underscores \ourM{}'s potential for discovering new materials with exceptional mechanical properties, thereby broadening the scope of material design and innovation.

\subsection{\ourM{}-Mat3D: a crystal structure predictor for materials}\label{sec:material_structure_predictor}
Crystal material structure prediction (CSP) is a critical problem. Previous works apply random search, particle swarm algorithm, and a few others search algorithms to look for stable crystal structures. More recently, generative models like VAE \cite{cdvae}, diffusion \cite{zeni2023mattergen} and flow matching based methods \cite{flowmm} are applied for such 3D structure generation.   There is also a growing trend towards using Large Language Models (LLMs) for crystal structure generation, which can autoregressively generate the structures \cite{gruver2024finetunedlanguagemodelsgenerate,antunes2024crystalstructuregenerationautoregressive,flowmm}. We fine-tune \ourM{} to act as a crystal structure prediction module that generates 3D structures in an autoregressive manner. 

Using \ourM{} for structure prediction is particularly meaningful because it aligns the sequential modeling capacity of LLMs with the sequential representation of crystal structures. This congruence allows the model to capture the intricate dependencies and patterns inherent in material structures, potentially leading to more accurate and efficient generation of stable crystal configurations.

We represent materials and their 3D structures as 1D sequences in three steps:
\begin{enumerate}
\item {\em Flatten the chemical formula}: Repeat each element according to its count (e.g., \texttt{A2B3} becomes \texttt{A A B B B}). 
\item {\em Add space group information}: Append special tokens $\langle$sg$\rangle$ and $\langle$sg{N}$\rangle$, where \texttt{N} is the space group number.
\item {\em Include coordinate information}: Use the token $\langle$coord$\rangle$ to indicate the start of coordinates. Flatten the lattice parameters into nine float numbers and the fractional atomic coordinates into sequences of float numbers. Numbers are retained to four decimal places and tokenized character-wise (e.g., \texttt{-3.1416} as \texttt{- 3 . 1 4 1 6}).
\end{enumerate}
For example, the sequence for a material \texttt{A2B3} with space group number 123 is:
\begin{example}
A A B B B $\langle$sg$\rangle$ $\langle$sg123$\rangle$ $\langle$coord$\rangle$ {9 float numbers for lattice} {15 float numbers for atoms}
\end{example}

We collect data from Materials Project \cite{materialsproject}, NOMAD \cite{nomad} and OQMD \cite{oqmd2013,oqmd2015} as our training data which are widely used database for materials with structure information, and test on MP-20, Perov-5 and MPTS-52 following previous works \cite{cdvae,diffcsp,flowmm}. Specially, we remove duplications in the merged training data and remove all the data that appear in the test set in these benchmarks. The final training data contains about 6.5M samples after deduplication and removal of the test set. After training, we also finetuned the model on the training set for each benchmark to mitigate the different distributions between our training data and the benchmark data. We evaluate the match rate of the generated material structures and compare to CDVAE \cite{cdvae}, DiffCSP \cite{diffcsp} and FlowMM \cite{flowmm}. The results are shown in Table \ref{tab:mat_struct_pred}. Experiment results show that our sequence based auto-regressive method achieves comparable or best
performance on MP-20 and MPTS-52 compared to other methods. We will use this for material structure generation in our following experiments. In future work, we will leverage and combine with more advanced methods like MatterGen \cite{zeni2023mattergen} for structure generation.

\begin{table}[!htbp]
\centering
\begin{tabular}{lcccccc}
        \hline
         & \multicolumn{2}{c}{Perov-5} & \multicolumn{2}{c}{MP-20} & \multicolumn{2}{c}{MPTS-52} \\
        & MR (\%)  & RMSE & MR (\%) & RMSE & MR (\%) & RMSE\\
        \hline
        CDVAE  & 45.31 & 0.1138 & 33.90 & 0.1045 & 5.34 & 0.2106 \\
        DiffCSP & 52.02 & \textbf{0.0760} & 51.49 & 0.0631 & 12.19 & 0.1786 \\
        FlowMM & \textbf{53.15} & 0.0992 & 61.39 & 0.0566 & 17.54 & 0.1726 \\
        \ourM{}-Mat3D (1B) & 50.78 & 0.0856 & \textbf{61.78} & \textbf{0.0436} & \textbf{30.20} & \textbf{0.0837} \\
        \hline
    \end{tabular}
    \caption{The match rate (MR) and RMSE on Perov-5, MP-20 and MPTS-52.}
    \label{tab:mat_struct_pred}
\end{table}

\ourM{}-Mat3D (1B) achieves performance that is comparable to or surpasses other state-of-the-art methods. The high match rates and low RMSE values demonstrate that our model effectively captures the complex spatial arrangements of atoms in crystal structures. Moreover we can see that \ourM{}-Mat3D performs better than other methods as the number of atoms increases, demonstrating the advantage of autoregressive sequence model. As a next step, we plan to further improve the structure prediction quality by incorporating 3D autoregressive data into the pre-training phase of the next version of \ourM{}.

%% file: chapters/Nucleotide.tex
\section{Nucleotide tasks}\label{sec:nucleotide}

The genome contains a vast amount of information regarding protein-coding genes and the regulatory DNA and RNA sequences that control their expression. In this section, we evaluated our model on nucleotide sequence generation tasks, including both unconditional generation and cross-domain generation, specifically DNA to RNA generation (guide RNA design) and protein to RNA generation.

\subsection{Unconditional RNA generation}\label{sec:unconditional_rna_generation}
Designing RNA molecules is crucial for advancing RNA vaccines, nucleic acid therapies, and various biotechnological applications. In this section, we evaluate the proficiency of NatureLM in generating RNA sequences without any conditional prompts. For evaluation purposes, we constrained the generated RNA sequences to a maximum length of 1024 nucleotides. An example of an unconditionally generated sequence is provided below:
\begin{example}
\noindent\textbf{Instruction}: \rna{} \\
\noindent\textbf{Response}: \rna{} C C A C G G A G C C \erna{}
\end{example}

We assessed the quality of the generated RNA sequences by calculating their Minimum Free Energy (MFE) using RNAfold~\cite{Lorenz2011} (see Section \ref{app:rna_generation} for details). A lower MFE value indicates a potentially more stable RNA secondary structure. For each model, we generated 5,000 sequences and computed their MFE values. To establish a baseline for comparison, we generated control sequences and computed their average MFE values. Specifically, for each generated sequence, we created:

\noindent(1) {\em Shuffled Sequences}: For each generated sequence, we created a new sequence by randomly shuffling its nucleotides, thereby preserving the original nucleotide composition and length but potentially disrupting any inherent structural motifs. 

\noindent(2) {\em Random Sequences}: For each generated sequence, we created an entirely random sequence of the same length, where each nucleotide position was independently sampled from the four nucleotides (A, G, C, U) with equal probability. This baseline represents sequences with no designed structure or composition bias.

As a reference for the MFE values of natural RNA sequences, we randomly sampled 5,000 sequences of length up to 1,024 nucleotides from RNAcentral\footnote{\url{https://rnacentral.org/}} and computed their MFE values.

The average MFE values are reported in Table \ref{tab:unconditional_rna_generation_mfe}.

\begin{table}[!htpb]
\centering
\begin{tabular}{lccccccc}
\toprule
               & MFE (kcal/mol) & Retrieved Rfam Families \\
\midrule
RNAcentral     & -165.4 \\
Shuffled sequences & -156.4 \\
Random sequences &  -142.0 \\
\ourM{} (1B)   & -160.6 & 23 \\
\ourM{} (8B)   & -170.6 & 38 \\
\ourM{} (8x7B) & -177.1 & 165 \\
\bottomrule
\end{tabular}
\caption{Average MFE values (in kcal/mol) of RNA sequences generated by different methods and the number of unique Rfam families retrieved by different models. \textbf{MFE} denotes the mean Minimum Free Energy of the sequences. \textbf{Retrieved Rfam Families} represents the count of unique RNA families identified in the generated sequences using \texttt{cmscan}. }
\label{tab:unconditional_rna_generation_mfe}
\end{table}

From the results, we observe that larger models tend to generate RNA sequences with lower (more negative) MFE values, indicating potentially more stable secondary structures. Additionally, shuffling and randomizing the sequences result in higher (less negative) MFE values, suggesting that the original sequences generated by our models have structural features that contribute to stability.

To evaluate the diversity of the RNA sequences generated by NatureLM, we compared them to known RNA families in Rfam~\cite{10.1093/nar/gkaa1047}. We used \texttt{cmscan} from the Infernal toolkit~\cite{10.1093/bioinformatics/btt509} to search for structural similarities between our generated sequences and the Rfam database (see Section \ref{app:rna_generation} for details). As shown in Table~\ref{tab:unconditional_rna_generation_mfe}, larger models retrieved a significantly higher number of unique Rfam families than smaller models: the 1B, 8B, and 8x7B models retrieved 23, 38, and 165 unique families, respectively, covering a wider range of RNA functions. These results suggest that larger models not only generate more stable sequences but also encompass a more diverse set of RNA structures and functions.


\subsection{Guide RNA design}
Guide RNA (gRNA) plays a critical role in CRISPR-Cas9 gene editing by directing the Cas9 nuclease to a specific genomic target site. A gRNA consists of two components: a crispr RNA (crRNA), which is approximately 20 nucleotides long and complementary to the target DNA sequence, and a tracrRNA, which binds to Cas9. The crRNA garners more attention as it features a variable sequence critical for targeting specific DNA, whereas the tracrRNA has a more fixed role, primarily aiding in Cas9 binding.  We evaluate NatureLM on two gRNA design tasks: the first is designing crRNA for a given DNA sequence, and the second is selecting the more effective crRNA from two candidates. Examples are provided below:

\begin{example}
\noindent$\rhd$gRNA generation\\
\noindent{\textbf{Instruction}}: \\
\texttt{Generate a guide RNA for targeting the DNA sequence} \\
\dna{}GACTGGCACCAG$\cdots$CCCTCGC\edna{}.\\	
\noindent{\textbf{Response}}: \rna{}AGACACAGCGGGTGCTCTGC\erna{}\\

\noindent$\rhd$More effective gRNA identification\\
\noindent{\textbf{Instruction}}: 
\noindent \texttt{Investigate which of} \rna{}ATGTAGAAGAATCCACC\\
ATA\erna{} 
\texttt{or} \rna{}GGAAGGGGTCAATATTCTCA\erna{} \texttt{results in better wild-type efficiency for the DNA sequence} \\
\dna{}AAGGGGTGGCA$\cdots$AGTGC\edna{}.	

\noindent{\textbf{Response}}: \rna{}ATGTAGAAGAATCCACCATA\erna{}
\end{example}

For the first task of designing crRNA, the NatureLM's identifies a targeted DNA region that includes the PAM sequence (Protospacer Adjacent Motif). To obtain the final crRNA, remove the 'NGG' PAM sequence, reverse complement the DNA sequence, and apply the central dogma principle by converting DNA into RNA (replacing all 'T' with 'U'). These quick and easy steps ensure the resulting crRNA is ready for use. 
A valid crRNA must meet the following criteria: (1) Be 17 to 24 nucleotides long. (2) Match a specific region in the provided DNA sequence. (3) Be followed by an "NGG" Protospacer Adjacent Motif (PAM) in the DNA sequence, where "N" represents any of the four standard nucleotide bases: adenine (A), cytosine (C), guanine (G), or thymine (T).

The example demonstrates that NatureLM accurately identifies the target sequence for guide RNA, ensuring the resulting guide RNA meets all necessary criteria--with an impressive accuracy of 95.7\% in designing valid gRNAs.

\ourM{} demonstrates a strong ability to generate valid gRNA sequences compared with  generalist models like GPT4, accurately targeting the specified DNA while maintaining the PAM sequence feature. Furthermore, \ourM{} shows proficiency in assessing the effectiveness of gRNAs, enabling it to select the more efficient gRNA from a given pair (Table \ref{tab:RNA:RNA generation tasks}). 
\begin{table}[!htbp]
\centering
\begin{tabular}{lccc}
\toprule
Model & Validity & Top 1 accuracy\\
\midrule
GPT-4 & 0.272 & 0.597 \\
Llama 3 8B & 0 & 0.38 \\
Mixtral 8x7B & 0 & 0.46 \\
\midrule
\ourM{} (1B) & 0.95 & 0.681 \\
\ourM{} (8B) & 0.765 & 0.657 \\
\ourM{} (8x7B) & 0.957 & 0.699 \\
\bottomrule
\end{tabular}
\caption{The performance of guide RNA design.}
\label{tab:RNA:RNA generation tasks}
\end{table}



\subsection{Protein binding RNA design}\label{sec:protein2rna}
RNA-binding proteins (RBPs) represent a large and diverse class of over 2,000 proteins that play a crucial role in regulating gene expression by interacting with RNA. Designing RNA decoys offers a powerful strategy to manipulate these interactions. Such decoys can sequester RBPs away from their natural RNA targets, act as competitors to displace natural RNA molecules from RBPs, or serve as scaffolds to recruit RBPs to specific RNA molecules or cellular locations.


\begin{example}
\noindent\textbf{Instructions}:  \texttt{Given} \pro{}MSEY$\cdots$SSGWGM\epro{}, \texttt{create an RNA molecule that binds to it}.	\\
\noindent\textbf{Response}:
\rna{}AAACAGG$\cdots$CGTACGACA\erna{}
\end{example}

We selected 200 targets and generated binding RNA for them. To evaluate the generation ability of \ourM{}, following \cite{xu2023prismnet}, we trained a predictor for each protein to predict the binding affinity between the RNA and the protein. Specifically, the final layer of the classifier is a sigmoid function, which produces an output value ranging from 0 to 1, indicating the strength of the RNA-protein binding. If the score is greater than 0.5, we consider the generated RNA to have successfully bound to the protein.

We compared the RNA sequences generated by \ourM{} 1B, 8B and 8x7B. Additionally, we used the predictors to evaluate the binding and non-binding RNA sequences from the test set. We also randomly selected RNA sequences of the same sizes from the unconditional generation setting for prediction (Section \ref{sec:unconditional_rna_generation}). 

The results are summarized in Table~\ref{tab:RNA:rbp2rna}, which reports the average and median prediction scores, as well as the success rate—the proportion of sequences with a prediction score above 0.5. We have the following observations: 
\begin{enumerate}
\item As expected, binding RNA sequences achieved the highest average prediction score of 0.714 and a success rate of 74.5\%, while the non-binding RNA sequences had the lowest average score of 0.274 and a success rate of 24.4\%. This confirms the reliability of the classifiers and serving as a benchmark for optimal performance.
\item For unconditioned RNA Sequences, with an average score of 0.391 and a success rate of 36.3\%, these sequences performed better than non-binding sequences but significantly worse than the binding sequences. This suggests that random RNA sequences have a moderate chance of being predicted as binders due to the intrinsic properties of RNA but lack the specificity achieved through conditioning.
\item For \ourM{} generated sequences, as we increase the model sizes, there is a clear trend that larger models perform better. The results also demonstrated that \ourM{} is more likely to generate RNA sequences that are likely to bind to the specified proteins when explicitly conditioned on them.
\end{enumerate}

\begin{table}[!h]
\centering
\begin{tabular}{lccc}
\toprule
Source          & AVG Score & Success rate (\%)\\
\midrule
Binding         & 0.714     & 74.5 \\
Non-binding     & 0.274     & 24.4 \\
Unconditioned   & 0.391     & 36.3 \\
\ourM{} (1B)    & 0.415     & 40.9 \\
\ourM{} (8B)    & 0.434     & 44.2 \\
\ourM{} (8x7B)  & 0.438     & 44.8 \\
\bottomrule
\end{tabular}
\caption{Performance of designing protein-binding RNA given proteins: ``AVG Scores" refers to the average prediction scores across all sequences. Success ratio refers to the percentage of scores that are greater than 0.5. See Figure \ref{fig:enter-label} for the detailed distribution of the predicted scores.}
\label{tab:RNA:rbp2rna}
\end{table}.

%% file: chapters/prediction.tex
\section{Prediction tasks}
In addition to the generation and design tasks studied in previous sections, we also studied the predictive capabilities of \ourM{}. 

\subsection{Small molecule prediction tasks}
We evaluated \ourM{} on three molecular property prediction tasks from MoleculeNet \cite{moleculenetpaper}: (i) predicting whether a molecule can cross the blood-brain barrier (BBBP); (ii) predicting whether a molecule can bind to the BACE receptor (BACE); (iii) predicting the toxicity of a molecule associated with 12 targets (Tox21). An illustrative example is presented below:

\begin{example}
\textbf{Instruction:}\\
\texttt{Can \mol{}C1(c2ccccc2)=CCN(C)CC1\emol{} traverse the blood-brain barrier?}\\ 
\textbf{Response:} \\
\texttt{Yes}.
\newline
\newline
\noindent\textbf{Instruction:} \\
\texttt{Could the compound \mol{}N(O)=C1CCC([NH2+]CC(O)C
\noindent(Cc2cc(F)cc(F)c2)NC(C)=O)(c2cccc(C(C)(C)C)c2)CC1\emol{} potentially restrain beta-secretase 1?}\\
\textbf{Response:} \\
\texttt{Yes}. 
\end{example}

To determine the probability of a ``Yes'' or ``No'' response, we first extract the probabilities output by the NatureLM, denoting the probability of ``Yes'' as $p_1$ and the probability of ``No'' as $p_2$. We then normalized these probabilities: the probability of ``Yes'' is calculated as $p_1/(p_1+p_2)$ while the probability of ``No'' is $p_2/(p_1+p_2)$. 

All tasks in this subsection are measured by AUROC\footnote{\url{https://scikit-learn.org/1.5/modules/generated/sklearn.metrics.roc_auc_score.html}}. The results are reported in Table \ref{tab:molnet}. Generally, larger models achieve better performance, while there is still a gap between the current \ourM{} and the state-of-the-art specialist models. 

\begin{table}[!htpb]
\centering
\begin{tabular}{lcccc}
\toprule
& BBBP & Bace & Tox21  \\
\midrule
DVMP \cite{2021jinhuaDVMP} & 78.1 & 89.3 & 78.8 \\
BioT5 \cite{PeiQizhi2023BioT5} & 77.7 & 89.4 & 77.9  \\
NatureLM (1B) & 71.1 & 79.4 & 68.3 \\
NatureLM (8B) & 70.2 & 82.0 & 69.8 \\
NatureLM (8x7B) & 73.7 & 83.1 & 72.0\\
\bottomrule
\end{tabular}
\caption{Molecular property prediction on MoleculeNet \cite{moleculenetpaper}. The evaluation metric is the AUROC score. }
\label{tab:molnet}
\end{table}

\subsection{Protein prediction tasks}

We evaluated the \ourM{} on four protein property classification tasks, including solubility prediction, stability prediction, and protein-protein interaction (PPI) prediction for both human and yeast proteins. These datasets as well as the data splits are adopted from the PEER benchmark~\cite{xu2022peer}. An example is provided below:

\begin{example}
\textbf{Instruction:}\\
\texttt{Does the sequence of this protein suggest it would be stable? Please answer `Yes' if it is stable and `No' if it is not.} \pro{}\text{TTIKVNG \ldots KVTR}\epro{} \\
\textbf{Response:} \\
\texttt{No}.
\newline
\newline
\noindent\textbf{Instruction:} \texttt{Could these proteins interact, considering their sequences? The first protein is} \\
\pro{}MPPS \ldots VETVV\epro{}, \texttt{the second protein is} \pro{}MSLHF \ldots PLGCCR\epro{}. \texttt{Please respond with `Yes' if the proteins can interact and `No' if they cannot.} \\
\textbf{Response:} \\
\texttt{Yes} 
\end{example}  

\begin{table}[!h]
\centering
\begin{tabular}{ccccc}
\toprule
Model Setting & Solubility & Stability & Human PPI & Yeast PPI \\
\midrule
Literature SOTA & 0.702 & - & 0.881 & 0.661 \\
SFM-Protein (650M) \cite{he2024sfm} & 0.744 & 0.583 & 0.852 & 0.628 \\
\midrule
\ourM{} (1B) & 0.684 & 0.682 & 0.781 & 0.561 \\
\ourM{} (8B) & 0.714 & 0.635 & 0.848 & 0.604 \\ 
\ourM{} (8x7B) & 0.698 & 0.723 & 0.776 & 0.586 \\ 
\bottomrule
\end{tabular}
\caption{Protein understanding task comparison (accuracy). Please note that the \ourM{} models are trained for various diverse tasks in a unified way and evaluated separately on these tasks by the same model, while the state-of-the-art models are trained and tested individually for each task. }
\label{tab:protein:classification}
\end{table}

Table \ref{tab:protein:classification} presents the accuracy of our models on various protein understanding tasks. Overall, the results highlight that our unified \ourM{} models perform competitively with task-specific models, even surpassing them in certain tasks like stability prediction. This demonstrates the effectiveness of our training strategy, where a single model can learn and generalize across diverse protein understanding tasks without the need for separate models for each task.

\subsection{DNA prediction tasks}\label{sec:dna_property_pred}
We selected two classification tasks to assess the model's capability of identifying significant sequence motifs implicated in human gene regulation. These tasks include the identification of promoters and transcription factor binding sites.  We utilized datasets from the Genome Understanding Evaluation (GUE, see Appendix B.2 of \cite{zhou2024dnabert2} for a summary), converting them into a format suitable for instruction tuning. 
An example is shown below:

\begin{example}
\textbf{Instruction:}\\
\texttt{Verify if there is a promoter region within \dna{}TGGACT$\cdots$TGAGCTC\edna{}?}  \\
\textbf{Response:} \\
\texttt{Yes}.
\newline
\newline
\noindent\textbf{Instruction:} \\
\texttt{Can the sequence \dna{}GCCTGCCAG$\cdots$AAAAC\edna{} be classified as a transcription factor binding site?}  \\
\textbf{Response:} \\
\texttt{No}. 
\end{example}

\begin{table}[!h]
\centering
\begin{tabular}{lccc}
         \toprule
         Model & Promoter detection & Core promoter detection & TF binding\\
         \midrule
         NT-2500M-multi \cite{dalla2023nucleotide} & 0.881 & 0.716 & 0.633 \\
         DNABERT2 \cite{zhou2023dnabert} & 0.842 & 0.705 & 0.701 \\
         \midrule
         \ourM{} (1B) & 0.805 & 0.571 & 0.524 \\
         \ourM{} (8B) & 0.827 & 0.595 & 0.549 \\
         \ourM{} (8x7B) & 0.835 & 0.602 & 0.560 \\
         \bottomrule
\end{tabular}
\caption{Performance comparison of various models on DNA property prediction tasks, evaluated using Matthews Correlation Coefficient (MCC).}
\label{tab:DNA:DNA prediction tasks}
\end{table}
Table \ref{tab:DNA:DNA prediction tasks} presents the results of our experiments, evaluated using the Matthews Correlation Coefficient (MCC). For the Transcription Factor Binding prediction task, we conducted separate predictions on specific ChIP-seq datasets: POLR2A ChIP-seq on human HUVEC, POLR2A ChIP-seq on human ProgFib, PAX5 ChIP-seq protocol v041610.1 on human GM12892, TRIM28 ChIP-seq on human U2OS, and MXI1 ChIP-seq on human H1-hESC produced by the Snyder lab. We then calculated the average performance across these datasets. Despite a performance gap between our models and the state-of-the-art, the observed improvements with increasing model sizes suggest potential for further advancements. These findings indicate that larger models may more effectively capture the complex regulatory motifs involved in human gene regulation.




%% file: chapters/performance_improvement.tex
\section{Strategies to further improve performance}
In this section, we examine two strategies to improve the model's performance: reinforcement learning for scenarios with limited labeled data for fine-tuning specific tasks, and dedicated fine-tuning for cases where sufficient labeled data is available for particular tasks.

\subsection{Reinforcement enhanced \ourM{}}
\label{sec:RL}
Reinforcement Learning with Human Feedback (RLHF) is well-established approach to enhance foundation models. This section explores how to utilize preference signals in RLHF, moving beyond reliance on direct supervised signals\footnote{It is important to note that direct signals have already been used in post-training (see Section~\ref{sec:post_train}.}. For many generative tasks, where answers are open-ended and do not have a single correct solution, training with preference signals offers a more intuitive approach.

For RLHF training, we curated preference data from nine property optimization tasks related to small molecules: BBBP, BACE, LogP, Donor, QED, CYP1A2, CYP2C9, CYP2D6, and CYP3A4. Detailed descriptions of each task and the corresponding data quantities can be found in Table \ref{tab:data-rlhf}. In total, we compiled 179.5k data points. Note that we used all the data to enhance the post-trained \ourM{} (8B), resulting in a single model for the nine tasks after RLHF.

The data is structured in a preference-based format, where each sample consists of a prompt, along with both an accepted and a rejected response. An example of this format is presented below:

\begin{example}
\textbf{Instruction: }\\\texttt{Enhance the effectiveness of the molecule }\mol{}COc1cc2c(c(OC)c1OC)-c1ccc(OC)c(=O)cc1[C@@H](NC(C)=O)CC2 \emol{}\texttt{ in penetrating the blood-brain barrier.}\\
\textbf{Accepted Response: }\\\mol{}COc1cc2c(c(OC)c1OC)-c1ccc(OC)c(=O)cc1C(NC(C)=O)CC2\emol{}\\
\textbf{Reject Response: }\\\mol{}COc1cc2c(c(OC)c1OC)-c1ccc(OC)c(=O)cc1[C@@H](NC(C)=O)\\CC2\emol{}
\end{example}  

In the example above, the compound in the accepted response is capable of crossing the BBB, whereas the compounds in the instruction and rejected response cannot.

We leveraged Direct Preference Optimization (DPO)~\cite{rafailov2024direct} to enhance the molecule optimization ability of \ourM{}. The loss of DPO algorithm is written as follows:

\begin{align}
    \mathcal{L}_{\text{DPO}}(\pi_\theta; \pi_{\text{ref}}) = -\mathbb{E}_{(x, y_w, y_l) \sim \mathcal{D}} \left[ \log \sigma \left( \beta \log \frac{\pi_\theta(y_w \mid x)}{\pi_{\text{ref}}(y_w \mid x)} - \beta \log \frac{\pi_\theta(y_l \mid x)}{\pi_{\text{ref}}(y_l \mid x)} \right) \right],
\end{align}
where $\pi_{\rm ref}$ is the model after post-training and fixed during DPO training, $\pi_\theta$ is the model to optimize and set to $\pi_{\rm ref}$ before DPO training, $x$ is the prompt, $y_w$ is the accepted response, $y_l$ is the reject response, and $\beta$ is a hyper-parameter.

\begin{table}[h]
\centering
\begin{tabular}{lc}
\toprule
Property & $\Delta$  \\
\midrule
QED  & 0.6 \\
LogP  & 0.6 \\
Donor  & 0.6 \\
BBBP  & 2.9 \\
BACE  & 3.5 \\
CYP1A2  & 2.3 \\
CYP2C9  & 0.7 \\
CYP2D6  & 0.7 \\
CYP3A4  & 1.0 \\
\bottomrule
\end{tabular}
\caption{Results of the reinforcement optimization. Let $r_1$ and $r_2$ represent the outcomes before and after applying reinforcement, respectively, and let $\Delta$ denote the percentage improvement, i.e., $\Delta = (r_2 - r_1)/r_1 \times 100\%$. }
\label{tab:dpo-result}
\end{table}

Table \ref{tab:dpo-result} shows the improvements of DPO training over the 9 property optimization tasks. Notably, the model had already undergone instruction tuning (the post-training in Section~\ref{sec:smallmol}) prior to DPO training, and no new data was introduced during the DPO process. The results highlight how reformatting the data into a preference-based structure allows the DPO algorithm to improve performance across multiple tasks simultaneously.

Looking ahead, we plan to generate data on the fly in RLHF and utilize additional reward models to evaluate the properties of newly generated molecules, thereby creating better preference-based data.

\subsection{Dedicated fine-tuning on retrosynthesis}\label{sec:dedicate_tune}

We dedicatedly fine-tuned our \ourM{} model to evaluate its performance against specialized models in the retrosynthesis prediction task, using a large-scale labeled dataset, the Pistachio reaction dataset~\cite{mayfield2017pistachio}, with 15 million reactions from U.S., European, and WIPO patents. 
To ensure data quality, we removed any invalid or duplicate reactions.
The cleaned dataset was then randomly split into a training set with 3.1 million reactions and a test set with 34,000 reactions.

Before training, we preprocessed the input products and output reactants using a root-aligned SMILES format~\cite{Zhong2022rsmiles}.
This format offers a clear one-to-one mapping between the product and reactant SMILES, 
thereby enhancing prediction efficiency.
Additionally, we augmented the training dataset tenfold to further improve the model's performance.
As shown in Table~\ref{tab:retro-pistachio}, NatureLM (1B) demonstrates competitive performance, rivaling leading template-based models (e.g., LocalRetro) and template-free models (e.g., R-SMILES) on the large Pistachio dataset.

\begin{table}[!h]

\centering
\begin{tabular}{lcc}
\toprule
& Top-1 accuracy & Top-3 accuracy \\
\midrule
LocalRetro~\cite{chen2021localretro} & 40.8\% & 56.6\% \\
R-SMILES~\cite{Zhong2022rsmiles} & 51.2\% & 67.1\% \\ 
\midrule
\ourM{}~(1B) & 51.4\% & 66.0\% \\
\bottomrule
\end{tabular}
\caption{Retrosynthesis prediction results on Pistachio dataset.}
\label{tab:retro-pistachio}
\end{table}

\subsection{Dedicated fine-tuning on Matbench}\label{sec:dedicate_tune_matbench}
We fine-tuned our \ourM{} 8B model on Matbench~\cite{matbench}, a benchmark for state-of-the-art machine learning algorithms that predict various properties of solid materials. Matbench is hosted and maintained by the Materials Project~\cite{materialsproject}. Following the approach in~\cite{xie2023darwin}, we fine-tuned a single model for three tasks from Matbench, rather than developing separate models for each task. 

The results are presented in Table~\ref{tab:pi-matbench-expt-gap} to  \ref{tab:pi-matbench-glass}. Results of baseline models are collected from the official leader board\footnote{\url{https://matbench.materialsproject.org/}}. As can be seen, \ourM{} achieves state-of-the-art performance on matbench\_expt\_gap and matbench\_is\_metal. 

\begin{table}[h]
  \centering
  \begin{minipage}{0.3\linewidth}
    \centering
    \begin{tabular}{lc}
    \toprule
    Model &  MAE $\downarrow$ \\
    \midrule
    Dummy\cite{matbench} & 1.1435\\
    gptchem\cite{Jablonka_2023} & 0.4544\\
    RF-SCM/ & \multirow{2}{*}{0.4461}\\
    Magpie\cite{matbench} & \\
    AMMExpress\cite{matbench} & 0.4161\\
    MODNet\cite{De_Breuck_2021} & 0.3327\\
    Ax/SAASBO & \multirow{2}{*}{0.3310} \\
     CrabNet\cite{Wang2021crabnet,erikssonHighDimensionalBayesianOptimization2021} & \\
    DARWIN\cite{xie2023darwin} & 0.2865 \\
    \midrule
    \ourM{} & \textbf{0.2858} \\
    \bottomrule
    \end{tabular}  
    \vspace{0.35cm} 
    \caption{Results on matbench\_expt\_gap.}
    \label{tab:pi-matbench-expt-gap}
  \end{minipage}
  \hfill  
  \begin{minipage}{0.3\linewidth}
    \centering
    \begin{tabular}{lc}
    \toprule
    Model &  F1 $\uparrow$ \\
    \midrule
    Dummy\cite{matbench} & 0.4913\\
    gptchem\cite{Jablonka_2023} & 0.8953\\
    MODNet\cite{De_Breuck_2021} & 0.9153\\
    RF-SCM/ & \multirow{2}{*}{0.9159}\\
    Magpie\cite{matbench} & \\
    AMMExpress\cite{matbench} & 0.9200\\
    DARWIN\cite{xie2023darwin} & 0.9599 \\
    \midrule
    \ourM{} & \textbf{0.9630} \\
    \bottomrule
    \end{tabular}
    \vspace{1cm} 
    \caption{Results on matbench\_is\_metal.}  
    \label{tab:pi-matbench-is-metal}
  \end{minipage}
  \hfill  
  \begin{minipage}{0.3\linewidth}
    \centering
    \begin{tabular}{lc}
    \toprule
    Model &  F1 $\uparrow$ \\
    \midrule
    Dummy\cite{matbench} & 0.7127 \\
    DARWIN\cite{xie2023darwin} & 0.8722 \\
    gptchem\cite{Jablonka_2023} & 0.8782\\
    RF-SCM/ & \multirow{2}{*}{0.9278}\\
    Magpie\cite{matbench} & \\
    AMMExpress\cite{matbench} & 0.9043\\
    MODNet\cite{De_Breuck_2021} & \textbf{0.9784}\\
    \midrule
    \ourM{} & 0.8720 \\
    \bottomrule
    \end{tabular}
    \vspace{1cm} 
    \caption{Results on matbench\_glass.}  
    \label{tab:pi-matbench-glass}
  \end{minipage}
\end{table}




%% file: chapters/text.tex
\section{Text capabilities }\label{sec:text_ability}
We assess the text-related capabilities of \ourM{} (8x7B) in this section. We first present several examples to provide readers with an intuitive understanding (Section \ref{sec:case_study}), and then benchmark results from AlpacaEval (Section \ref{sec:alpahaeval}).

\subsection{Case study}\label{sec:case_study}
Here we demonstrate our model's language capabilities in scientific discovery research. Specifically, we assessed the model's understanding of CRISPR gene editing, cancer immunotherapy, and drug discovery. A few selected cases are presented in this section for brevity. Additional cases can be found in Fig. \ref{SI_fig:case_study:CRISPR}, \ref{SI_fig:case_study:cancer_immunotherapy} and \ref{SI_fig:case_study:drug_discovery}.

\subsubsection{CRISPR gene editing}
 CRISPR gene editing has rapidly emerged as a powerful tool in biotechnology. Our model demonstrates a good understanding of CRISPR and is able to effectively analyze the advantages and disadvantages of commonly used CRISPR tools.

\begin{mdframed}[backgroundcolor=white, linecolor=black, linewidth=1pt]
    \noindent
    \textbf{Instruction:} Could you provide me the advantages and disadvantages of Cas9? \\
    \textbf{Response:} \\
    Advantages: 1. Highly efficient and specific gene editing with a low risk of off-target effects. 2. Can target a wide range of sequences.  3. Cam be used to make specific alterations to genes in virtually any organism.  Disadvantages: 1. Potential for off-target effects when using Cas9. 2. The delivery of Cas9 can be inefficient. 3. Lack of control over the repair mechanism. 4. May cause chromosomal abnormalities. \\

    \noindent
    \textbf{Instruction:} Could you provide me the advantages and disadvantages of Cas12a, another widely used CRISPR protein? \\
    \textbf{Response:} \\
    Advantages: \\
    1. High specificty: Cas12a has higher specificity compared to Cas 9, which reduces off-target effects. 
    2. Simpler PAM requirements: Cas 12a requires a simpler PAM sequence than Cas9, making it easier to target specific genomic loci. 
    3. Multiplexing: Cas12a can be used for multiplexing genome editing, meaning multiple genomic loci can be edited in a single reaction. 
    Disadvantages: \\
    4. Lower efficiency: Cas12a is less efficient in terms of indel formation than Cas9. 
    5. Larger size: Cas12a is a larger protein than Cas9, which can make it harder to deliver the Cas12a system into cells. \\
    ... 
\end{mdframed}  
\captionof{figure}{Cases about CRISPR gene editing.} \label{fig:case_study:CRISPR}

\subsubsection{Cancer immunotherapy}
Our model is knowledgeable about existing checkpoint inhibitors used in cancer therapy and provides accurate explanations of their mechanisms of action. Specifically, in our experiment, it generates a valid peptide sequence for inhibiting the overexpressed PD-L1 protein (see Figure \ref{SI_fig:case_study:cancer_immunotherapy}), although the length of the peptide is incorrect. It demonstrates a solid understanding of the inhibition cascade between drugs and proteins.

\begin{mdframed}[backgroundcolor=white, linecolor=black, linewidth=1pt]  
    \noindent
    \textbf{Instruction:} Tell me existing checkpoint inhibitors for PD-L1. \\
    \textbf{Response:} The checkpoint inhibitors for PD-L1 that are currently approved by the FDA include atezolizumab, avelumab, duravalumab, and perbrolizumab.\\
    
    \noindent
    \textbf{Instruction:} PD-L1 is a protein that inhibits PD-1, a recptor found on T cells. This new drug X, is designed to inhibit PD-L1. If we introduce drug X, will T cell function be restored? Explain it step by step. \\
    \textbf{Response:}
    PD-1 is inhibited by interaction with PD-L1. If drug X inhibits PD-L1, PD-1 will no longer be inhibited. Therefore, T cell function will beestored. Therefore, the answer is yes.

\end{mdframed}  
\captionof{figure}{Cases about cancer immunotherapy.} \label{fig:case_study:cancer_immunotherapy}

\subsubsection{Drug discovery}
Drug discovery is an evolving process that incorporates empirical rules, computational design, and experimental validation. The model demonstrates an understanding of the fundamental principles of drug discovery. Specifically, when being asked questions about Lipinski’s Rule of Five and strategies (see Figure \ref{SI_fig:case_study:drug_discovery}) to improve a drug’s ADME properties, the model provides mostly accurate answers.

\begin{mdframed}[backgroundcolor=white, linecolor=black, linewidth=1pt]  
    \textbf{Instruction:} What approaches can drug experts use to enhance a drug's ADME properties? \\
    \textbf{Response:} Structure-based drug design: Computational techniques can be used to modify existing drugs or design new ones with desired ADME properties. 2. Lipophilicity: Lipophilicity can affect the ADME profile of a drug. 3. Hepatic and extracellular clearance: Understanding the hepatic and extracellular clearance of a drug is important for assessing its ADME profile. 4. Microsomal stability: Microsomal stability is a measure of how quickly a drug is broken down by liver enzymes. 5. Oral absorption: Oral absorption is determined by the solubility and permeability of a drug. 6. Permeability: The ability of a drug to cross cell membranes affects its bioavailability. 
    ... 

\end{mdframed}  
\captionof{figure}{Case about drug discovery.} \label{fig:case_study:drug
_discovery}

\subsection{Comparison on AlpacaEval evaluation}\label{sec:alpahaeval}
To assess the linguistic capabilities of \ourM{}, we compared \ourM{} (8x7B) with the original Mixtral  (8x7B) using the AlpacaEval \cite{dubois2024length} dataset and method. AlpacaEval is a benchmark for LLM designed to gauge a model's proficiency in executing general user instructions. \ourM{}'s responses are compared against reference responses generated by original Mixtral using GPT-4 as the annotator. We focused on two metrics from AlpacaEval: the general win rate and the LC win rate (length-controlled win rate), which controls the bias introduced by the length of the responses. Our NatureLM model achieved a LC win rate of 31.8\% and a win rate of 25.2\% when benchmarked against Mixtral, showcasing that while the model possesses a foundational level of language comprehension, it exhibits constraints in its linguistic capacity.   A potential contributing factor to this limited performance is the overrepresentation of scientific data within the training dataset, which may skew the model's proficiency towards technical language and away from a more generalized understanding required for diverse user instructions. Fig. \ref{fig:case_study:alpacaeval} presents two illustrative examples from the AlpacaEval comparison. We will improve the text capabilities of \ourM{} in our future work. 

%% file: chapters/discussion.tex
\section{Ablation study}\label{sec:ablation_study}
\begin{figure}[!htbp]
\centering
\includegraphics[width=0.8\linewidth]{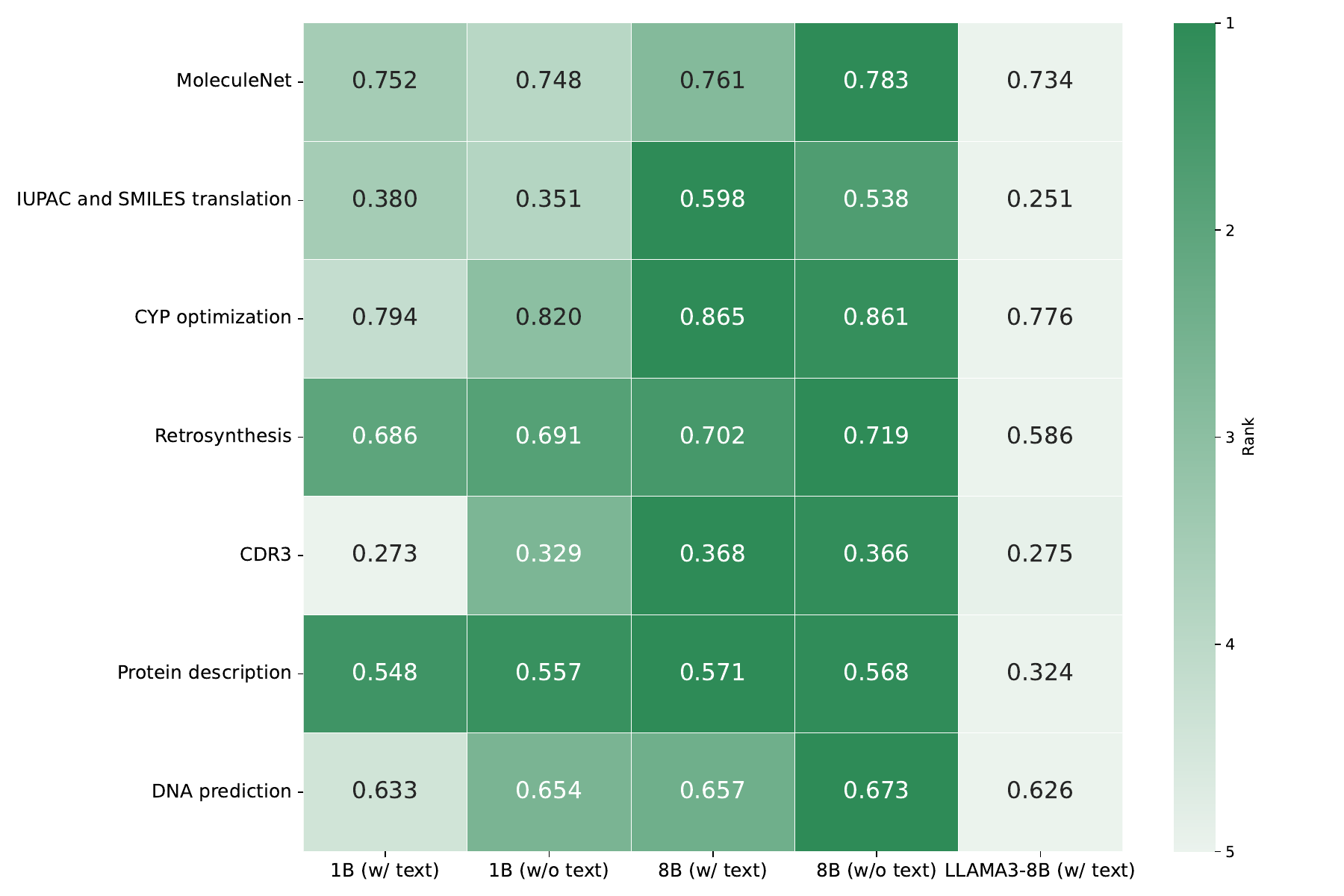}
\caption{Ablation study results. NatureLM models are denoted as ``(w/ text)''. The NatureLM models without text instruction data for post-training are denoted as ``(w/o text)''. Additionally, we fine-tuned the original Llama 3 model, denoted as ``Llama 3 8B (w/ text)''. Performance metrics are displayed in the cells (larger values indicate better performance), with rank represented by the color intensity (darker colors signify higher rankings).}
\label{fig:ablation_study}
\end{figure}

To better understand the contributions of different components in our model and training process, we conducted ablation studies focusing on two aspects: (1) the impact of including general text-based instruction tuning data in post-training, and (2) the effectiveness of continuing pretraining on scientific data before post-training, as opposed to directly fine-tuning from the baseline model. We evaluated the results across 7 tasks, including four small molecule tasks (Molecular property prediction, IUPAC to SMILES translation, Retrosynthesis, Metabolism optimization), two protein tasks (Protein description generation, CDR-H3 generation) and two DNA tasks (DNA property prediction). The results are reported in Fig. \ref{fig:ablation_study}. 

\subsection{Impact of text-based post-training data}
We investigated whether incorporating general text instruction tuning data into the post-training phase affects the performance of \ourM{}. To this end, we compared the results of \ourM{} pest-trained with or without the inclusion of general text-based instruction data during the post-training phase. \ourM{} (1B w/o text) and \ourM{} (8B w/o text) denote the models without text data for post-training. We have several observations from Fig. \ref{fig:ablation_study}:
\begin{itemize}
\item For the 1B parameter models, post-training without general text-based instruction data leads to better performance on scientific tasks, as evidenced by the superior performance on 5 out of 8 tasks. This suggests that at smaller scales, including text-based instruction data may not provide benefits and could potentially dilute the model's focus on scientific instructions due to limited model capacity.
\item In contrast, for the 8B parameter models, post-training with text-based instruction data does not adversely affect performance. This indicates that the larger model has sufficient capacity to incorporate both general text-based and science-based instruction data without detrimental effects on its performance on scientific tasks.
\end{itemize}

\subsection{Impact of continued pre-training on scientific data}
One might wonder whether it is necessary to pre-train a foundation model on scientific data or if directly fine-tuning a large language model (LLM) with scientific instruction data suffices. To address these questions, we compared our NatureLM (8B) model, which initially continues pre-training of Llama 3 8B on scientific data before undergoing post-training with scientific instruction data, against a model that directly fine-tunes the Llama 3 8B model without the pre-training step. As shown in Fig. \ref{fig:ablation_study}, \ourM{} (8B) outperforms the directly fine-tuned Llama 3 8B across all tasks, highlighting the importance of pre-training on a scientific corpus.

\section{Discussions}
\subsection{Summary}
In this work, we developed Nature Language Model (\ourM{}), a sequence-based science foundation model for scientific discovery across multiple domains. Users can interact with the model using text-based instructions to generate novel scientific entities. It supports cross-domain generation and has been demonstrated in phases of drug discovery, protein generation, RNA generation, and enables predictive capabilities for small molecules, proteins, and DNA. Among the 22 tasks tested, larger models showed better performance on 18 tasks. We believe \ourM{} is a significant step towards transforming scientific discovery with foundation model approaches. 

\subsection{Limitations}
Despite the progress of \ourM{}, we have identified several limitations and are committed to addressing them in future versions:

{\em Language Capabilities}: Interacting with scientific models using human language will be an essential feature to enable scientific discoveries. Although \ourM{} demonstrates general language capabilities, it achieves only a 31.8\% winning rate on the AlpacaEval benchmark when compared to the original Mixtral 8x7B. To enhance this, we plan to incorporate more high-quality text data in pre-training in the future.

{\em Few-shot Capabilities}: The capability of few-shot learning is critical for a foundation model. Currently, our NatureLM does not exhibit strong few-shot capabilities. We aim to enhance this by refining our training strategies and increasing the model size.

\subsection{Cross-domain applications}
\ourM{} is a unified model that spans multiple domains, including text, small molecules, proteins, materials, and nucleotides. One significant advantage of this multi-domain unification is that it allows for the integration of knowledge from diverse fields, enabling us to tackle important cross-domain tasks that domain-specific models cannot address. While we have already provided a few examples of cross-domain tasks, here are several more that we plan to study in the future:
\begin{enumerate}
\item \emph{Design of Biocompatible Materials:} Developing biocompatible materials requires the simultaneous consideration of material properties and protein interactions. Examples included the Titanium Alloys and Cobalt-Chromium Alloys used in hip replacement.
\item \emph{Ribozyme and Bio-Catalyst Development:} Designing effective ribozymes and bio-catalysts necessitates a detailed understanding of RNA structures, protein functions, and small molecule interactions.
\item \emph{Enabling Complex System Understanding:} Systems biology aims to understand the complex interplay of various components in a system, including biomolecules such as proteins, DNA, RNA, lipids, carbohydrates, and small molecules like metabolites.
\end{enumerate}

\clearpage

\section*{Author list}\label{sec:authorlist}
Yingce Xia$^{*}$, Peiran Jin$^{*}$, Shufang Xie$^{*}$, Liang He$^{*}$, Chuan Cao$^{*}$, 

Renqian Luo$^{*}$, Guoqing Liu$^{*}$, Yue Wang$^{*}$, Zequn Liu$^{*}$, Yuan-Jyue Chen$^{*}$, 

Zekun Guo$^{*}$, Yeqi Bai, Pan Deng, Yaosen Min, Ziheng Lu, 

Hongxia Hao, Han Yang, Jielan Li, Chang Liu, Jia Zhang, 

Jianwei Zhu, Ran Bi, Kehan Wu, Wei Zhang, Kaiyuan Gao, Qizhi Pei, 

Qian Wang, Xixian Liu, Yanting Li, Houtian Zhu, Yeqing Lu, 

Mingqian Ma, Zun Wang, Tian Xie, Krzysztof Maziarz, Marwin Segler, 

Zhao Yang, Zilong Chen, Yu Shi, Shuxin Zheng, Lijun Wu, 

Chen Hu, Peggy Dai, Tie-Yan Liu, Haiguang Liu, Tao Qin

\vspace{0.5cm}

\noindent$^*$ indicates co-first authors. 
 
\noindent Corresponding authors: Tao Qin and Yingce Xia 

\noindent Contact emails: \{taoqin, yingce.xia\}@microsoft.com

\noindent This work was conducted in Microsoft Research AI for Science.

\section*{Acknowledgements}
We extend our gratitude to Dr. Fan Yang and Dr. Jilong Xue for their support with large-scale model training. We thank Likun Dong and Junren Li for conducting the case study on retrosynthesis and the SMILES-to-IUPAC translation. Our thanks also go to Dr. Claudio Zeni, Dr. Robert Pinsler, Dr. Daniel Z\"{u}gner, Dr. Andrew Fowler, Dr. Matthew Horton, and Dr. Ryota Tomioka for their assistance with material tasks. We appreciate the constructive feedback from Dr. Bichlien Nguyen, Dr. Jake Smith, and Dr. Frank No\'{e}. We thank Dr. Han Guo for visualizing the molecules in our paper. We acknowledge Jingyun Bai for  improving the quality of the figures. We thank Dr. Christopher Bishop for his invaluable guidance and sponsorship of this work.

This work was done when Zekun Guo, Kehan Wu, Wei Zhang, Kaiyuan Gao, Qizhi Pei, Qian Wang, Xixian Liu, Yanting Li, Houtian Zhu, Yeqing Lu, Mingqian Ma, Zhao Yang, Zilong Chen were interns at Microsoft Research AI for Science. 

%% file: chapters/SI.tex
\section{Supplementary figures}
\begin{figure}[!htb]
    \centering
    \includegraphics[width=0.5\linewidth]{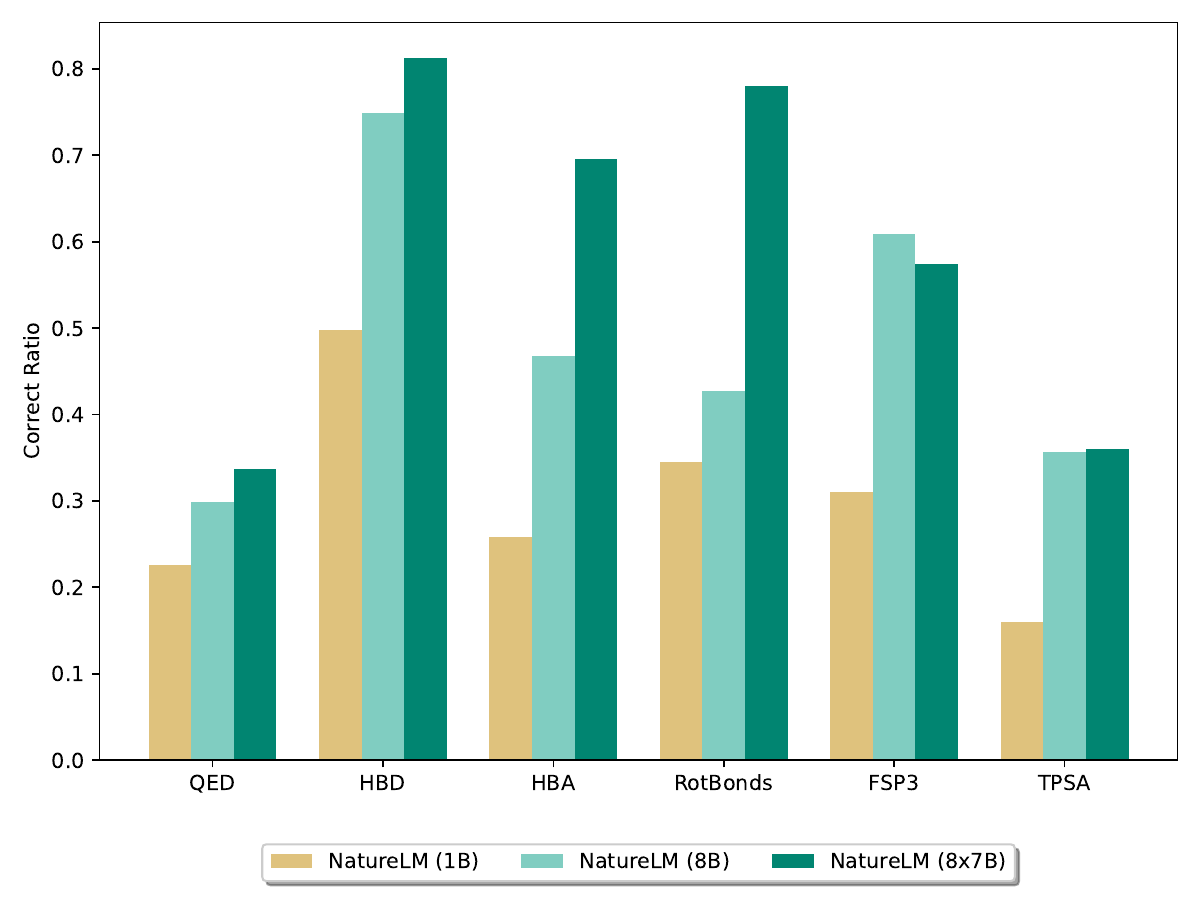} 
    \caption{Correct ratio of property-to-molecule generation. We treat the generated molecule as a correct one if $\lvert v' - v \lvert \leq \delta$, where $v'$ is its property value and $v$ is the input value. Tolerance threshold $\delta$ is set to 0 for HBA, HBD, RotBonds, 0.05 for QED and FSP3, and 5 for TPSA. }
    \label{fig:prop2mol_correct}
\end{figure}
\clearpage
\begin{figure}[!htb]
\centering
\subfigure[QED]{
\includegraphics[width=0.5\linewidth]{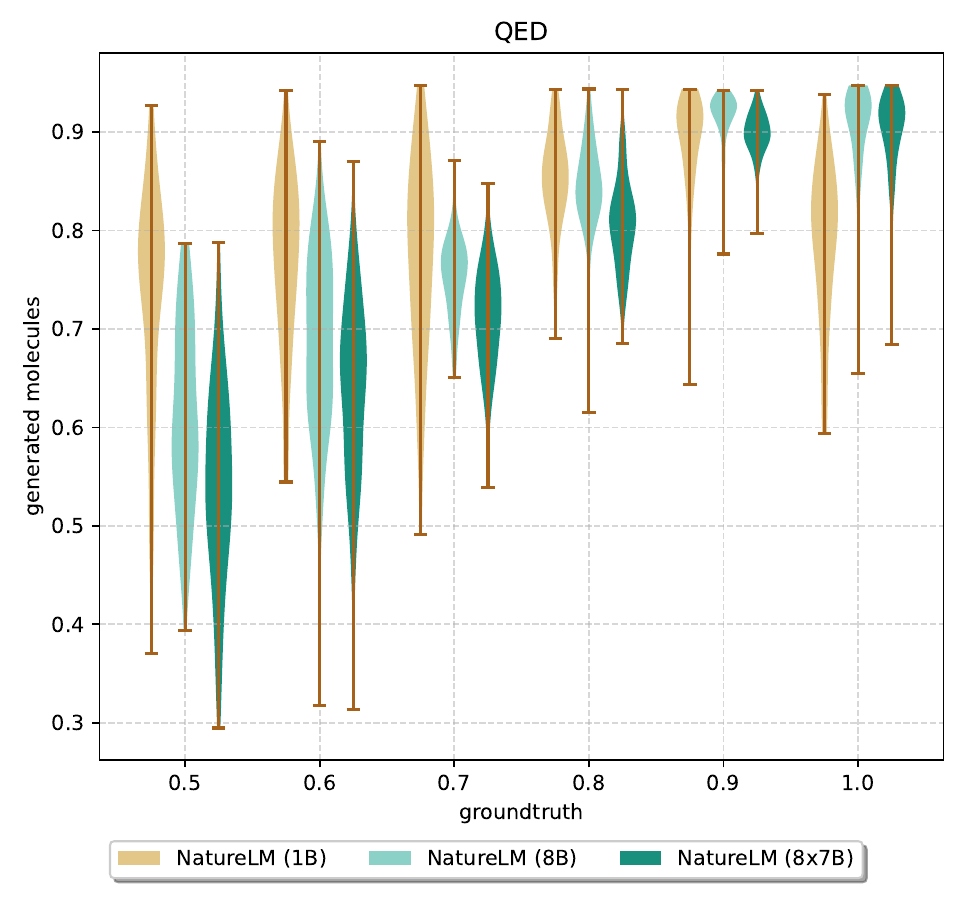} 
}%
\subfigure[HBA]{
\includegraphics[width=0.5\linewidth]{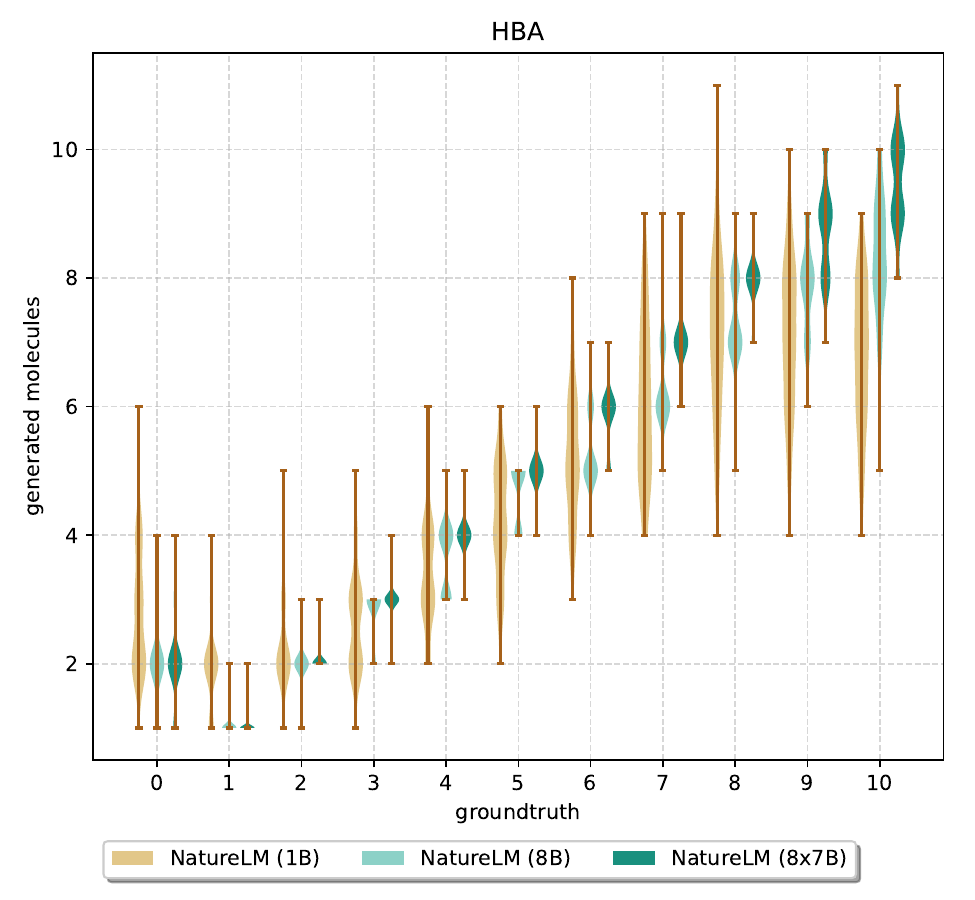} 
}
\subfigure[HBD]{
\includegraphics[width=0.5\linewidth]{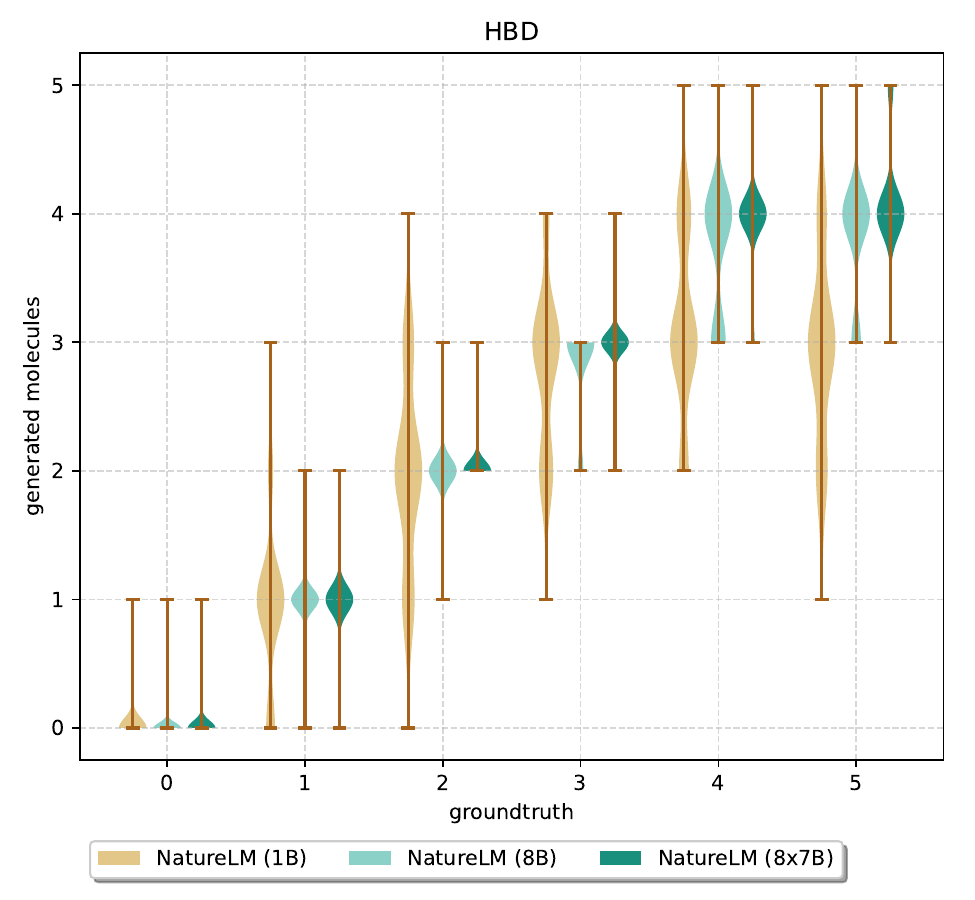} 
}%
\subfigure[Rotatable bonds]{
\includegraphics[width=0.5\linewidth]{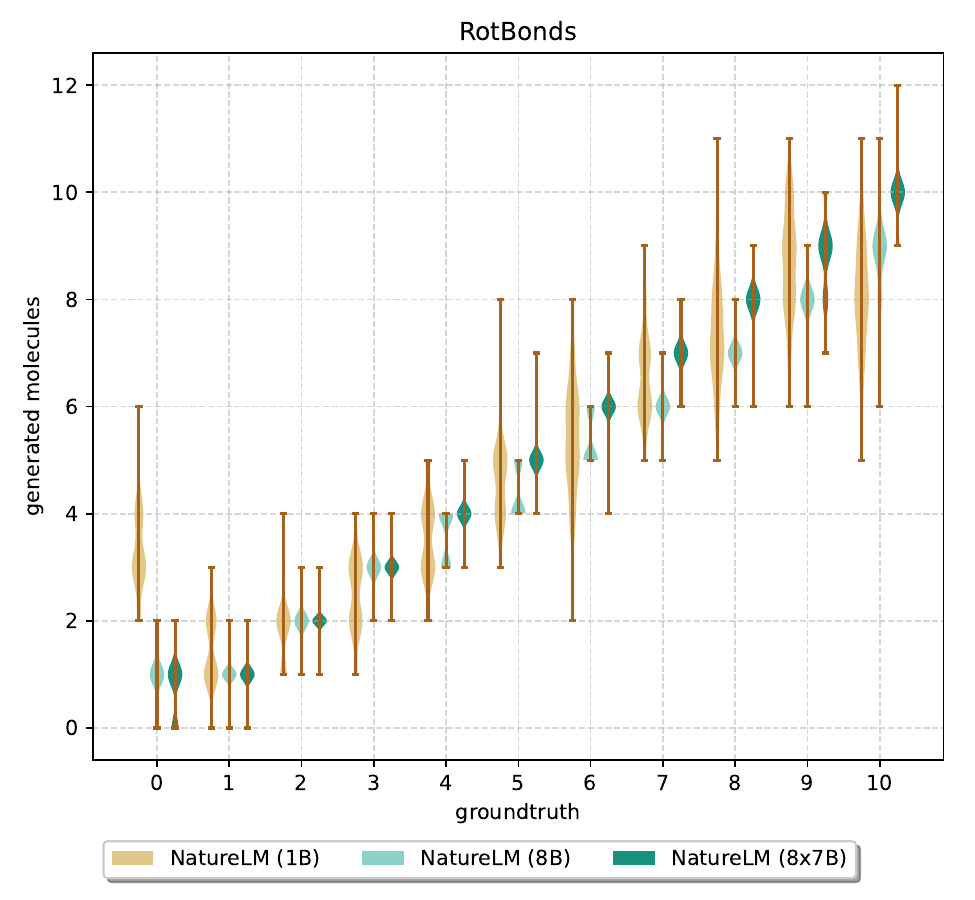} 
}
\caption{Violin plot of basic molecular properties for molecule generation, including QED, the number of hydrogen bond acceptors (HBA), the number of hydrogen bond donors (HBD) and the number of rotatable bonds.}
\label{fig:basic_to_cmpd_violinplot}
\end{figure}

\clearpage
\begin{figure}[!htb]
\centering
\subfigure[QED=0.8, FSP3=0.4]{
\includegraphics[width=0.5\linewidth]{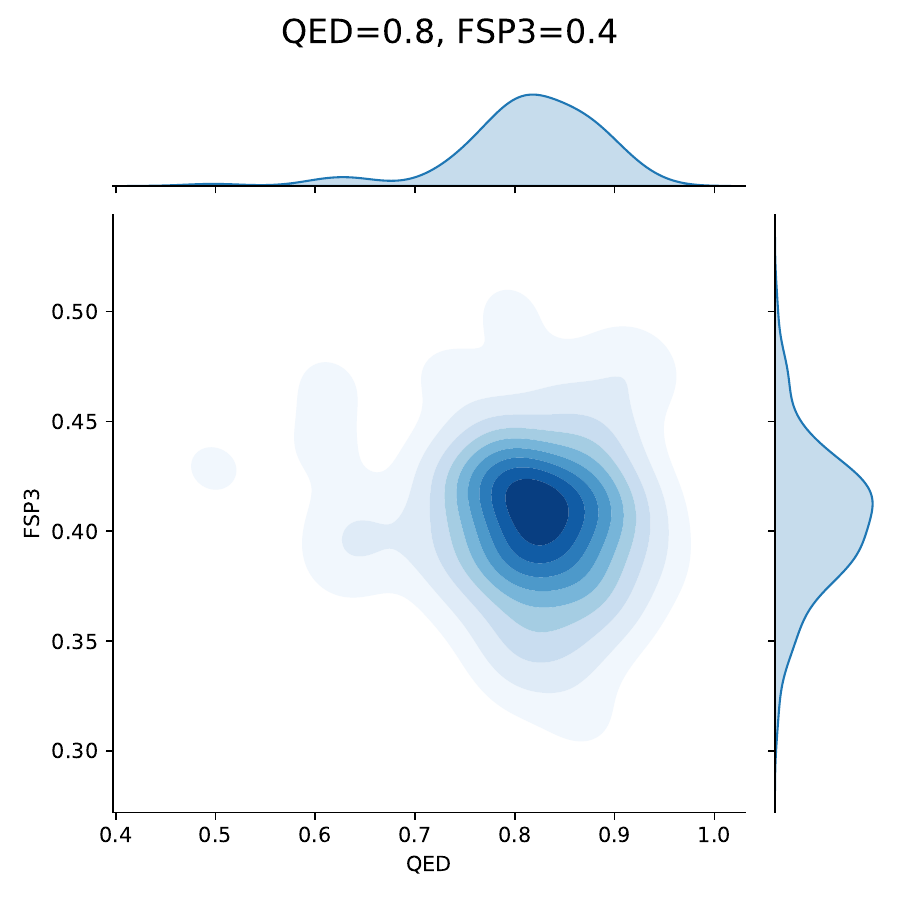}
}%
\subfigure[QED=0.8, FSP3=0.6]{
\includegraphics[width=0.5\linewidth]{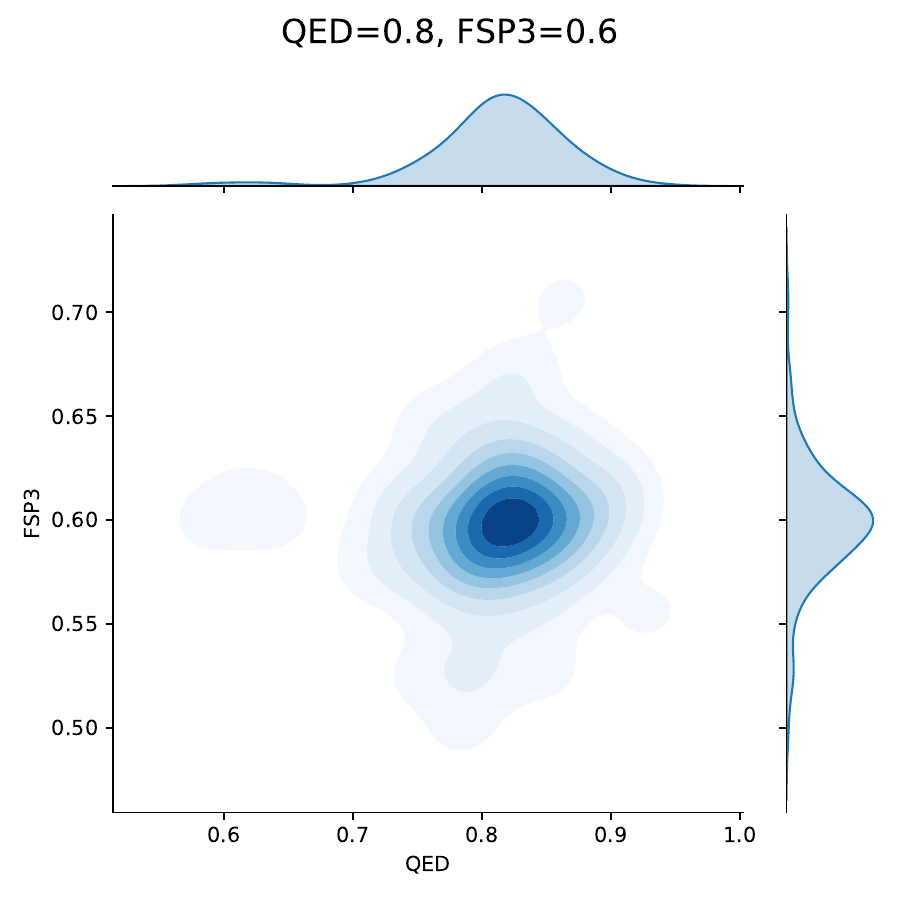}
}
\caption{Heatmap of molecule generation based on QED and fraction of sp³ (FSP3) properties. Each generated compound's QED and FSP3 values are calculated using RDKit and visualized in the heatmap.}
\label{fig:qed_fsp3_joint_optim}
\end{figure}
\begin{figure}[!htb]
    \centering
    \includegraphics[width=0.7\linewidth]{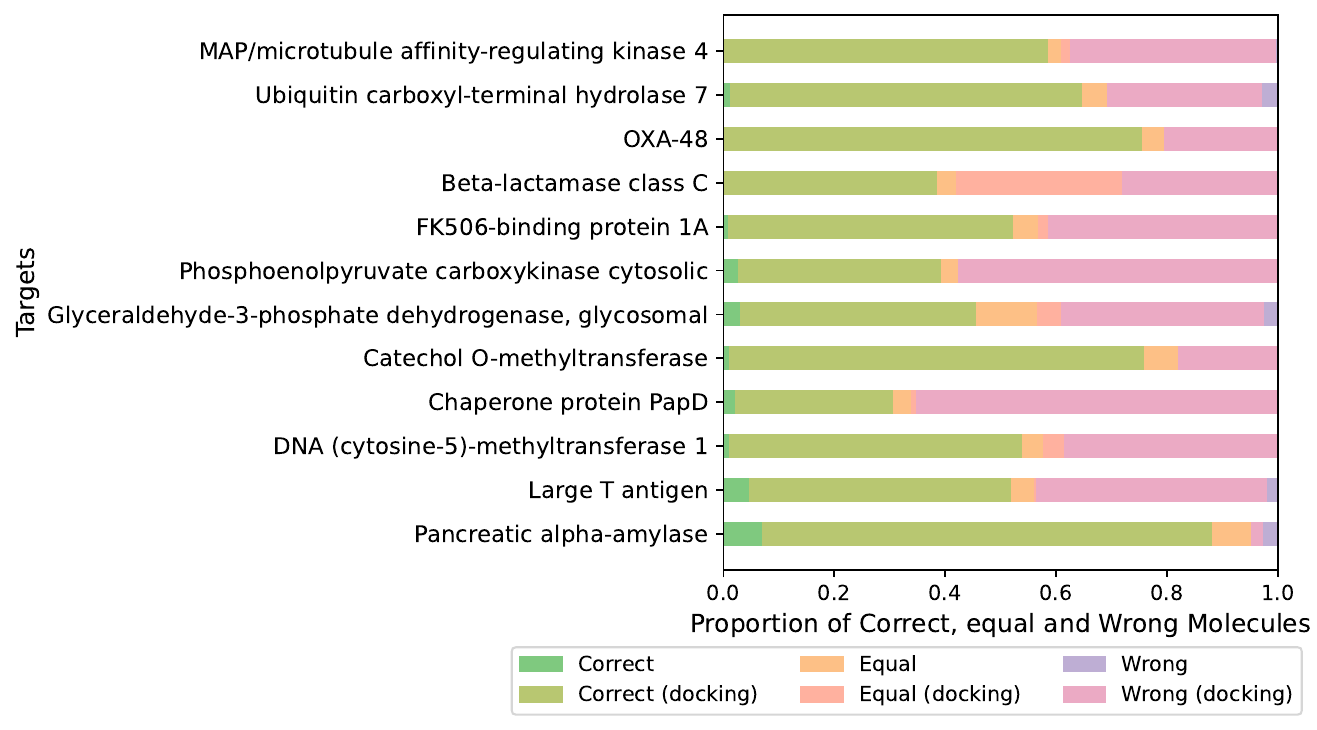} 
    \caption{Bar plot of the proportion of correct, equal and wrong generated molecules. Molecules evaluated by retrieval and molecules evaluated by docking are distinguished using different colors.}
    \label{fig:binding_docking}
\end{figure}
\begin{figure}[!htb]
    \centering
    \includegraphics[width=0.7\linewidth]{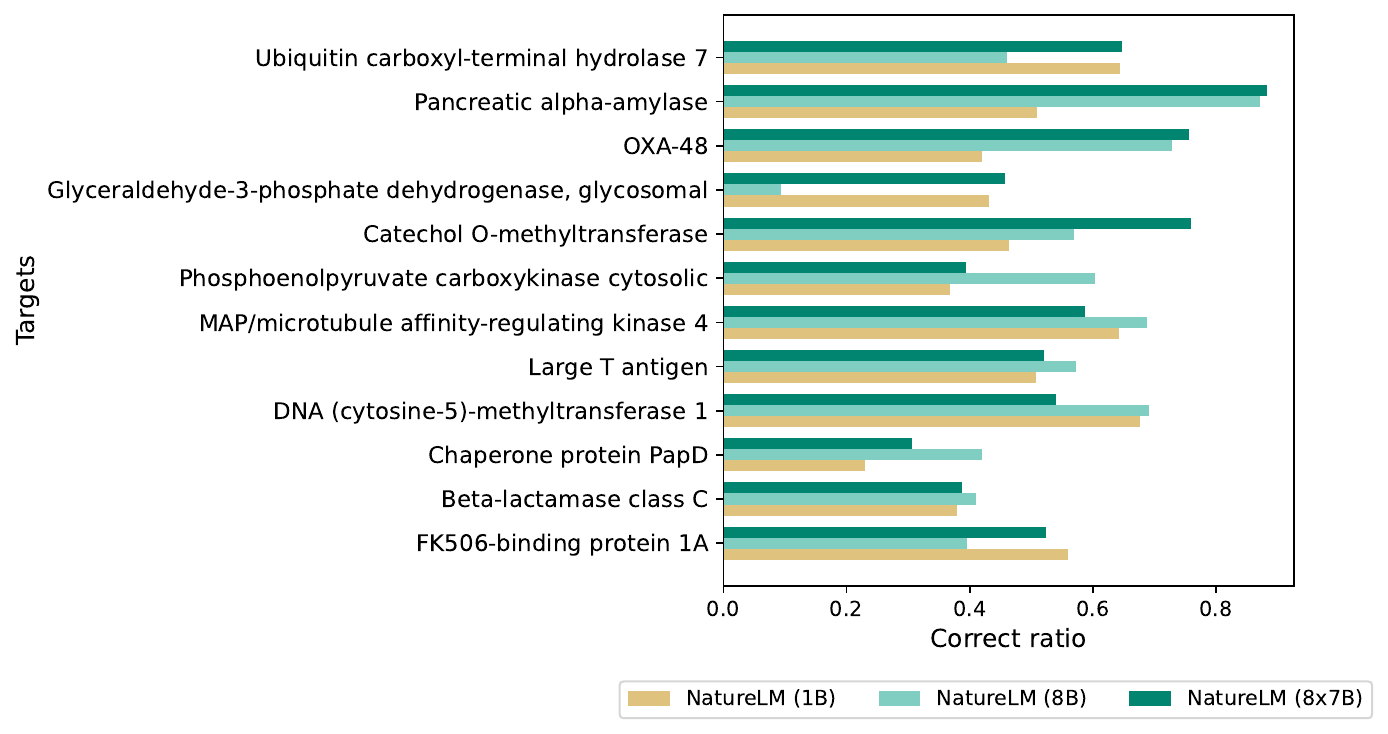} 
    \caption{Bar plot of the correct ratio of \ourM{} (1B), \ourM{} (8B) and \ourM{} (8x7B) on each target.}
    \label{fig:binding_correct}
\end{figure}

\begin{figure}
    \centering
    \subfigure[\ourM{} (1B)]{
    \includegraphics[width=0.45\linewidth]{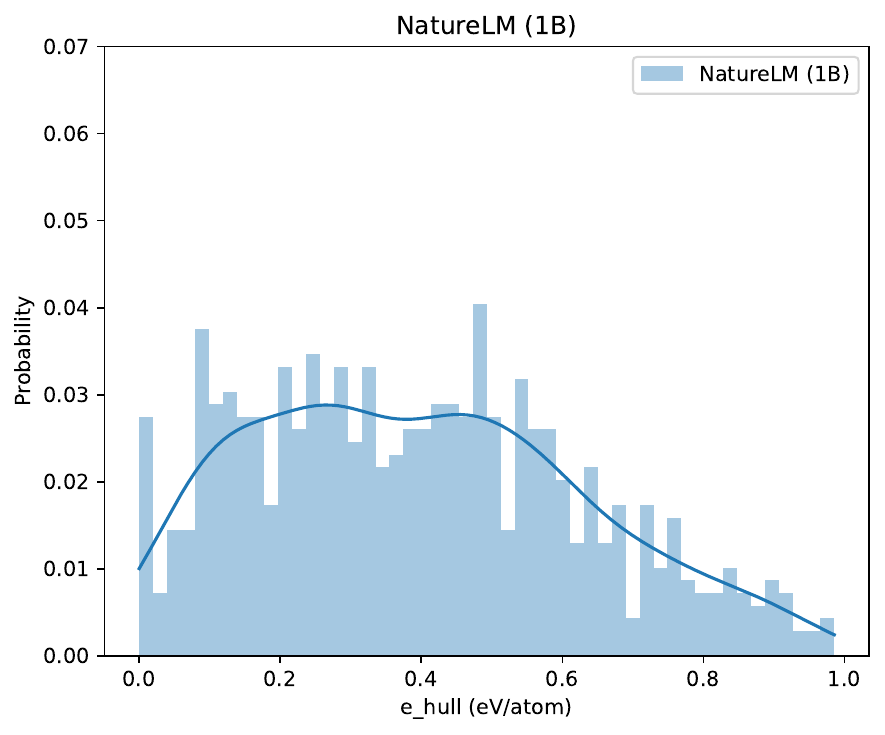}
    }%
    \subfigure[\ourM{} (8B)]{
    \includegraphics[width=0.45\linewidth]{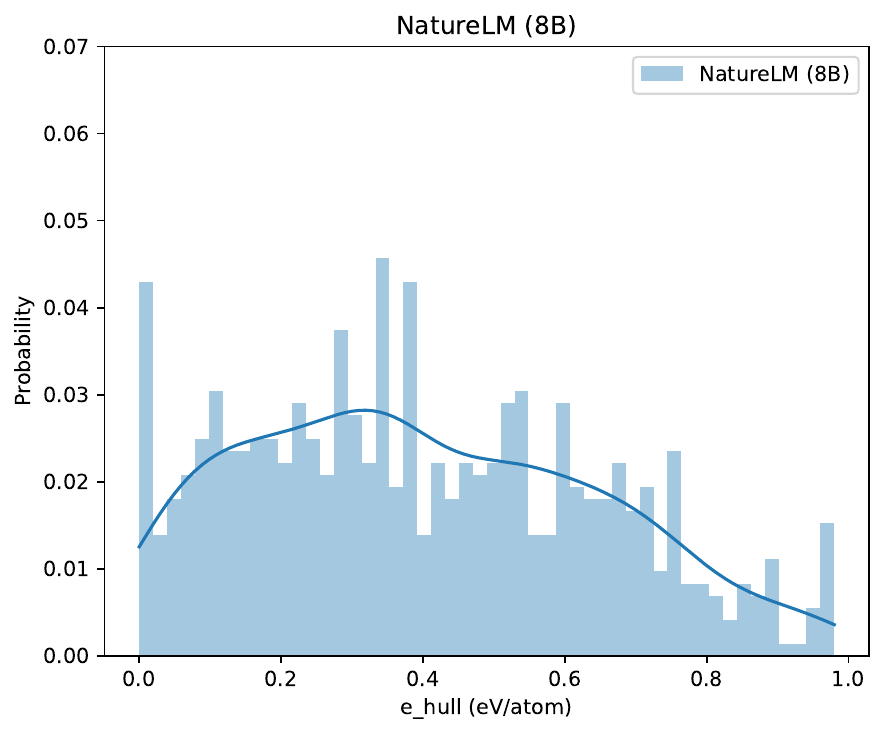}
    }%
    \vskip\baselineskip
    \subfigure[\ourM{} (8x7B)]{
    \includegraphics[width=0.45\linewidth]{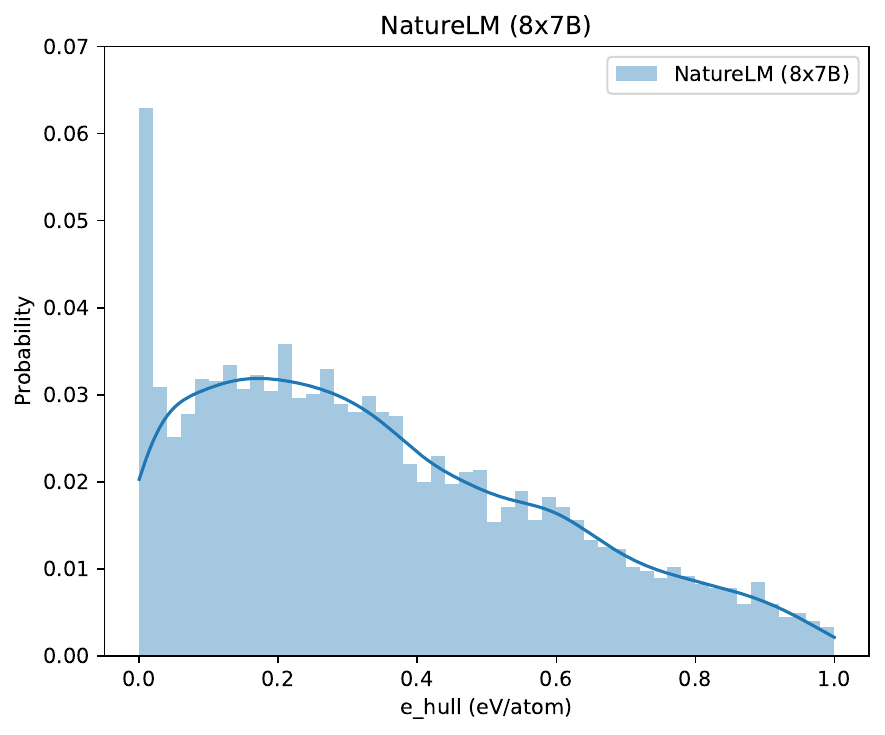}
    }
    \subfigure[Accumulated distribution]{
    \includegraphics[width=0.45\linewidth]{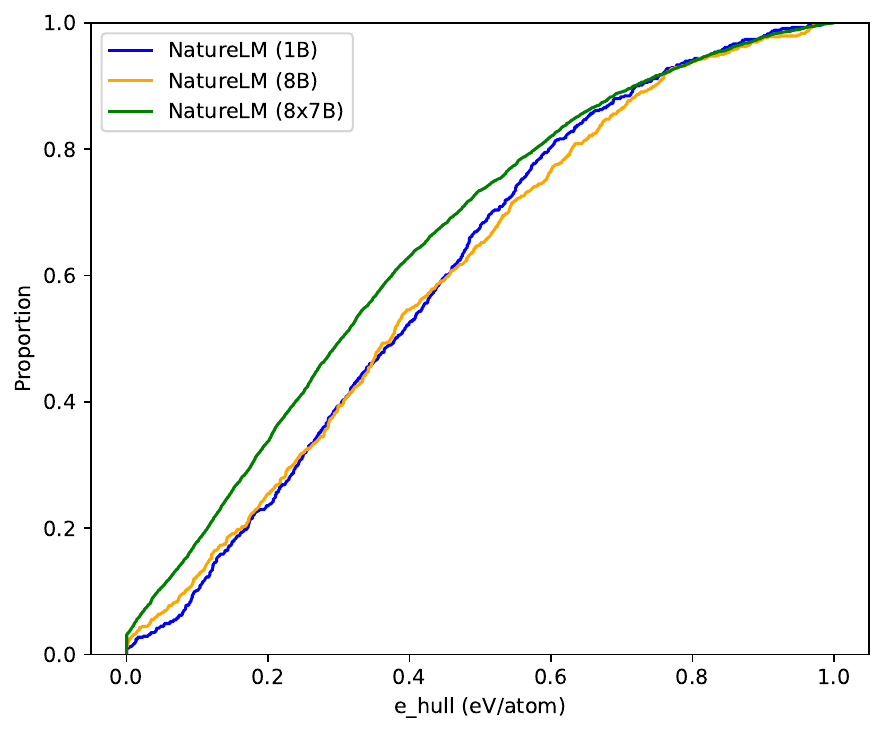}
    }
    \caption{Energy above hull (ehull) distribution for unconditional material generation.}
    \label{fig:mat_uncon}
\end{figure}

\begin{figure}
    \centering
    \subfigure[\ourM{} (1B)]{
    \includegraphics[width=0.45\linewidth]{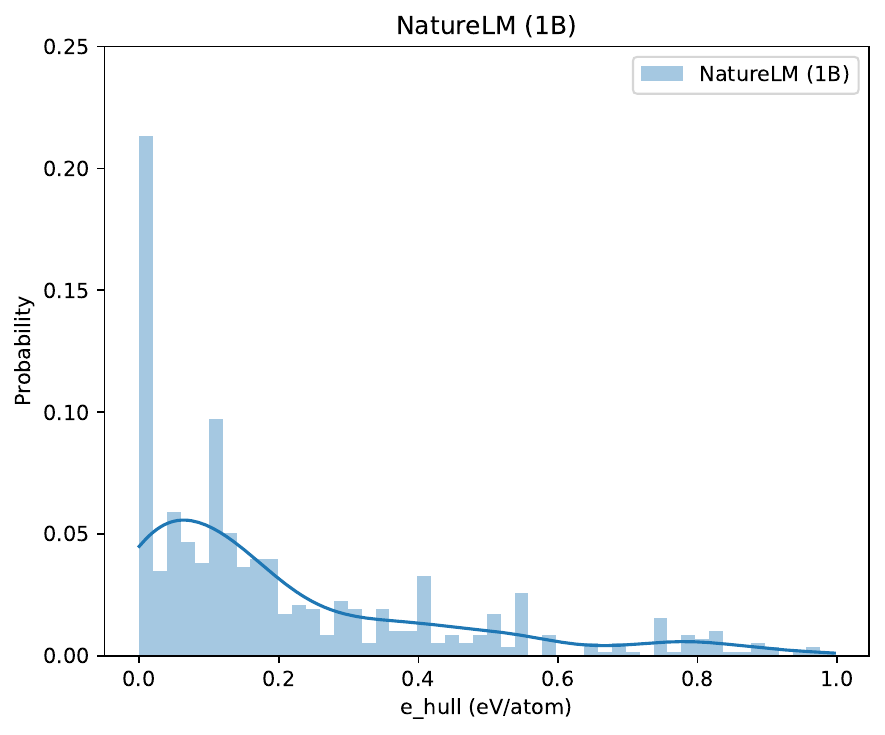}
    }%
    \subfigure[\ourM{} (8B)]{
    \includegraphics[width=0.45\linewidth]{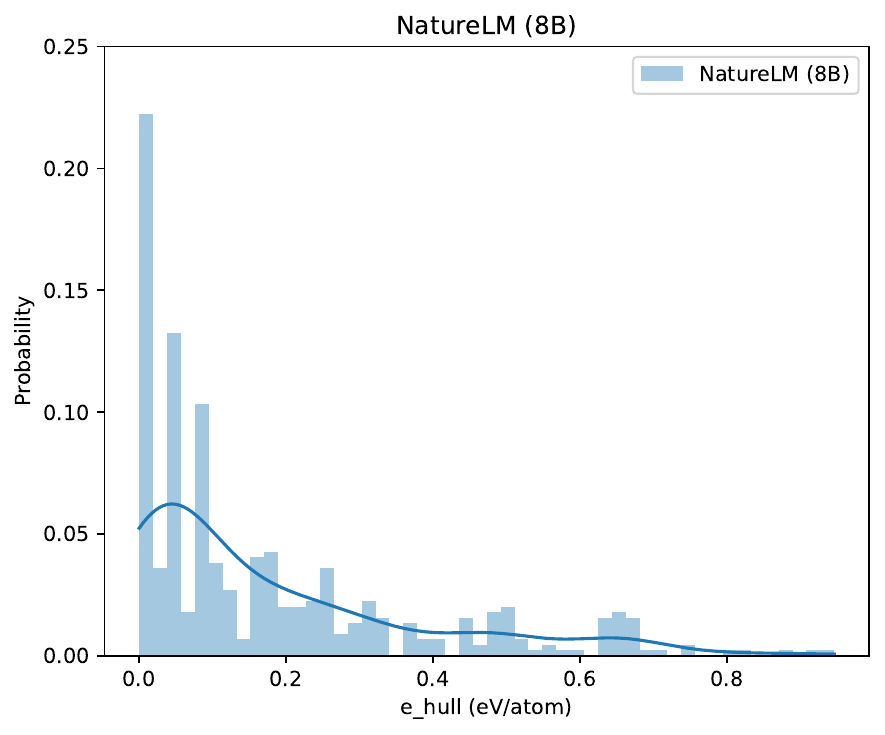}
    }%
    \vskip\baselineskip
    \subfigure[\ourM{} (8x7B)]{
    \includegraphics[width=0.45\linewidth]{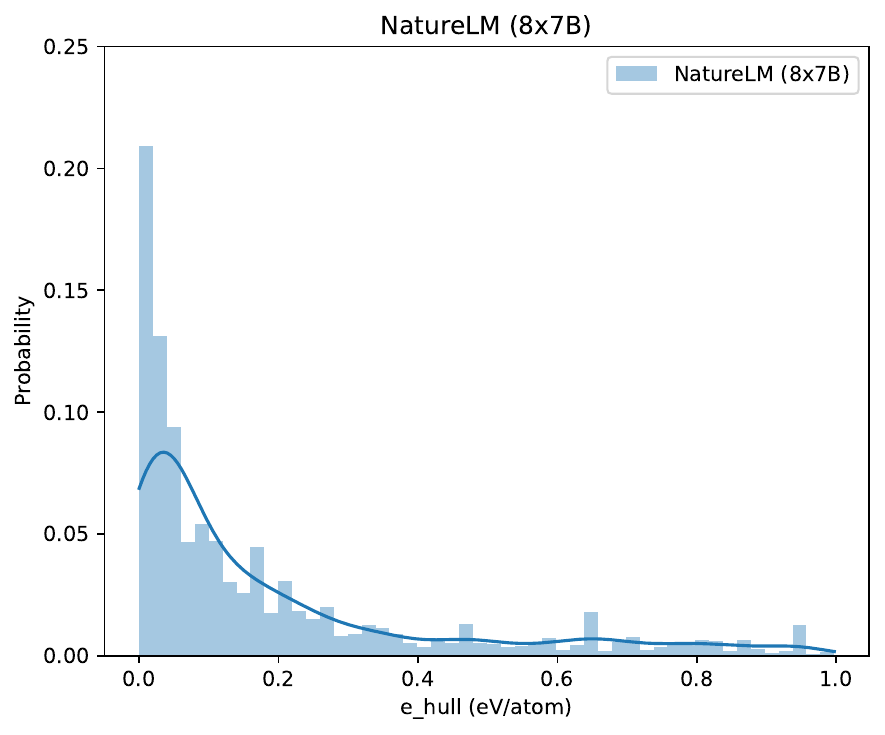}
    }
    \subfigure[Accumulated distribution]{
    \includegraphics[width=0.45\linewidth]{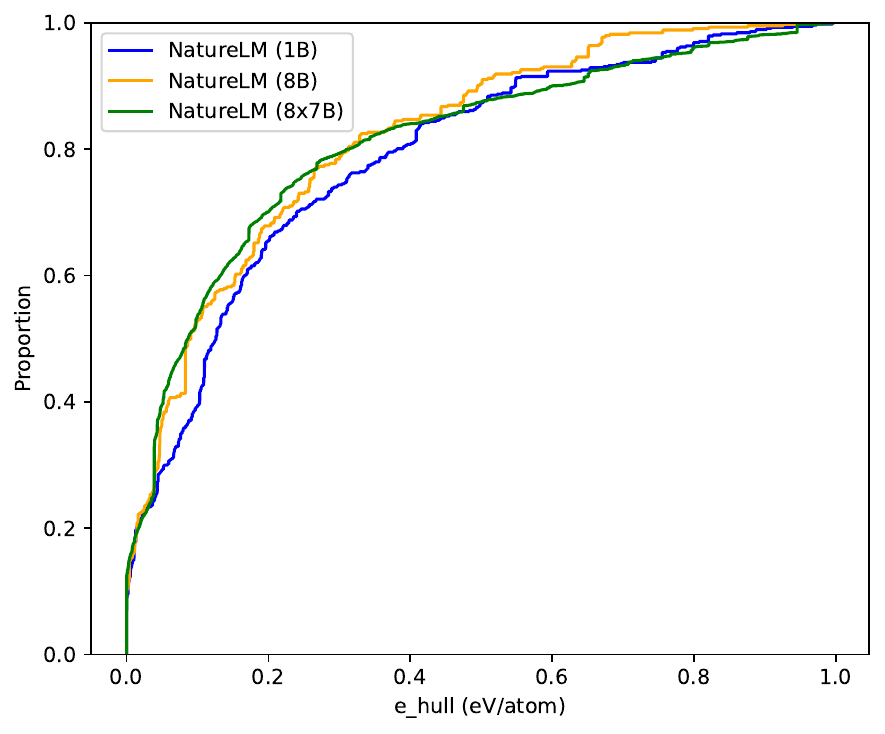}
    }
    \caption{Energy above hull (ehull) distribution for bulk modulus to material generation.}
    \label{fig:mat_bulk_to_mat_ehull}
\end{figure}

\begin{figure}[!htbp]
    \centering
    \includegraphics[width=0.8\linewidth]{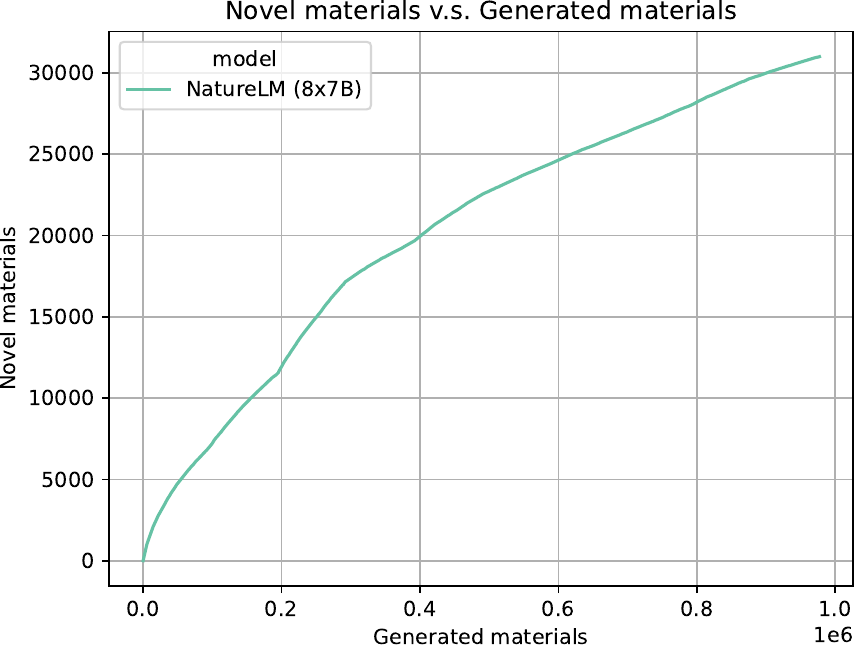}
    \caption{Novel materials w.r.t generated materials.}
    \label{fig:mat_novelty}
\end{figure}

\clearpage
\begin{figure}[!htbp]
\centering
\includegraphics[width=\linewidth]{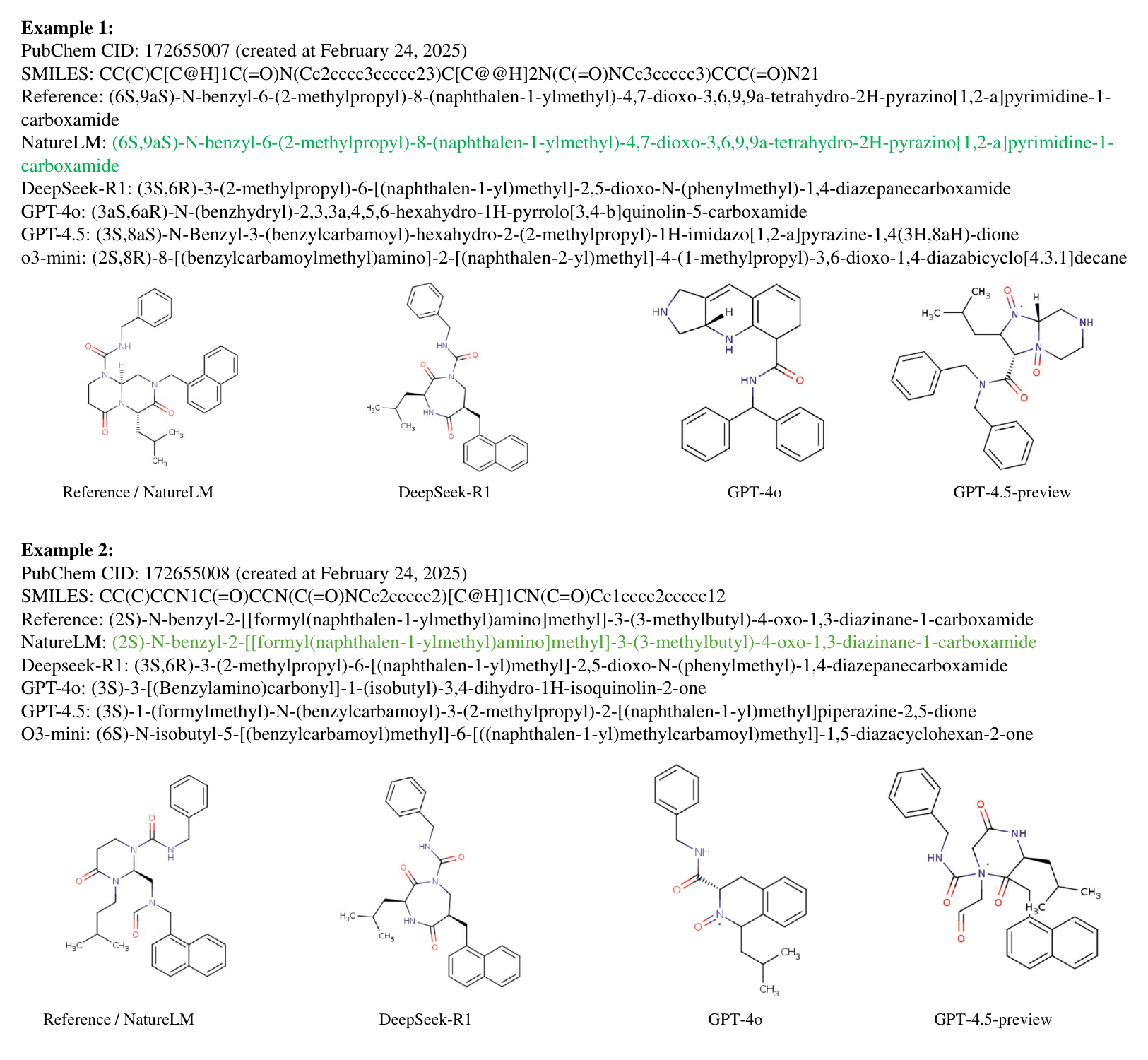}
\caption{We selected SMILES strings from PubChem with IDs 172655007 and 172655008, which were available as of February 24, 2025, and were excluded from our training set. The performance of \ourM{}, DeepSeek-R1 \cite{deepseekai2025r1}, GPT-4o, GPT-4.5-preview, and o3-mini was evaluated for SMILES-to-IUPAC translation. The generated IUPAC names are presented in the accompanying figure. These IUPAC names were subsequently converted back to SMILES for validation. The IUPAC name produced by o3-mini could not be processed due to the high structural complexity of the corresponding molecule. \ourM{} successfully generated the correct result. It is important to emphasize that our objective is not to criticize the limitations of general language models but to better understand their current capabilities and explore how they can be complemented by \ourM{} for enhanced performance. The molecular structures were visualized using the ChemDB Chemoinformatics Portal \cite{Chen2007-il} \url{https://cdb.ics.uci.edu/cgibin/Smi2DepictWeb.py}.}
\label{fig:case_study_iupac_to_smiles}
\end{figure}

\clearpage

\begin{figure}[!htbp]
\centering
\includegraphics[width=1.0\linewidth]
{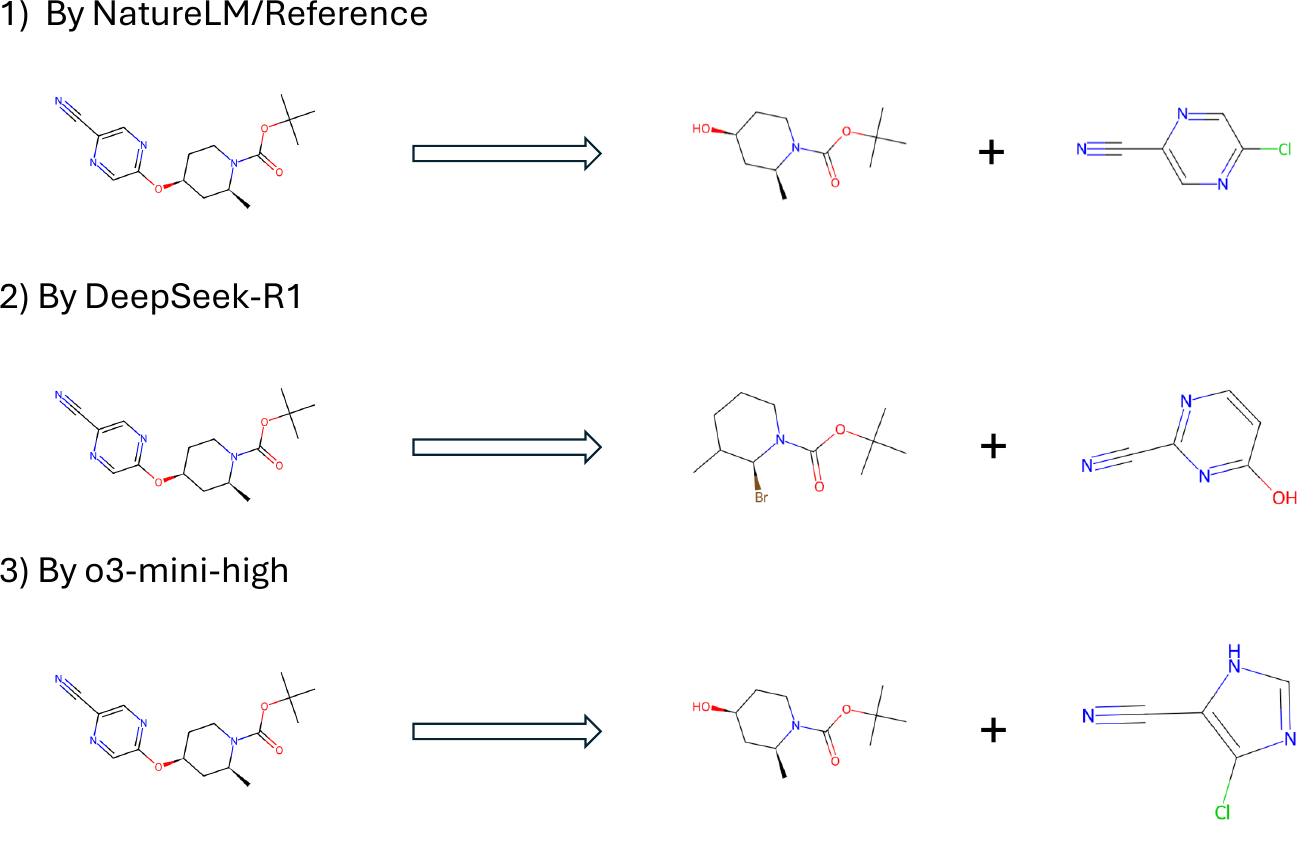}

\caption{Additional examples on retrosynthesis prediction. 
We evaluated the performance of \ourM{}, DeepSeek-R1, and o3-mini-high using a reaction from U.S. Patent ID US11999726B2, granted to Eli Lilly on June 04, 2024. 
The product features two ring systems with a protecting functional group, suggesting that the previous synthesis step likely involved a reaction to connect these rings. Notably, an ether bond links the two rings, with a pyrazine ring on one side and a piperidine ring on the other. Substitution on the pyrazine ring is a common strategy due to its electrophilicity, which often leads to substitution reactions. In this case, NatureLM accurately predicted the cleavage site of the molecule, incorporated a common chlorine atom on the pyrazine ring, and preserved the molecule's stereochemistry, providing a reasonable synthetic strategy. In contrast, both DeepSeek-R1 and o3-mini-high models correctly identified the reactive sites but failed to predict the correct reactants due to poor handling of SMILES representations. For instance, DeepSeek-R1 predicted the pyrazine as pyrimidine, altering the nitrogen atom's position, while o3-mini-high converted the six-membered pyrazine directly into a five-membered imidazole. These errors indicate that these general-purpose language models do not fully understand the relationship between chemical structures and their SMILES representations, hindering their ability to perform accurate reaction predictions.
}
\label{fig:case_study_reaction2}
\end{figure}

\clearpage

\begin{figure}[!htpb]
\centering
    \includegraphics[trim=1cm 0 3cm 0, clip, width=\linewidth]{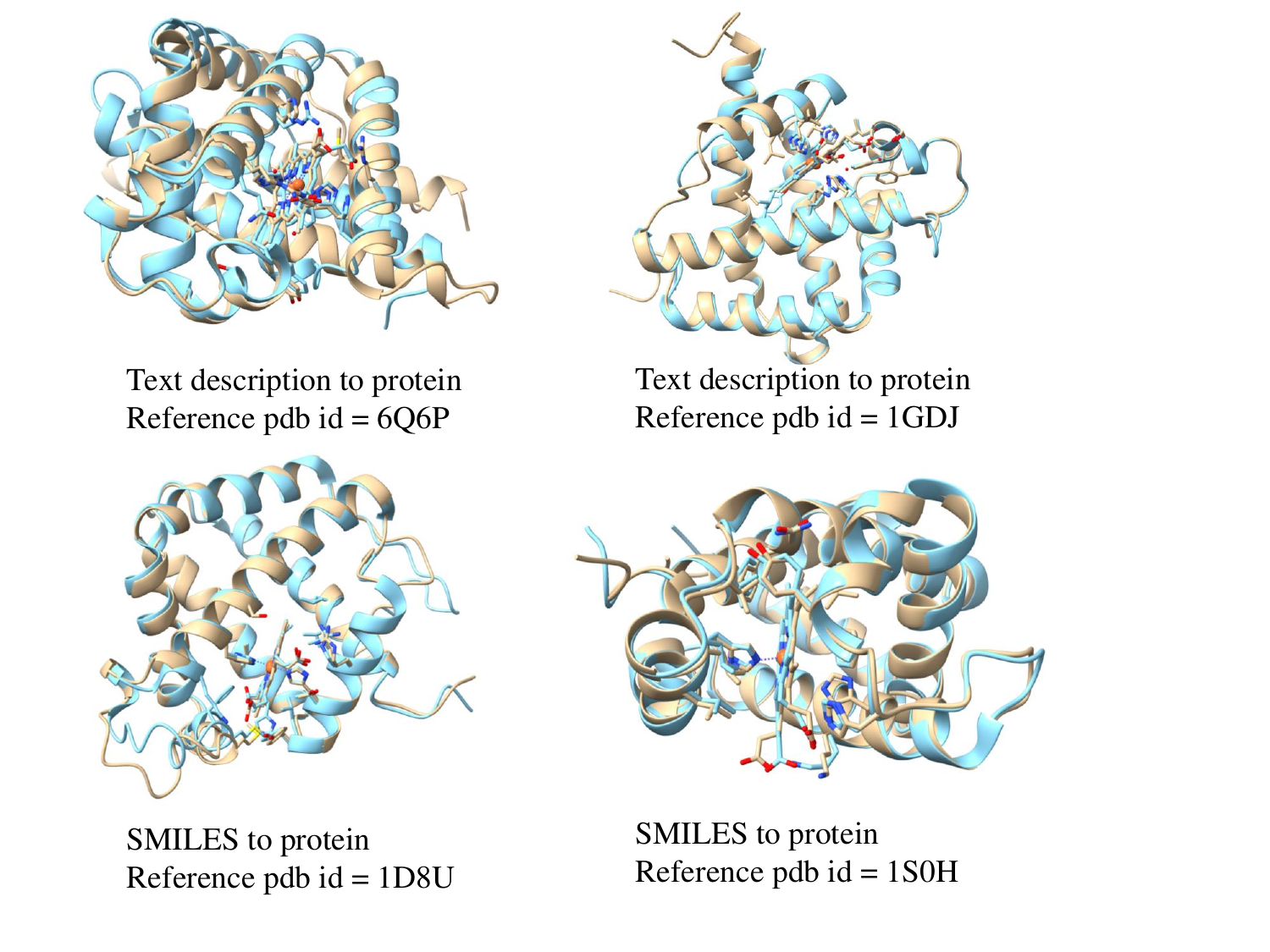}
    \caption{Additional examples of designing heme-binding proteins based on text or SMILES instructions are shown. The first two rows display results from the text-based design, while the second row corresponds to the SMILES-based design. The yellow models represent structures generated by \ourM{}, whereas the blue models are the reference structures retrieved using the built-in Chimera function. The structures of the generated proteins were predicted using Protenix \cite{Protenix2025}. }
    \label{fig:SI:moreHemeCases}
\end{figure}

\clearpage

\begin{figure}[!htpb]
\centering
\includegraphics[width=0.6\linewidth]{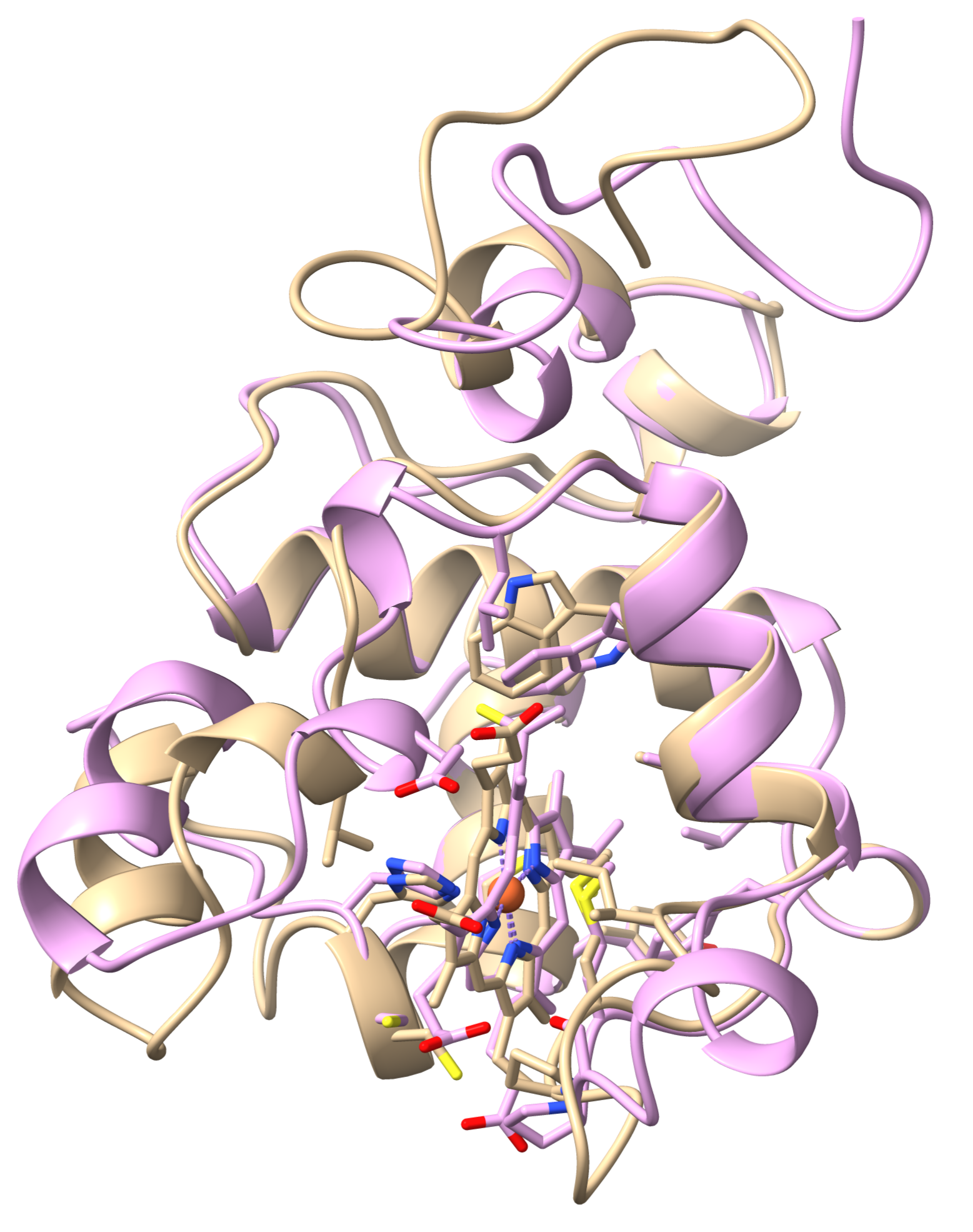}
\caption{Comparison of the complex structure of the generated protein with heme (yellow model) and heme C (pink model). The protein was obtained using the SMILES-to-protein approach described in Section \ref{sec:heme_case_study}. We observe that they share common structural features. The structures of the generated proteins were predicted using Protenix \cite{Protenix2025}. \\
For the retrieved PDB structure 3MK7 in Fig. \ref{fig:heme_bind_prot}, our generated protein aligns to the pocket region that binds to heme C. To further validate this, we used Protenix to predict the binding of our generated protein to both heme C (PubChem CID: 11987638) and heme. The results demonstrate that heme C fits properly into the designed pocket, supporting the structural compatibility of the generated protein with heme C.\\
This discrepancy arises from the high structural similarity between heme and heme C, as their SMILES representations are nearly identical. Despite this slight misalignment, the output remains biologically relevant because heme-binding proteins often interact with multiple heme derivatives. Furthermore, generating a protein that binds to heme C from the SMILES of heme highlights the algorithm's ability to capture the inherent structural flexibility and functional overlap within the heme family. We will continue improving the algorithm to enhance ligand specificity in future iterations.}
\label{fig:SI:prot_hem_hec}
\end{figure}

\clearpage
\begin{figure}[!htpb]
    \centering
    \includegraphics[trim=5cm 0.5cm 8cm 2cm, clip, width=\linewidth]{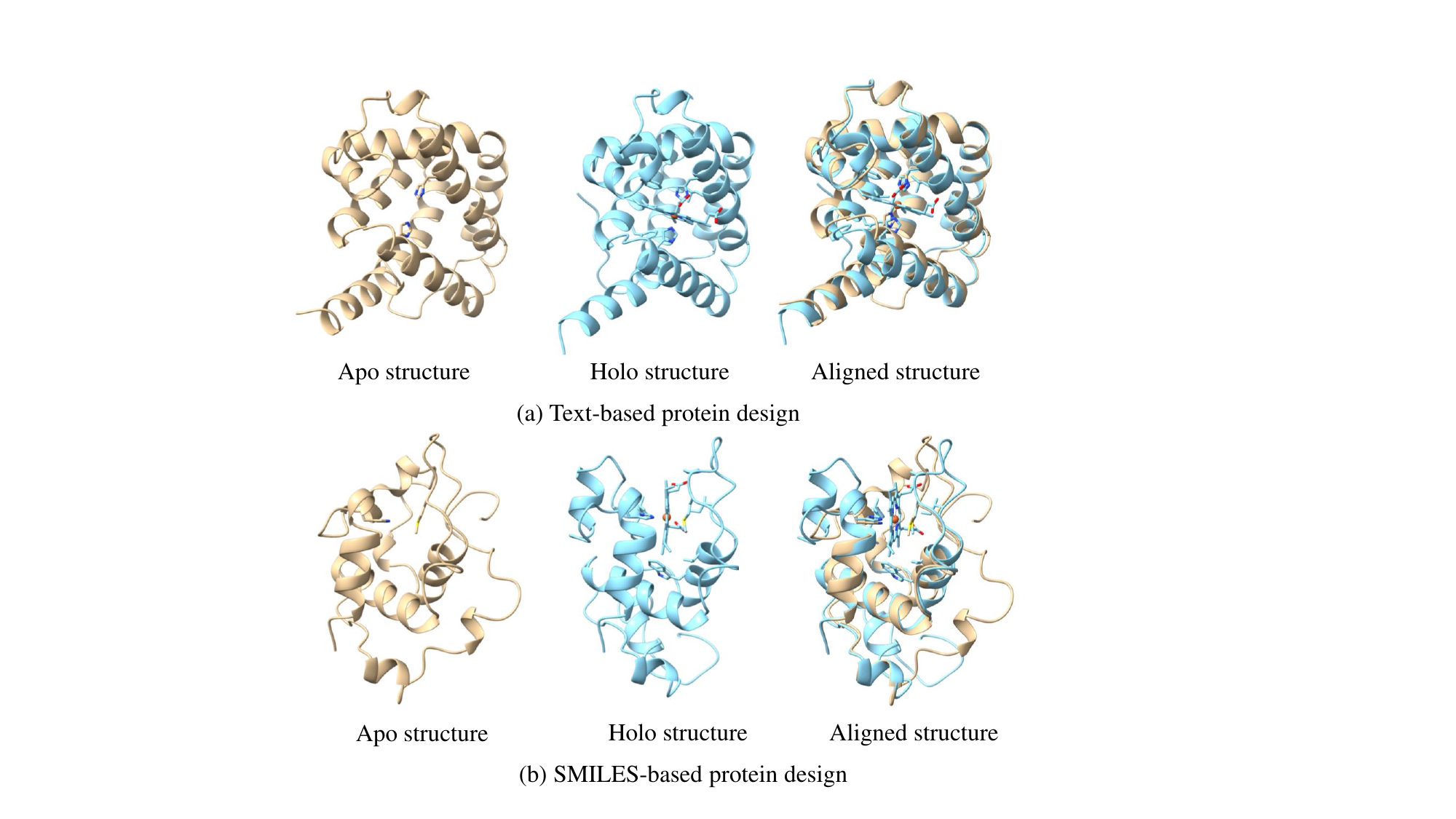}
    \caption{Comparison of the apo structure of the generated protein, the holo structure in complex with heme, and their aligned structures. Key residues, such as histidine and methionine, occupy similar positions in the pocket region in both the apo and holo structures. This observation suggests that the generated proteins are not only capable of binding heme but also exhibit a structurally pre-formed or conserved binding pocket even in the absence of the ligand. These findings validate the structural plausibility of the designed proteins and their suitability for heme binding.}
    \label{fig:compare_apo_holo}
\end{figure}

\clearpage

\begin{figure}[!h]
\centering
\includegraphics[width=0.75\linewidth]{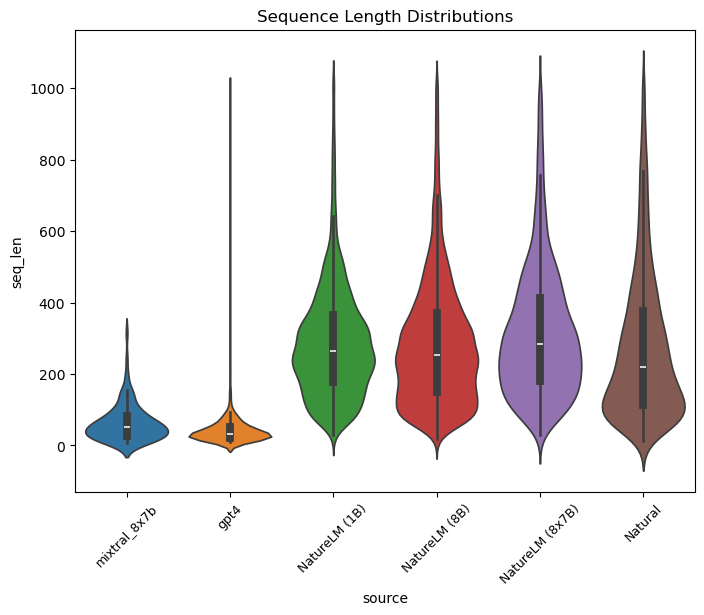}
\caption{Sequence length distribution of generated proteins. The \ourM{} models demonstrate a more natural distribution that closely resembles the reference UR50 sequences, while Mixtral 8x7B and GPT-4 tend to generate shorter sequences.}
\label{fig:protein:unconditioned_generation_sequence_length}
\end{figure}

\begin{figure}
\centering
\includegraphics[width=\linewidth]{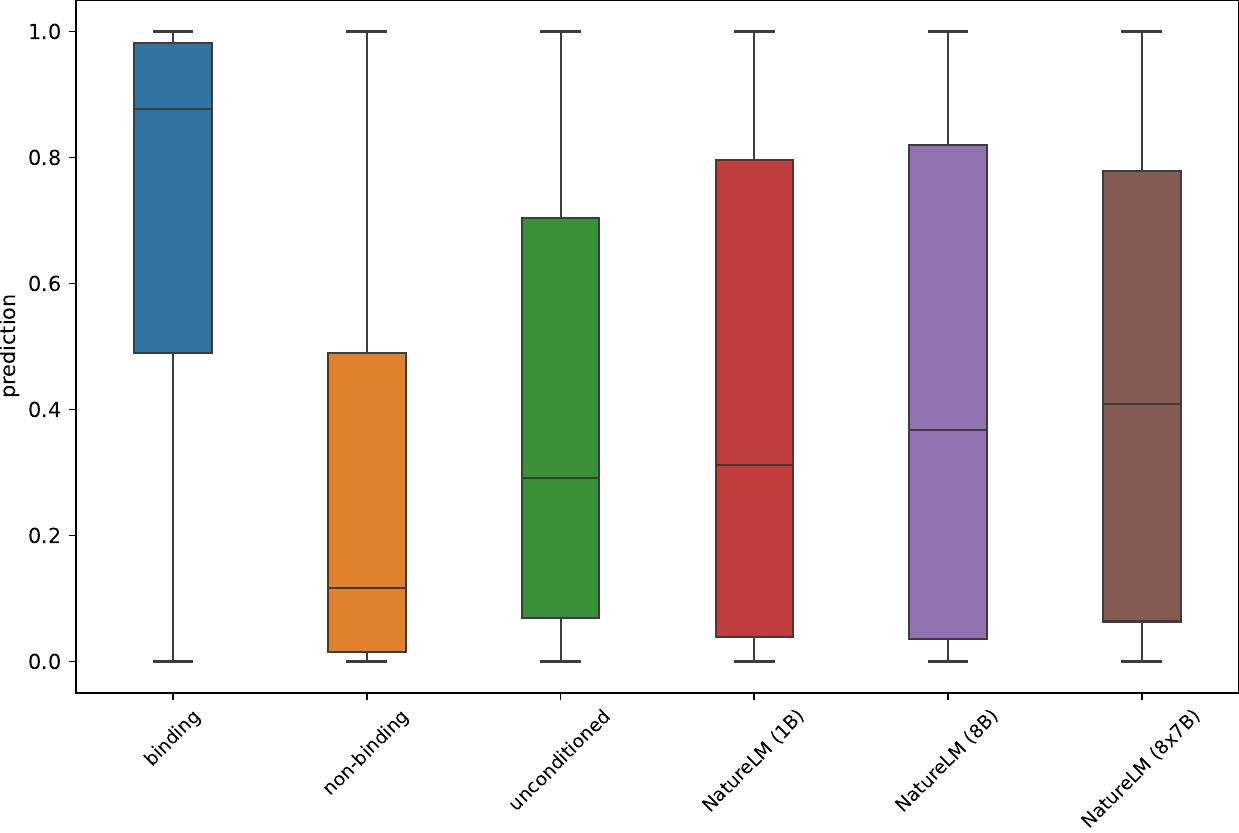}
\caption{The distribution of the predicted scores for the RNA sequences in the test set and the generated RNA sequences shows a clear trend. In terms of median values, larger models consistently achieve better predicted scores, indicating stronger binding affinity.}
\label{fig:enter-label}
\end{figure}

\clearpage

\begin{figure}[h]  
    \centering  
    \begin{mdframed}[backgroundcolor=white, linecolor=black, linewidth=1pt]  
    \textsc{Stable}
    \textit{
    \begin{itemize} 
        \item[-] Please produce a protein sequence that exhibits stability.
        \item[-] I require a stable protein sequence, kindly generate one.
        \item[-] Generate a protein sequence ensuring its stability.
        \item[-] I need a protein sequence that's stable. Please generate it.
        \item[-] Create a stable protein sequence.
        \item[-] Produce a stable protein sequence.
        \item[-] Kindly generate a protein sequence with stability.
        \item[-] I would like you to generate a stable protein sequence.
        \item[-] Please create a protein sequence that ensures stability.
        \item[-] Make a protein sequence that is stable.
    \end{itemize}
    }
    \textsc{Soluble}
    \textit{
    \begin{itemize} 
        \item[-] Generate a soluble protein sequence.
        \item[-] Produce a protein sequence that is soluble.
        \item[-] Create a soluble protein sequence, please.
        \item[-] I require a soluble protein sequence, kindly generate one.
        \item[-] Please produce a protein sequence that exhibits solubility.
        \item[-] Make a protein sequence that is soluble.
        \item[-] Kindly generate a protein sequence with solubility.
        \item[-] I need a protein sequence that's soluble. Please generate it.
        \item[-] I would like you to generate a soluble protein sequence.
        \item[-] Please create a protein sequence that ensures solubility.
    \end{itemize}
    }
    \textsc{Insoluble}
    \textit{
    \begin{itemize}
        \item[-] Generate an insoluble protein sequence.
        \item[-] Produce a protein sequence that is not soluble.
        \item[-] Create an insoluble protein sequence, please.
        \item[-] I require an insoluble protein sequence, kindly generate one.
        \item[-] Please produce a protein sequence that lacks solubility.
        \item[-] Make a protein sequence that is not soluble.
        \item[-] Kindly generate a protein sequence without solubility.
        \item[-] I need a protein sequence that's insoluble. Please generate it.
        \item[-] I would like you to generate an insoluble protein sequence.
        \item[-] Please create a protein sequence that ensures insolubility.
    \end{itemize}
    }
    \end{mdframed}  
    \caption{Prompts for conditioned generation} \label{fig:protein:conditioned_prompts_full}
\end{figure} 

\clearpage

\begin{mdframed}[backgroundcolor=white, linecolor=black, linewidth=1pt]
    \textbf{Instruction:} \textit{Tell me about Cas9 regarding its cutting mechanism?} \\
    \textbf{Response:} \\
    Cas9 is a CRISPR-associated protein that can be programmed to cut DNA at a specific location.  It recognizes a short DNA sequence, called a protospacer, and cuts both strands of the DNA at that site. This DNA cutting is done using two nuclease domains, HNH and RuvC, which are repsonsible for cutting the top and bottom strands of the DNA, respectively. The HNH nuclease domain cleaves the top strand, while the RuvC nuclease domain cleaves the bottom strand. The cut made by the nuclease domains leaves the DNA with a sticky end that can ligate to another DNA molecule, such as the chromosome of the cell.  \\

    \textbf{Instruction:} \textit{Could you provide me the advantages and disadvantages of Cas12a, another widely used CRISPR protein?} \\
    \textbf{Response:} \\
    Advantages: \\
    1. High specificty: Cas12a has higher specificity compared to Cas 9, which reduces off-target effects. \\
    2. Simpler PAM requirements: Cas 12a requires a simpler PAM sequence than Cas9, making it easier to target specific genomic loci. \\
    3. Multiplexing: Cas12a can be used for multiplexing genome editing, meaning multiple genomic loci can be edited in a single reaction. \\
    Disadvantages: \\
    4. Lower efficiency: Cas12a is less efficient in terms of indel formation than Cas9. \\
    5. Larger size: Cas12a is a larger protein than Cas9, which can make it harder to deliver the Cas12a system into cells. \\
    6. Lack of donor specificity: Cas12a has a lower efficiency of HDR than Cas9, which makes it difficult to target specific nucleotides for insertion or deletion.
    7. Higher toxicity: Cas12a may casue higher levels of cell death than Cas9, which could impact the overall efficiency of genome editing. \\
    8. Differing DNA repair mechanism: Cas12a and Cas9 utilize DNA different repair mechanisms, which can lead to different outcomes when targeting specific genomic loci. \\
    9. Less versatile: Cas12a is not as versatile as Cas9, as it can only target DNA sequences that contain a TTN PAM sequence. \\
    10. Unclear off-target effects: Cas12a has not been fully characterized in terms of off-target effects, which could lead to unwanted changes in the genome. \\
    11. Difficulty in designing gRNAs: Cas12 requires a longer gRNA sequence compared to Cas9, making it more difficult to design effective gRNAs. \\
\end{mdframed}  
\captionof{figure}{Response of CRISPR gene editing.}\label{SI_fig:case_study:CRISPR}

\begin{mdframed}[backgroundcolor=white, linecolor=black, linewidth=1pt]  
    \noindent
    \textbf{Instruction:} \textit{PD-L1, Programmed Death Ligand-1, is a protein that expressed on cancer cells.  Design a peptide with 10 amino acids that binds to PD-L1.} \\ 
    \textbf{Response:} 
        \text{$\langle$protein$\rangle$}\text{TDEDVLQKIVKRATEDSSYTVKLSNGQTLSVTE} \\
        \text{DSEIKKIQKFRS}\text{$\langle$/protein$\rangle$} \\
\end{mdframed}  
\captionof{figure}{Response of cancer immunotherapy.} \label{SI_fig:case_study:cancer_immunotherapy}

\begin{mdframed}[backgroundcolor=white, linecolor=black, linewidth=1pt]
    \textbf{Instruction:} \textit{Could you explain Lipinski's Rule of Five and its role in evaluating the drug-likeness of chemical compounds?} \\
    \textbf{Response:} Lipinski's Rule of Five, also known as the Rule of Five, is a set ofules developed by Dr. Lipinski that outlines the physical properties of successful orally active drugs. It is a set of simple guidelines that can be used to evaluate the drug-likeness of chemical compounds. It indicates whether a chemical compound with a certain set of properties will have desirable drug-like properties and is likely to be successful as an oral drug. The fiveules that comprise Lipinski's Rule of Five are:
1. The number of hydrogen bond donors should be limited to no more than 
2. The number ofotatable bonds should be limited to no more than 
3. The molecular weight should be less than 500 daltons.
4. The number of hydrogen bond acceptors should be limited to no more than 10.
5. The logD value (a logarithmic measure of a compound's hydrophobicity) should be less than 3.
The application of Lipinski's Rule of Five has been widely adopted in the pharmaceutical industry to prioritize lead compounds for drug development. It is a tool that is used to identify potentially promising compounds and can help in the elimination of compounds with poor pharmaceutical properties.\\
\end{mdframed}  
\captionof{figure}{Response of drug discovery.} \label{SI_fig:case_study:drug_discovery}

\clearpage

\begin{mdframed}[backgroundcolor=white, linecolor=black, linewidth=1pt]  
    \textbf{Instruction:} \textit{How do I wrap a present neatly?} \\
    \textbf{\ourM{}:} To wrap a present neatly, start by deciding on the type of wrapping paper you want to use. Next, tear the edges of the paper to make it easier to fold. Wrap the paper around the gift and use tape to hold it in place. Fold the ends of the paper over to form a neat tuck. Finally, tie a bow around the gift to finish it off.\\
    \textbf{Mixtral:} To wrap a present neatly, follow these steps: 1. Gather your supplies: wrapping paper, scissors, tape, and a ruler or measuring tape. 2. Measure the length and width of the present, and cut a piece of wrapping paper that is at least 2 inches longer and wider than the present. 3. Place the present in the center of the wrapping paper, with the bottom of the present facing up. 4. Fold one side of the wrapping paper over the present, and tape it in place. 5. Fold the other side of the wrapping paper over the present, and tape it in place. 6. Fold the top and bottom edges of the wrapping paper over the present, and tape them in place.7. Flip the present over, and repeat steps 4-6 on the other side. 8. Add a ribbon or bow to the top of the present for a finishing touch.

    \textbf{Instruction:} \textit{Write a captivating movie review for a recently released science fiction film, discussing its plot, characters, and special effects.} \\
    \textbf{\ourM{}:} It's still quite worth seeing.\\
    \textbf{Mixtral:} Captivating Movie Review: "Interstellar" "Interstellar" is a science fiction film directed by Christopher Nolan, starring Matthew McConaughey, Anne Hathaway, and Jessica Chastain. The film follows a group of astronauts as they travel through a wormhole in search of a new home for humanity. The plot of "Interstellar" is complex and thought-provoking, exploring themes of time, space, and the human condition. The characters are well-developed and relatable, with McConaughey delivering a powerful performance as the lead astronaut, Cooper. The special effects in "Interstellar" are breathtaking, with stunning visuals of space and otherworldly landscapes. The film also features a haunting score by Hans Zimmer, which adds to the emotional impact of the story. Overall, "Interstellar" is a captivating and thought-provoking film that will leave audiences pondering the mysteries of the universe long after the credits roll.
    
\end{mdframed}  
\captionof{figure}{Response of AlpacaEval.} \label{fig:case_study:alpacaeval}

\clearpage
\section{Supplementary tables}

\begin{table}[!htbp]
\centering
\begin{tabular}{lcccc}
\toprule
& Samples   & Tokens  & Samples & Tokens \\
Data source & (by million) &  (by billion) &  (\%)& (\%)\\
\midrule
Text           & 46.3  & 14.4 & 11.6 & 10.0\\
Small molecule & 68.0  &  4.2 & 17.0 & 2.9 \\
Protein        & 192.0 & 65.2 & 47.9 & 45.3 \\
DNA            & 13.4  & 19.8 & 3.3  & 13.8\\
RNA            & 37.8  & 27.5 & 9.4  & 19.1\\
Material       & 1.1   & 0.02 & 0.3  & 0.014\\
Cross-domain & 41.9  & 12.7 & 10.5 & 8.8\\
\midrule
Total & 400.5& 143.8 & 100 & 100 \\
\bottomrule
\end{tabular}
\caption{Tokens numbers and their distribution of each domain. }
\label{tab:statistics_pretrain_data}
\end{table}

\begin{table}[!htbp]
\centering
\begin{tabular}{cccccccc}
\toprule
Model Parameters & 1B & 8B & 8x7B \\
\midrule
Learning Rate & 1e-4 & 1e-4 & 2e-4 \\
Batch Size (Sentences) & 4096 & 2048 & 1536 \\
Context Length (Tokens) & 8192 & 8192 & 8192 \\
GPU number (H100) & 64 & 256 & 256 \\
\bottomrule
\end{tabular}
\caption{Training recipe of different models.}
\label{tab:training_recipe}
\end{table}

\begin{table}[!htbp]
\centering
\begin{tabular}{lcc}
\toprule
Porperty & Value \\
\midrule
QED & 0.5, 0.6, 0.7, 0.8, 0.9, 1.0\\
HBA & 0, 1, 2, 3, 4, 5, 6, 7, 8, 9, 10\\
HBD & 0, 1, 2, 3, 4, 5\\
FSP3 & 0.0, 0.1, 0.2, 0.3, 0.4, 0.5, 0.6, 0.7, 0.8, 0.9, 1.0\\
RotBonds & 0, 1, 2, 3, 4, 5, 6, 7, 8, 9, 10\\
TPSA & 20, 40, 60, 80, 100, 120\\
\bottomrule
\end{tabular}
\caption{Input property values for property-to-molecule generation}
\label{tab:property_values}
\end{table}

\begin{table}[!htbp]
\centering
\begin{tabular}{lc}
\toprule
Target & Spearman correlation\\
\midrule
Pancreatic alpha-amylase &0.569\\
Large T antigen &0.572\\
DNA (cytosine-5)-methyltransferase 1 &0.517\\
Chaperone protein PapD &0.739\\
Catechol O-methyltransferase &0.638\\
Glyceraldehyde-3-phosphate dehydrogenase, glycosomal &0.503\\
Phosphoenolpyruvate carboxykinase cytosolic &0.501\\
FK506-binding protein 1A &0.606\\
Beta-lactamase class C &0.560\\
OXA-48 &0.680\\
Ubiquitin carboxyl-terminal hydrolase 7 &0.764\\
MAP/microtubule affinity-regulating kinase 4 &0.782\\
\bottomrule
\end{tabular}
\caption{Spearman correlation between docking scores and binding affinity on the selected targets for evaluation.}
\label{tab:targets}
\end{table}

\begin{table}[]
\centering
\begin{tabular}{lcccccccccc}
\toprule
Basic property & QED & QED & donor & donor & LogP & LogP \\
Enzyme & CYP2C9 & CYP3A4  & CYP2C9 & CYP3A4 & CYP2C9 & CYP3A4 & Average \\
\midrule
1B & 0.352 & 0.357 & 0.501 & 0.497 & 0.276 & 0.280 & 0.377 \\
8B & 0.404 & 0.428 & 0.548 & 0.522 & 0.332 & 0.340 & 0.429 \\
8x7B & 0.429 & 0.427 & 0.515 & 0.501 & 0.355 & 0.347 & 0.429 \\
\bottomrule
\end{tabular}
\caption{Joint optimization of metabolism and a basic property.}
\label{tab:joint_basic_cyp}
\end{table}

\clearpage
\begin{table}[b]
\centering
\begin{tabular}{ ccc }
\toprule
Property Name & Training samples & Testing samples \\ 
\midrule
BBBP          & 1272             & 199             \\  
BACE          & 90677            & 152             \\  
LogP          & 8491             & 473             \\  
Donor         & 8526             & 478             \\  
QED           & 8466             & 476             \\  
CYP1A2        & 8076             & 103             \\  
CYP2C9        & 21589            & 199             \\  
CYP2D6        & 8067             & 165             \\  
CYP3A4        & 24376            & 171             \\ 
\midrule
Total         & 179540           & 2416            \\ 
\bottomrule
\end{tabular}
\caption{Statistics of preference data used in RLHF}
\label{tab:data-rlhf}
\end{table}

\clearpage
\section{Supplementary notes}

\subsection{Text-guided basic property optimization of small molecule compounds}
We focus on optimizing the basic molecular properties in this section. The input of \ourM{} includes a text command and a SMILES sequence to be optimized.  We evaluate the optimization results of Quantitative Estimation of Drug-likeness (QED), LogP, and the number of hydrogen bond donors. Following DrugAssist \cite{ye2023drugassist}, we curated a fine-grained procedure. An illustrative example is provided below and the example is from DrugAssist \cite{ye2023drugassist}:

\begin{example}
\noindent\texttt{Instruction: With a molecule represented by the SMILES string }
\newline
\mol{}CC(N)=[NH+]CC(=O)N1CCC(O)(Cn2cnc3c(cnn3-c3ccc(N4CCC5(CCOCC5)CC4)cc3)c2=O)CC1\emol{}, \texttt{propose adjustments that can increase its QED value by at least 0.1 compared to the pre-optimized value to make it more drug-like. }   
\newline
\texttt{Response:}\mol{}CC(C)(C)OC(=O)N1CCC(c2ncc(-c3ccc(CC[B-](F)(F)F)cc3)cn2)CC1\emol{}.
\end{example}

For QED and hydrogen bond donor property optimization, our instructions cover the following scenarios: (i) increase or decrease the property by $\delta$, where both $\delta=0$ and $\delta>0$ are considered, aiming to verify the ability of the model; (2) maintain the properties. For LogP, the instruction is to adjust the LogP value from one specified region to another. 

\begin{table}[!htbp]
\centering
\begin{tabular}{lccc}
\toprule
Model            & QED     & \#Donor & LogP \\
\midrule 
LLAMA 3 8B$^*$   & 0.62 / 0.43    &  0.75 / 0.43   & 0.84 / 0.45   \\ 
\ourM{} (1B)    & 0.58 / 0.57    &  0.74 / 0.58   & 0.63 / 0.60 \\ 
\ourM{} (8B)         & 0.65 / 0.45    &  0.81 / 0.44   & 0.80 / 0.42 \\ 
\ourM{} (8x7B) & 0.66 / 0.48 & 0.80 / 0.47  & 0.80 / 0.47  \\ 
\bottomrule
\end{tabular}
\caption{Comparison between the basic property optimization. In each cell, the success rate and uniqueness ratio are reported.}
\label{tab:basic_property_optimization}
\end{table}

The results are in Table \ref{tab:basic_property_optimization}. Notably, as the model size of \ourM{} increases, there is a marked improvement in performance metrics across all properties. For instance, \ourM{} (8B) surpasses \ourM{} (1B) in all categories, indicating enhanced comprehension and manipulation of molecular structures and properties as model complexity grows. Despite DrugAssist$^*$ achieving the highest scores overall, our results demonstrate that by further increasing the model size and fine-tuning the training process, there is significant potential to outperform this baseline. The trend observed with the \ourM{} models underscores the importance of model scale and suggests that with continued advancements in model architecture and training methodologies, even better optimization outcomes can be achieved. This validates the proficiency of \ourM{} in understanding and applying the given instructions to revise molecular properties accordingly.

\subsection{Supplementary information of RNA generation}\label{app:rna_generation}

Minimum free energy (MFE) calculation: 

\texttt{./ViennaRNA-2.7.0/src/bin/RNAfold -p --MEA \$\{input\_file\}}

Usage of cmscan: 

\texttt{cmscan --rfam --cut\_ga --nohmmonly --tblout results\_tblout --fmt 2 --clanin Rfam/Rfam.clanin Rfam/Rfam.cm \$\{input\_file\}}

\subsection{POSCAR files of crystal structures in Fig. \ref{fig:bulk_caseStudy}}
{{
\footnotesize
\begin{example}
\begin{verbatim}
VASPStructure of Re3C
 1.000000
    7.1831247561033589    0.0000000000000000    0.0000000000000000
    0.0000000000000000    1.4245311887791383    2.4673932588460490
    0.0000000000000000   -1.4245311887791383    2.4673932588460490
   Re   C 
     3     1
Direct
    0.5000000000000000    0.6666666666666643    0.6666666666666643     Re1
    0.8049243600558619    0.3333333333333357    0.3333333333333357     Re2
    0.1950756399441381    0.3333333333333357    0.3333333333333357     Re3
    0.0000000000000000    0.6666666666666643    0.6666666666666643      C1
\end{verbatim}
\end{example}
\begin{example}
\begin{verbatim}
VASPStructure of Os3Re
 1.000000
    8.7432980292995008    0.0000000000000000    0.0000000000000000
    0.0000000000000000    1.3846334542329621    2.3982883660601972
    0.0000000000000000   -1.3846334542329621    2.3982883660601972
   Re   Os
     1     3
Direct
    0.0000000000000000    0.3333333333333355    0.3333333333333355     Re1
    0.7492665073023750    0.6666666666666643    0.6666666666666643     Os1
    0.2507334926976250    0.6666666666666643    0.6666666666666643     Os2
    0.5000000000000000    0.3333333333333355    0.3333333333333355     Os3
\end{verbatim}
\end{example}
}}

\subsection{Supplementary information for evaluation metrics}\label{app:more_eval_method}
\subsubsection*{Success Rate for BBBP and CYP Optimization}

For BBBP optimization, our goal is to enhance the BBBP ability of the given compounds. These compounds are selected from the test set of the BBBP dataset in MoleculeNet, and initially, none can cross the BBB. For compounds generated by our AI method, we use BioT5 to predict their ability to cross the BBB. If a compound is predicted to cross, the optimization is considered successful.

For CYP optimization, the objective is to decrease the inhibition ability. Our prediction model uses a sigmoid function in the final layer, where $0$ indicates inhibition and $1$ indicates no inhibition. For an input molecule A and output molecule B, with predicted values $p_a$ and $p_B$, if $p_a>p_b$, the optimization is deemed successful.

\subsection{Shift the focus from general text to scientific sequences}\label{app:compare_with_galactica}
Although there are certain sequence-based foundation models for scientific tasks, their main focus is on text-based tasks and scientific understanding, instead of scientific discovery, i.e., discovering new molecules, proteins, and material. In Table~\ref{tab:comparison_galactica_ourM}, we compare \ourM{} with several sequence models.    

\begin{table}[!htbp]  
    \centering
    \begin{tabular}{@{}p{1.9cm}p{5cm}p{5cm}@{}} 
        \toprule  
        \textbf{Model} & \textbf{BioGPT} & \textbf{MolXPT} \\   
        \midrule  
        \textbf{Scope} & Biomedical literature & Text and SMILES \\   
        \midrule  
        \textbf{Core Capabilities} &   
        \begin{tabular}[t]{@{}p{5cm}@{}}  
            Biomedical natural language processing
        \end{tabular} &   
        \begin{tabular}[t]{@{}p{5cm}@{}}
        SMILES understanding and generation
        \end{tabular} \\   
        \midrule
        \textbf{Representative Tasks} &   
        \begin{tabular}[t]{@{}p{5cm}@{}}  
            \textbullet\ Biomedical relation extraction \\  
            \textbullet\ Biomedical question answering \\
            \textbullet\ Biomedical document classification \\
        \end{tabular} &   
        \begin{tabular}[t]{@{}p{5cm}@{}} 
            \textbullet\ Molecule property prediction \\
            \textbullet\ Text-molecule translation \\  
        \end{tabular} \\   
        \midrule
        \textbf{Training Data} &   
        \begin{tabular}[t]{@{}p{5cm}@{}}  
            \textbullet\ Text only\\ 
            \textbullet\ PubMed items before 2021\\ 
            \textbullet\ 15M paper titles and abstracts \\
        \end{tabular} &   
        \begin{tabular}[t]{@{}p{5cm}@{}} 
        \textbullet\ 67\% pure text tokens \\
            \textbullet\ 30M paper titles and abstracts from PubMed \\  
            \textbullet\ 30M SMILES from PubChem \\
            \textbullet\ 8M interleaved sequences between SMILES and text
        \end{tabular} \\   
        \midrule  
        \textbf{Training Strategy} &   
        Trained from scratch &  Trained from scratch \\
        \bottomrule
    \end{tabular}  
    \begin{tabular}{@{}p{1.9cm}p{5cm}p{5cm}@{}} 
        \toprule  
        \textbf{Model} & \textbf{Galactica} & \textbf{\ourM{}} \\   
        \midrule  
        \textbf{Scope} & Academic literature & Broader ``language of nature'' \\   
        \midrule  
        \textbf{Core Capabilities} &   
        \begin{tabular}[t]{@{}p{5cm}@{}}  
            \textbullet\ Scientific knowledge and reasoning \\  
            \textbullet\ Scientific writing assistance \\  
        \end{tabular} &   
        \begin{tabular}[t]{@{}p{5cm}@{}}  
            \textbullet\ Scientific entity generation \\  
            \textbullet\ Scientific entity optimization \\ 
        \end{tabular} \\   
        \midrule
        \textbf{Representative Tasks} &   
        \begin{tabular}[t]{@{}p{5cm}@{}}  
            \textbullet\ Scientific Q\&A \\  
            \textbullet\ Citation prediction \\
            \textbullet\ Equation recall \\
        \end{tabular} &   
        \begin{tabular}[t]{@{}p{5cm}@{}} 
            \textbullet\ Molecule optimization \\
            \textbullet\ Protein-to-molecule design \\  
            \textbullet\ Guide RNA engineering \\ 
        \end{tabular} \\   
        \midrule
        \textbf{Training Data} &   
        \begin{tabular}[t]{@{}p{5cm}@{}}  
            \textbullet\ More than 90\% pure text tokens \\  
            \textbullet\ Academic text (e.g., papers, knowledge bases)  
        \end{tabular} &   
        \begin{tabular}[t]{@{}p{5cm}@{}}  
            \textbullet\ 10\% pure text tokens \\  
            \textbullet\ Diverse scientific sequences (e.g., SMILES, FASTA, DNA, RNA, material, text.)  
        \end{tabular} \\   
        \midrule  
        \textbf{Training Strategy} &   
        Trained from scratch &  Continual pre-training on existing LLMs. Incorporates domain-specific instructions \\
        \bottomrule
    \end{tabular}  
    \caption{Comparison between existing sequence models and \ourM{}.}  
    \label{tab:comparison_galactica_ourM}  
\end{table}  


BioGPT~\cite{biogpt2022} and MolXPT~\cite{liu2023molxpt} are designed for the biomedical and (small) molecular domains. BioGPT is trained with titles and abstracts from PubMed items. MolXPT is trained with PubMed items as well as SMILES from PubChem. Their core capabilities are natural language tasks.
Galactica~\cite{galactica2022} is primarily designed for understanding and reasoning about academic literature. Its core capabilities include recalling equations, answering scientific questions, and performing domain-specific reasoning such as predicting chemical reactions and deriving mathematical proofs. 
It is trained on a highly curated corpus, primarily consisting of academic texts such as research papers (e.g., arXiv, PMC), reference materials, knowledge bases, LaTeX equations, and structured factual datasets. 
Notably, \textbf{over 90\%} of Galactica's training data consists of pure text, reflecting its emphasis on ``academic text'' and its key application in scientific writing.

In contrast, \ourM{} envisions a broader ``language of nature'' that unifies multiple scientific domains and modalities. It is explicitly designed to process diverse sequence-based data, including small molecules (SMILES), proteins (FASTA), materials (composition, space group, and atomic coordinates), as well as DNA and RNA sequences. 	Unlike Galactica, which focuses on understanding and reasoning within scientific text, \ourM{} focuses on generative tasks for scientific discovery, especially cross-domain generation and optimization tasks, such as protein-to-molecule design or guide RNA engineering. 

Only \textbf{10\%} training data of \ourM{} is pure text. The remaining \textbf{90\%} consists of scientific entities and cross-domain sequences. Furthermore, \ourM{} incorporates cross-domain data where text is interlinked with SMILES, FASTA, and material representations, enabling it to span multiple scientific disciplines through sequence-based formats. This emphasis on structured scientific data allows \ourM{} to bridge multiple domains and facilitates discovery-oriented tasks beyond text-based scientific reasoning.